\documentclass[letterpaper]{article} 
\usepackage{aaai2026}  
\usepackage{times}  
\usepackage{helvet}  
\usepackage{courier}  
\usepackage[hyphens]{url}  
\usepackage{graphicx} 
\usepackage{natbib}  
\usepackage{caption} 
\usepackage{algorithm}
\usepackage{algorithmic}

\usepackage{amssymb}
\def\P{\mathbb{P}}
\def\E{\mathbb{E}}
\def\R{\mathbb{R}}

\def\S{\mathbb{S}}
\def\A{\mathbb{A}}
\def\P{\mathbb{P}}
\def\g{\gamma}

\newcommand{\argmin}{\mathop{\mathrm{argmin}}}

\def\dt{{\Delta t}}
\newcommand{\coef}[1]{a^{(#1)}}
 
\usepackage{dsfont}
\usepackage{graphicx,graphics}
\usepackage{amsmath}
\usepackage{enumitem}
\usepackage{booktabs} 
\usepackage{nicefrac} 
\usepackage{yfonts}
\usepackage[euler]{textgreek}
\usepackage{listings}
\usepackage{MnSymbol}
\usepackage{silence}

\usepackage{caption}
\usepackage{subcaption}

\def\P{\mathbb{P}}
\def\E{\mathbb{E}}
\def\R{\mathbb{R}}

\newtheorem{theorem}{Theorem}

\newtheorem{proof}{Proof}[section]
\newtheorem{lemma}{Lemma}
\newtheorem{corollary}[theorem]{Corollary}

\title{Variance Reduction via Resampling and Experience Replay}

\makeatletter
\newcounter{corrfn}
\def\corrauthor{%
  \ifnum\value{corrfn}=0%
    \thanks{Co-corresponding authors}%
    \setcounter{corrfn}{\value{footnote}}%
  \else%
    \footnotemark[\value{corrfn}]%
  \fi%
}
\makeatother

\author {
    Jiale Han,
    Xiaowu Dai\corrauthor,
    Yuhua Zhu\corrauthor
}
\affiliations {
    Department of Statistics and Data Science, UCLA\\
    \{jialehan, daix, yuhuazhu\}@ucla.edu
}

\begin{document}

\maketitle

\begin{abstract}
Experience replay is a foundational technique in reinforcement learning that enhances learning stability by storing past experiences in a replay buffer and reusing them during training. Despite its practical success, its theoretical properties remain underexplored. In this paper, we present a theoretical framework that models experience replay using resampled $U$- and $V$-statistics, providing rigorous variance reduction guarantees. We apply this framework to policy evaluation tasks using the Least-Squares Temporal Difference (LSTD) algorithm and a Partial Differential Equation (PDE)-based model-free algorithm, demonstrating significant improvements in stability and efficiency, particularly in data-scarce scenarios. Beyond policy evaluation, we extend the framework to kernel ridge regression, showing that the experience replay-based method reduces the computational cost from the traditional $O(n^3)$ in time to as low as $O(n^2)$ in time while simultaneously reducing variance. Extensive numerical experiments validate our theoretical findings, demonstrating the broad applicability and effectiveness of experience replay in diverse machine learning tasks.
\end{abstract}

\begin{links}
    \link{Code}{https://github.com/JialeHan22/Variance-Reduction-via-Resampling-and-Experience-Replay}
\end{links}

\section{Introduction}\label{sec:intro}
Experience replay is widely recognized for enhancing learning stability by storing past experiences in a memory buffer and reusing them during training \citep{lin1992self, mnih2015human}. Rather than processing each experience only once, experience replay randomly samples batches of experiences to update learning targets, increasing sample efficiency and improving model performance. This approach has become a key component in modern reinforcement learning (RL), driving breakthroughs in applications such as Atari games \citep{mnih2015human} and AlphaGo \citep{silver2016mastering}.
However, despite its widespread success, the theoretical understanding of experience replay remains limited, often requiring extensive trial and error for effective application \citep{zhang2017deeper, fedus2020revisiting}. To address this gap, we propose a theoretical framework that connects experience replay to resampled $U$- and $V$- statistics \citep{frees1989infinite, shieh1994infinite}. This framework establishes rigorous variance reduction guarantees, providing a deeper understanding of how experience replay enhances learning stability. 

Building on prior work on $U$- and $V$- statistics  \citep{zhou2021v, peng2019asymptotic}, which primarily focused on decision-tree-based methods like random forests, we extend this framework to encompass a broader class of learning functions. We derive the asymptotic variance of learned estimators, demonstrating that estimators employing experience replay achieve asymptotically lower variance compared to their original methods. To validate our framework, we analyze variance reduction through experience replay in two important machine-learning problems: policy evaluation in RL and supervised learning in reproducing kernel Hilbert space (RKHS). 

\begin{figure}[htb]
    \centering
   \begin{subfigure}{0.23\textwidth}
        \includegraphics[width=1\textwidth]{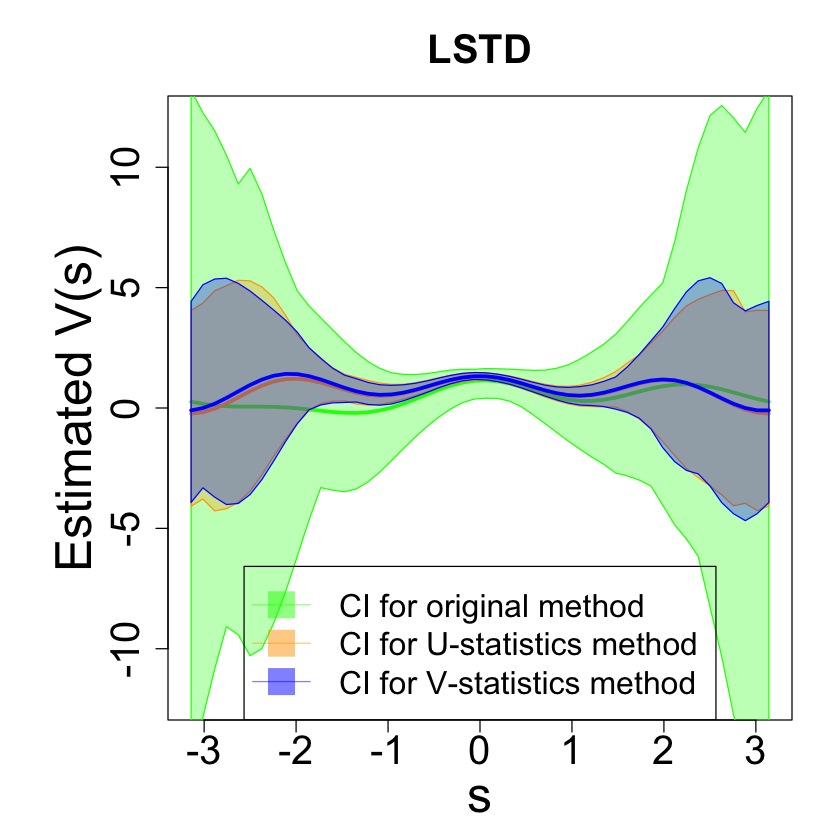} 
        \caption{LSTD approach.}
        \label{fig:figure1}
   \end{subfigure}
        \begin{subfigure}{0.23\textwidth}
        \includegraphics[width=1\textwidth]{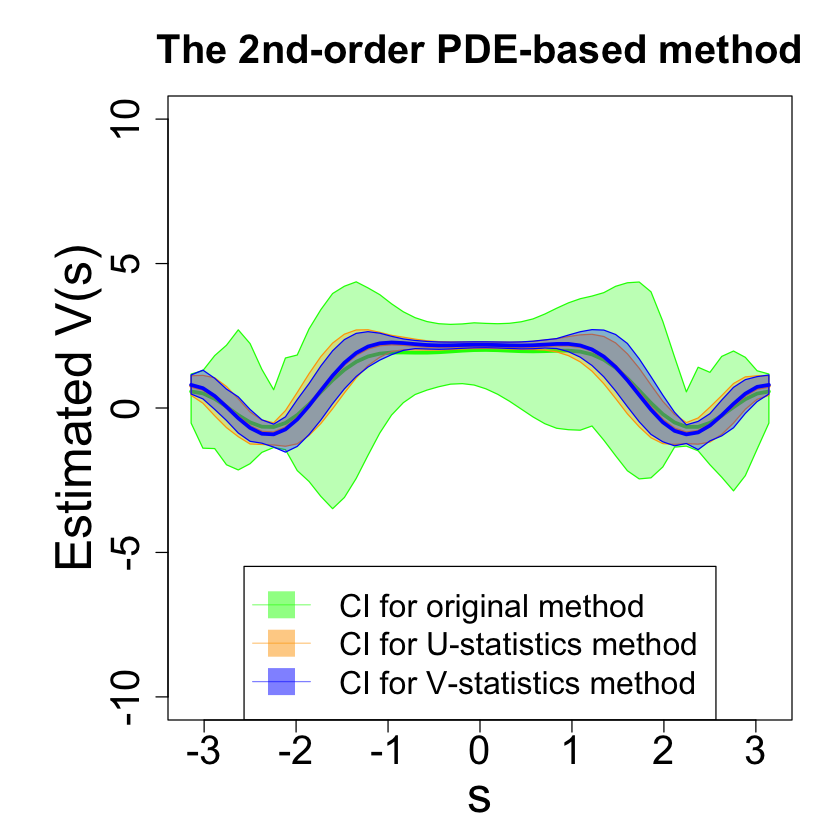} 
        \caption{PDE-based approach.}
       \label{fig:figure3}
    \end{subfigure}
    \caption{{Variance reduction achieved by experience replay in policy evaluation using two approaches. \textcolor{black}{$U$- and $V$- statistics methods incorporate experience replay without and with replacement, respectively, into the original method.     The solid lines represent the mean estimates, and the shaded areas denote the 95\% confidence intervals (CIs), calculated from 50 data replications.}}}
    \label{fig:three_figures}
\end{figure}

Policy evaluation is a critical component of RL, where the goal is to estimate the value function representing the expected cumulative reward under a given policy. Stable and accurate policy evaluation significantly impacts the overall performance of RL algorithms.  
We demonstrate the effectiveness of experience replay in two policy evaluation algorithms: (i) the Least-Squares Temporal Difference (LSTD) algorithm for Markov Decision Process (MDP) \citep{bradtke1996linear}, and (ii) the PDE-based algorithm for environments with continuous-time state dynamics \citep{zhu2024phibe}.
LSTD is a widely used data-efficient policy evaluation method that approximates value functions within a linear space by solving a least-squares problem derived from the Bellman equation \citep{bradtke1996linear}. The PDE-based approach is a novel method that employs a PDE framework to approximate the continuous-time value function and is tailored for environments where the state variable evolves continuously over time, which is a common scenario in applications such as autonomous driving \citep{kong2015kinematic} and robotics \citep{kober2013reinforcement}, where discrete-time models like MDP may fail to capture the complexity of the environment. 
Incorporating experience replay into these algorithms significantly enhances their stability by reducing variance, as illustrated in Figure \ref{fig:three_figures}. Rather than the original method which uses the entire dataset at once, the experience replay method resamples subsets, either without or with replacement, from the original dataset. These subsets are used to generate predictions, which are then averaged using resampled $U$- or $V$-statistics to produce the final prediction. This resampling approach enables data duplication, mitigates variability in predictions due to changes in the dataset, and enhances stability through the averaging process. 
This improvement is particularly important in practice, as the numerical results in \citet{zhu2024phibe} indicate that the original  RL algorithm solution can exhibit substantial instability.

While experience replay methods have been extensively validated empirically in RL, our contribution lies in providing a theoretical framework that explains why experience replay is effective in practice, particularly for policy evaluation. The experience replay technique can be further improved by incorporating extensions such as prioritized experience replay based on importance sampling \citep{schaul2015prioritized}. Our theoretical framework can also be extended to analyze these variants.

Besides RL, we apply our framework to supervised learning tasks using kernel ridge regression, where each regression sample is treated as an experience. Kernel ridge regression enhances modeling flexibility by leveraging reproducing kernel methods to map data into RKHS \citep{wahba1990spline, shawe2004kernel}. Unlike existing divide-and-conquer strategies that partition datasets into disjoint subsets to reduce computational costs \citep{zhang2013divide}, our approach employs experience replay to repeatedly draw random subsamples, providing a novel strategy to improve computational efficiency.
With appropriate parameter choices, our method 
reduces the computational cost of kernel ridge regression from the traditional  $O(n^3)$ to  $O(n^{2})$. 
At the same time, our theoretical results guarantee that the variance of the predictions is lower than that of the original kernel ridge regression method.
Hence, incorporating experience replay leads to
more stable and faster predictions in supervised learning tasks.

We validate the effectiveness of our proposed framework through extensive experiments. 
The results consistently demonstrate that experience replay significantly reduces the variance of predictions compared to methods without it, highlighting its ability to enhance stability in both reinforcement learning and supervised learning tasks. Additionally, it reduces the computational cost in kernel ridge regression with an appropriate choice of parameters. Notably, experience replay generally leads to both reduced variance and lower mean squared error in predictions.

The rest of the paper is organized as follows. Section \ref{sec:background} introduces the background of experience replay and its connection to resampled $U$- or $V$-statistics. 
Section \ref{sec:main} defines the resampled estimators, establishes their variance reduction guarantees, and discusses applications in policy evaluation and supervised learning tasks. 
Section \ref{sec:experiments} presents numerical experiments to validate the theoretical findings. Section \ref{sec:conclusion} concludes the paper with potential future directions. All technical proofs are provided in the Appendix. 

\section{Background}\label{sec:background}

\paragraph{Experience Replay.}
Experience replay stores past data in a replay buffer, denoted as $\mathcal{D}_n = \{Z_1, \dots, Z_n\}$, where $n$ represents the sample size, commonly referred to as the replay capacity in the context of experience replay \citep{lin1992self}. The replay ratio $B\geq 1$ denotes the number of batches sampled from the buffer during each update step. 
In practice, uniform sampling is the most common strategy for selecting data from the replay buffer, although more computationally expensive alternatives, such as prioritized experience replay, are also used \citep{zhang2017deeper, schaul2015prioritized}. This paper establishes theoretical guarantees for replay-based methods under uniform sampling. The proposed framework, however, can be extended to non-uniform (importance) sampling, as discussed in Appendix~\ref{extend}.

At each update step, we sample $B$ subsets of data points, $\{b_1, \dots, b_B\}$, where each subset $b_i$ $ ( i=1,\dots, B)$ contains $k\leq n$ data points. 
The learning method is represented by a function $h_k$, which takes $k$ data points as input. The response with experience replay is then computed as the average over these $B$ subsets:
\begin{equation}\label{replay 1}
       \frac{1}{B} \sum_{i} h_k(b_i). 
\end{equation}
In experience replay for Q-learning, each data point in the replay buffer $\mathcal{D}_n$ corresponds to a single transition, and $k=1$ \citep{fedus2020revisiting}. 
This paper studies algorithms such as LSTD, where $k$ could increase with $n$. LSTD is a foundational and actively studied RL algorithm for policy evaluation \cite[e.g., ][]{tu2018least, duan2024optimal}, which serves as an essential step to theoretically understand experience replay in other RL methods, such as Q-learning.

\paragraph{Connection to Resampled Statistics.}
To analyze the properties of experience replay \eqref{replay 1}, we consider, for clarity of exposition, a setting where the replay buffer $\mathcal{D}_n$ contains $n$ i.i.d. observations drawn from an underlying distribution $F_Z$ over the space $\mathcal{Z}$. 
The i.i.d. assumption can be relaxed in various ways without affecting our results (see Appendix \ref{relax}). We allow $B$ and $k$ to depend on $n$, with $k$ increasing in $n$.
This ensures that the function $h_k$ can use more information as the data size grows.

When the sampling strategy is uniform sampling \emph{without} replacement, the computation in \eqref{replay 1} takes the form of an incomplete, infinite order (or \emph{resampled}) $U$-statistics \citep{frees1989infinite, zhou2021v}, defined as:
\begin{equation}\label{3}
    U_{n,k,B} = \frac{1}{B}\sum_{i}h_k(Z_{i_1},\dots, Z_{i_{k}}).
\end{equation}
where infinite order  means that $k$ and $B$ depend on the value of $n$, and $\{Z_{i_1},\dots, Z_{i_{k}}\}$ are drawn without replacement from $\{Z_1,\dots, Z_n\}$. 
In contrast, with uniform sampling \emph{with} replacement, the computation in \eqref{replay 1} follows the form of an incomplete, infinite order (or \emph{resampled}) $V$-statistics \citep{shieh1994infinite, zhou2021v}, given by: 
\begin{equation}\label{5}
    V_{n,k,B} = \frac{1}{B}\sum_{i}h_k(Z_{i_1},\dots, Z_{i_{k}}),
\end{equation}
where $k$ and $B$ again depend on $n$, and the $B$ subsets are drawn with replacement from all size-$k$ permutations of $\{1, \dots , n\}$.

Under appropriate regularity conditions, both resampled $U$-statistics and $V$-statistics are asymptotically normal \citep{mentch2016quantifying,zhou2021v}.
The variances of these statistics can be expressed as a linear combination of $ \frac{k^2}{n} \zeta_{1,k}$ and $ \frac{1}{B}\zeta_{k,k}$.
For a given $c$, $1\leq c\leq k$, the variance components $\zeta_{c,k}$ are defined as 
\begin{equation*}
   \text{Cov}\Big(h_k(Z_1,\dots,Z_{k}), h_k(Z_1,\dots, Z_c,Z_{c+1}^{'},\dots,Z_{k}^{'})\Big),
\end{equation*}
where $Z_{c+1}^{'},\dots,Z_{k}^{'}$ are i.i.d. copies from  $F_Z$, independent of the original data set $\mathcal{D}_n$.
\paragraph{Learning Target.}
We focus on estimating the quantity defined as,
\begin{equation}
\label{eqn:defoftheta}
    \theta = \big[\E[g(Z)]\big]^{-1}\big[\E[f(Z)]\big] \in\R^q,
\end{equation}
where $g(\cdot): \mathcal{Z} \to \R^{q \times q}$ is a function returning an invertible matrix, and $f(\cdot): \mathcal{Z} \to \R^q$.
The target $\theta$ arises in various machine learning applications, including policy evaluation algorithms in reinforcement learning \citep{bradtke1996linear, zhu2024phibe}, and supervised learning with kernel ridge regression \citep{wahba1990spline, rahimi2007random}. We will discuss the application of experience replay to these methods in Section \ref{sec:app_eg}.

To estimate $\theta$ in \eqref{eqn:defoftheta}, we use a function $h_k$ based on $k\leq n$ data points $Z_{1}^*,Z_{2}^*, \dots,Z_{{k}}^*$ for any $Z_{i}^*\in\mathcal{D}_n$, $i=1,\dots, k$, where $h_k$ in \eqref{replay 1} is defined as:
\begin{equation}\label{new_hkn}
    h_{k}(Z_{1}^*,\dots,Z_{{k}}^*):=
    \Big[\sum\limits_{i=1}^{k}g(Z_{i}^*) \Big]^{-1}\Big[\sum\limits_{i=1}^{k}f(Z_{i}^*)\Big]\in\R^q.
\end{equation}
{The learning function in \eqref{new_hkn} provides a unified framework that applies to several algorithms, including the LSTD algorithm in reinforcement learning and kernel ridge regression in supervised learning. }
We will theoretically show that incorporating the experience replay approach \eqref{replay 1} reduces the variance of the estimate of $\theta$ and thus improves stability.

\section{Main Results}\label{sec:main}

\subsection{Theoretical Guarantees}\label{sec:theoretical}
\paragraph{Estimators without Experience Replay.}
When the experience replay approach is not used, and each data point in the replay buffer $\mathcal{D}_n$ is used only once, a plug-in estimator for $\theta$ in \eqref{new_hkn} is:
\begin{equation}\label{new_9}
    \tilde{\theta}_n :=\Big[\sum_{i=1}^n g(Z_i)\Big]^{-1}\Big[\sum_{i=1}^n f(Z_i)\Big].
\end{equation}
The asymptotic property of  $\tilde{\theta}_n$ is described in the following lemma. The proof relies on the central limit theorem and the delta method. 
\begin{lemma}\label{new_T1}
    Let $Z_1,Z_2,\ldots,Z_n\stackrel{iid}{\sim}F_Z$ and $\tilde{\theta}_n$ defined in \eqref{new_9}, we have that $ \sqrt{n}\left[ \tilde{\theta}_n-\theta\right]\xrightarrow{d} N(0,\Sigma_{}),$
where $\Sigma$ is a constant matrix given by 
\begin{equation*}
    G\begin{pmatrix}

    \text{Var}(f(Z)) &  \text{Cov}(f(Z),\text{vec}( g(Z))) \\
     \text{Cov}(f(Z),\text{vec}( g(Z))) & \text{Var}(\text{vec}( g(Z))

    \end{pmatrix}G^\top,
\end{equation*}
with $ G=\left([\E[g(Z)]]^{-1},-\theta^\top\otimes[\E[g(Z)]]^{-1}  \right),$ where $\otimes$ denotes the Kronecker product, and $\text{vec}(A)$   reshapes a matrix $A$ into a column vector by stacking its columns sequentially.
\end{lemma}
\noindent

\paragraph{Estimators with Experience Replay.}
Using the experience replay approach, we propose two new estimators for $\theta$ that leverage resampling methods based on $U$- and $V$-statistics.  These estimators are constructed using the learning method $h_k$ defined in \eqref{new_hkn},
\begin{equation}\label{new_mu_u}
    \hat{\theta}_U:=U_{n,k,B}=\frac{1}{B}\sum_{i}h_{k}(Z_{i_1},\dots,Z_{i_{k}}),
\end{equation}
\begin{equation}\label{new_mu_v}
    \hat{\theta}_V:=V_{n,k,B}=\frac{1}{B}\sum_{i}h_{k}(Z_{i_1},\dots,Z_{i_{k}}),
\end{equation}
where $U_{n,k,B}$ and $V_{n,k,B}$ are resampled $U$- and $V$-statistics defined in \eqref{3} and \eqref{5}, respectively. Algorithm \ref{alg:method} outlines the procedure for computing these estimators. The following theorem establishes that the $U$-statistics-based estimators achieve lower variances than the original estimator under general conditions.

\begin{algorithm}[t!]
\caption{ \normalsize{Estimating $\theta$ via Different Methods}}
\begin{algorithmic}[1]
\STATE   \normalsize{\textbf{Input:} Replay buffer $\mathcal{D}_n=\{Z_1,\dots,Z_n\}$; Functions $f$ and $g$;   Replay ratio (number of subsamples) $B$; Subsample size $k$. }
\STATE  \normalsize{\textbf{Original Estimator:} Compute $\tilde{\theta}_n$ using \eqref{new_9}. }
\STATE  \textbf{Resampled Estimators Based on $U (V)$-statistics:}
\STATE \textbf{for} $i=1$ to $B$ \textbf{do}
\STATE  \quad Randomly drawn $k$ samples $\{Z_{i_1},\dots, Z_{i_k}\}$ without (for $U$-statistics) or with replacement (for $V$-statistics).
\STATE  \textbf{end for}
\STATE   Compute $\hat{\theta}_U$ or $\hat{\theta}_V$ using   \eqref{new_mu_u} or  \eqref{new_mu_v}, respectively. 
\STATE  \textbf{Output:} Estimators $\tilde{\theta}_n$, $\hat{\theta}_U$, and $\hat{\theta}_V$.
\end{algorithmic}
\label{alg:method}
\end{algorithm}

\begin{theorem}[Variance Reduction for $U$-Statistics]
\label{new_u_incomplete}
Let $Z_1,Z_2,\ldots,Z_n\stackrel{iid}{\sim}F_Z$, and define $ \hat{\theta}_U$ as in \eqref{new_mu_u} and $\tilde{\theta}_n$  as in \eqref{new_9}. 
Under the assumption that $\lim_{n\to \infty}\frac{1}{n}\zeta_{k,k}[\zeta_{1,k}]^{-1}\to 0$ and $\lim_{n\to \infty}n/(Bk)\to 0$,
we have  
\begin{equation*}
    \liminf_{n\to \infty}[\text{Var}( \tilde{\theta}_n) - \text{Var}(\hat{\theta}_U)] \geq 0.
 \end{equation*}
\end{theorem}
\noindent
The assumption $\lim_{n \to \infty}\frac{1}{n}\zeta_{k,k}[\zeta_{1,k}]^{-1}\to 0$ used by \citet{peng2019asymptotic}, ensures the asymptotic normality of the resampled $U$-statistics.  As noted in their work, this condition is typically satisfied if  $\frac{1}{k}\zeta_{k,k}[\zeta_{1,k}]^{-1}$ remains bounded, with $k = o(n)$ being sufficient.
Additionally, the theorem requires $n/Bk$ to be small, which can be achieved by selecting a large replay ratio $B$.

To analyze the variance reduction for $V$-statistics-based estimators, we define the following class of functions $\mathcal{H}=\big\{h_k: \sup\limits_{k}||\E[h_{k}(Z_{i_1},\dots,Z_{i_{k}})h_{k}(Z_{i_1},\dots,Z_{i_{k}})^\top]||_{\infty}<\infty\big\},$
where $(i_1,\dots, i_{k})$ are indices selected with replacement from  $\{1,\dots, k\}$.  
This condition, used by \citet{zhou2021v}, ensures the boundedness of the expected outer product of $h_k$. 

\begin{theorem}[Variance Reduction for $V$-Statistics]
\label{the_2}
Let $Z_1,Z_2,\ldots,Z_n\stackrel{iid}{\sim}F_Z$, and define $ \hat{\theta}_V$ as in \eqref{new_mu_v} and $\tilde{\theta}_n$ as in \eqref{new_9}, with $h_{k}\in\mathcal{H}$.
Under the assumptions  $k=o(n^{1/4})$, $\lim_{n\to\infty} k^2\zeta_{1,k}>0$, and $\lim_{n\to \infty}n/(Bk)\to 0$, 
we have  
\begin{equation*}
\liminf_{n\to \infty}[\text{Var}( \tilde{\theta}_n) - \text{Var}(\hat{\theta}_V)] \geq 0.
\end{equation*}
\end{theorem}
\noindent
The condition $\lim_{n \to \infty} k^2 \zeta_{1,k} > 0$, which is satisfied by many base learners and has been used in prior work \citep{song2019approximating, zhou2021v}, is further discussed in Appendix \ref{dis_condition}, where we show that it holds in our framework.

Theorems \ref{new_u_incomplete} and \ref{the_2} show that incorporating experience replay via resampled $U$- and $V$-statistics asymptotically reduces variance compared to the original estimator, enhancing the stability of parameter estimation. 
{Our results remain valid under more general data-generating processes beyond the i.i.d. setting, including dependent sequences such as stationary ergodic Markov chains, $\beta$-mixing processes with summable coefficients, and $m$-dependent sequences; see Appendix~\ref{relax} for details.}

\subsection{Applications to Machine Learning Problems}\label{sec:app_eg}
\paragraph{Policy Evaluation for MDP.}
Consider a MDP defined by the tuple $(\S, \A, \g, r, \P)$ \citep{sutton2018reinforcement}. Here $s\in\mathbb{S}$ denotes the state space, $a\in\A$ represents the action space, $\gamma\in(0,1)$ is a given discounted factor, $r: \S\times \A \to \R$ is the reward function, and $\P : \S\times\A \to \Delta (\S)$ denotes the probability distribution of the next state given the current state and action. The goal of MDP is to find the optimal policy $\pi^*(s)$ that maximizes the value function. Here the policy is a mapping from the state space $\S$ to action space $\A$, while the value function ${V}_{*}^\pi(s)$ measures the expected cumulative reward of an agent over the long run, defined as:
\begin{equation}\label{def of Vpi}
    {V}_{*}^\pi(s) =  \E\left[\sum_{j =  0 }^\infty \gamma^j {r}_{*}^\pi(s_{j }) \Big|s_0 = s\right],
\end{equation}
where $s_0 = s$ is the initial state, ${r}_{*}^\pi(s) = r(s,\pi(s))$ is a known reward function under the current policy, and the state at time step $j+1$ follows the transition distribution under the policy $\pi$,  $s_{j+1}\sim P^\pi(\cdot|s_j) = P(\cdot|s_j,\pi(s_j))$. In RL, one usually divides the RL problem into two parts, one is policy evaluation, which is given a policy $\pi(s)$, calculates the value function ${V}_{*}^\pi(s)$; Another is policy improvement, that improves the policy according to gradient ascent or policy iteration.

The focus of this paper is policy evaluation, which is one of the most fundamental RL problems. In the setting of RL, one does not have access to the transition distribution. Instead, the agent applies an action $a_j = \pi(s_j)$ according to the policy at each time step $ j $, and observes the next step $ s_{j+1} $, receives a numerical reward $ r^{\pi}_{*}(s_{j+1}) $. Due to the finite length of the trajectory data, it is usually impossible to compute the value function directly according to the cumulative sum \eqref{def of Vpi}. Note that the value function ${V}^{\pi}_{*}(s)$ also satisfies the following Bellman equation (BE), 
 \begin{equation}\label{BE}
      {V}_{*}^\pi(s) = {r}_{*}^\pi(s) +  \gamma\E_{s_{j+1} \sim P^\pi(s'|s_0)}[{V}_{*}^\pi(s_{j+1})|s_0 = s].
\end{equation}
Therefore, the goal of the policy evaluation problem is to find the value function that solves BE \eqref{BE} given a set of trajectory data,
\begin{equation}\label{D_n in rl}
    \mathcal{D}_n=\{(s^l_0,s^l_{1}, \dots, s^l_{L}) \}_{l=1}^n.
\end{equation}
Here the data set contains $n$ independent trajectories and each contains $L+1$ data points.
The initial state $s^l_0$ of each trajectory is sampled from a distribution $\rho^{\pi}_0(s)$. 
Our method also extends to settings with dependent data and variable-length trajectories (see Appendix~\ref{relax}).

LSTD \citep{bradtke1996linear} is a popular RL algorithm for linear approximation and can be directly used to estimate ${V}_{*}^\pi(s)$ using the trajectory data. 
LSTD approximates the value function ${V}_{*}^\pi(s) = \Phi(s)^\top \theta$ in the space expanded by $q$ given bases $\{\phi_i(s)\}_{i=1}^q$, where $\theta\in\R^q$ is a unknown parameter and $\Phi(s) = (\phi_1(s), \cdots, \phi_q(s))^\top$.
By projecting the value function into the finite bases, LSTD solves the parameter $\theta$ in the form of 
\begin{equation}\label{theta_0}
\left[\mathbb{E}_s[ \Phi(s) (\Phi(s) -\g\E[\Phi(s_1)|s_0 = s])^\top]\right]^{-1} \mathbb{E}_s[r^{\pi}_{*}(s) \Phi(s) ].
\end{equation}
Using any trajectory data subset with $ k \leq n$ data points $\{(s^{l_{(1)}}_{j})_{j=0}^L, \dots,(s^{l_{(k)}}_{j})_{j=0}^L\} $ 
 for any $(s^{l_{(i)}}_{j})_{j=0}^L\in\mathcal{D}_n,$  $i=1,\dots, k$, the estimator of $\theta$ is 
\begin{equation*}\label{rl form_1}
\left[\sum_{i=1}^{k} g((s^{l_{(i)}}_{j})_{j=0}^L)\right]^{-1}\left[\sum_{i=1}^{k}f((s^{l_{(i)}}_{j})_{j=0}^L)\right],   
\end{equation*} 
corresponds to the structure of \eqref{new_hkn}, 
where 
\begin{equation*}
    g((s^{l_{(i)}}_{j})_{j=0}^L)= \sum_{j=0}^{L-1}\Phi(s_{j}^{l_{(i)}})[\Phi(s_{j}^{l_{(i)}})-\gamma\Phi(s_{(j+1)}^{l_{(i)}})]^\top, 
\end{equation*}
\begin{equation}\label{g and f}
    f((s^{l_{(i)}}_{j})_{j=0}^L)=\sum_{j = 0}^{L-1} r^{\pi}_{*}(s^{l_{(i)}}_j)\Phi(s^{l_{(i)}}_{j}).\quad\quad\quad\quad\quad\
\end{equation}

This setup aligns with our framework, where \( Z_i = (s^i_{j})_{j=0}^L \) for \( i = 1, \dots, n \), \( \theta \) is defined in \eqref{theta_0}, and the functions \( g \) and \( f \) are defined in \eqref{g and f}.

Our theories also help explain prior empirical findings on experience replay in Q-learning \citep{zhang2017deeper, fedus2020revisiting}; see Appendix~\ref{insight} for details.

\paragraph{Policy Evaluation for Continuous-Time RL.}

In the second application,  we aim to solve the policy evaluation problem for continuous-time RL \citep[e.g.,][]{zhu2024phibe}. 
Given a policy $\pi(s)$, unlike the MDP where the value function is a cumulative sum over discrete time steps defined as \eqref{def of Vpi}, the value function in continuous-time RL is an expected integral over continuous time,
\begin{equation}\label{def of value}
    V^{\pi}(s) = \E\left[\int_{0}^\infty e^{-\beta t}{r}^{\pi}(s_t)dt\Big| s_0 = s\right]. 
\end{equation}
Here $\beta>0$ is a given discounted coefficient, $r^{\pi}(s)\in\R$ is a known reward function under the current policy. When the state $s_t\in\mathbb{S} \subset \R^d$ is driven by the stochastic differential equation (SDE),  
\begin{equation}\label{def of dynamics}
    d s_t = \mu(s_t)dt + \sigma(s_t)dB_t,
\end{equation} 
by Feynman–Kac theorem \cite{stroock1997multidimensional}, the value function $V(s)$ satisfies the equation $    \beta V^{\pi}(s) = r^{\pi}(s) + \mathcal{L}_{\mu,\Sigma} V^{\pi}(s) $,
$\text{where\ } \mathcal{L}_{\mu,\Sigma} = \mu(s)\cdot\nabla + \frac{1}{2}\Sigma : \nabla^2$ with $\Sigma = \sigma \sigma^\top$, and $\Sigma : \nabla^2 = \sum_{i,j}\Sigma_{ij}\partial_{s_i}\partial_{s_j}$. 
Similar to the classical RL setting, one does not have access to the drift function $\mu(s)\in \R^d$ and diffusion function $\sigma(s)\in \R^{d\times d}$. Therefore, one cannot solve the above equation directly.  The goal of continuous-time policy evaluation is to find the value function satisfying \eqref{def of dynamics} with a set of trajectory data $\mathcal{D}_n$ defined in \eqref{D_n in rl}. Here the data at time step $j$ are collected at time $j\dt$ with a given time interval $\dt$.

\citet{zhu2024phibe} introduced an algorithm to approximate the value function by solving a Physics-informed Bellman equation (PhiBE) defined as follows
\begin{equation}\label{def of PhiBE}
    \beta \bar V^{\pi}_{\alpha}(s) - \mathcal{L}_{
    \hat{\mu}_{\alpha},\hat{\Sigma}_{\alpha}}  \bar V^{\pi}_{\alpha}(s) = r^{\pi}(s),\quad \alpha = 1, 2
\end{equation}
where $\hat \mu_{\alpha}(s) =\frac1\dt \sum_{j=1}^{\alpha}\E_{s_{j}}\left[\coef{{\alpha}}_j(s_{j} - s_0)|s_0 = s\right]$, $\hat \Sigma_{\alpha}(s) =\frac1\dt \sum_{j=1}^{\alpha}\E_{s_{j}}\left[\coef{{\alpha}}_j(s_{j} - s_0)(s_{j} - s_0)^\top|s_0 = s\right] $ and 
\begin{equation}\label{def of A b}
\alpha = 1:\ \coef{1}_1 = 1; \quad \alpha = 2:\ \coef{2}_1 = 2, \ \coef{2}_2 = -\frac12.
\end{equation}
Here $\bar{V}^{\pi}_\alpha(s)$ serves as $\alpha$-th order approximation to the continuous-time value function $V^{\pi}(s)$, where $\alpha\in\{1,2\}$.    

Similar to \eqref{theta_0}, if one approximates the solution $\bar V(s) = \Phi(s)^\top \theta$ to the PhiBE \eqref{def of PhiBE} in the linear function space spanned by $\Phi(s)$, one ends up solving for the parameter $\theta$ in the following form
\begin{equation}\label{theta_1}
    \left[\mathbb{E}_s[ (\beta \Phi(s)^ \top - \mathcal{L}_{\hat{\mu}_{\alpha},\hat{\Sigma}_{\alpha}}  \Phi(s)^ \top ) \Phi(s)]\right]^{-1}\mathbb{E}_s[r^{\pi}(s) \Phi(s) ].
\end{equation}

\citet{zhu2024phibe} gives the model-free algorithm to estimate the $\theta$ using only trajectory data. Specifically, for the $\alpha$-th order case, using any data subset with $ k \leq n$ data points $\{(s^{l_{(1)}}_{j})_{j=0}^L, \dots,(s^{l_{(k)}}_{j})_{j=0}^L\}$ for any $(s^{l_{(i)}}_{j})_{j=0}^L\in\mathcal{D}_n,\ i=1,\dots, k$, the estimator of $\theta$ is 
\begin{equation*}\label{rl form}
    \left[\sum_{i=1}^{k} g((s^{l_{(i)}}_{j})_{j=0}^L)\right]^{-1}\left[\sum_{i=1}^{k}f((s^{l_{(i)}}_{j})_{j=0}^L)\right],
\end{equation*}
corresponds to the structure of \eqref{new_hkn}, 
where 
\begin{equation*}
    g((s^{l_{(i)}}_{j})_{j=0}^L)= \sum_{j = 0}^{L-{\alpha}}\Phi(s^{l_{(i)}}_{j})\left[ \beta \Phi (s^{l_{(i)}}_{j})- \mathcal{L}_{
    \bar{\mu}_{\alpha},\bar{\Sigma}_{\alpha}} \Phi(s^{l_{(i)}}_{j})\right]^\top, 
\end{equation*}
\begin{equation}\label{f and g 2}
        f((s^{l_{(i)}}_{j})_{j=0}^L)=\sum_{j = 0}^{L-{\alpha}} r^{\pi}(s^{l_{(i)}}_{j})\Phi(s^{l_{(i)}}_{j}),\quad\quad\quad\quad\ \quad\ 
\end{equation}
and the estimators of $\mu(s)$ and $\sigma(s)$ are defined as $\bar{\mu}_{\alpha}(s_{j}^l) = \frac{1}{\dt}\sum_{k=1}^{\alpha}\coef{\alpha}_k(s^l_{(j+k)} - s^l_{j}) $, $\bar{\Sigma}_{\alpha}(s_{j}^l) = \frac{1}{\dt}\sum_{k=1}^{\alpha}\coef{{\alpha}}_k(s^l_{(j+k)} - s^l_{j})(s^l_{(j+k)} - s^l_{j})^\top$  
with $\coef{{\alpha}}$ defined as \eqref{def of A b}.
\textcolor{black}{Compared to LSTD, the second-order PDE-based algorithm with \(\alpha = 2\) incorporates two future steps, resulting in improved accuracy, as illustrated in Figure \ref{fig:three_figures}.}
This setup aligns with our framework with $Z_i=(s^i_{j})_{j=0}^L,\  i=1,\dots, n$ and $\theta$ defined in \eqref{theta_1}, and functions $g$ and $f$ defined in \eqref{f and g 2}. 

For both LSTD and the PDE-based approach,  once we obtain $\tilde{\theta}_n$, $\hat{\theta}_U$, and  $\hat{\theta}_V$ by applying Algorithm \ref{alg:method}, the corresponding estimations of the value function at a test point $s$ are defined as
$\tilde{V}(s) = \Phi(s)^\top \tilde{\theta}_n, \hat{V}_U(s) = \Phi(s)^\top \hat{\theta}_U,  \text{\ and } \quad  \hat{V}_V(s) = \Phi(s)^\top \hat{\theta}_V,$
where the superscript $\pi$ is omitted.
The variances are, $\text{Var}(\tilde{V}(s))=\Phi(s)^\top\text{Var}(\tilde{\theta}_n)\Phi(s)$, $\text{Var}(\hat{V}_U(s))=\Phi(s)^\top\text{Var}(\hat{\theta}_U)\Phi(s)$, and $\text{Var}(\hat{V}_V(s))=\Phi(s)^\top\text{Var}(\hat{\theta}_V)\Phi(s)$. Thus the reduction of the variance of estimators of $\theta$ could be directly evaluated by the reduction in the variance of these estimations.

\paragraph{Kernel Ridge Regression.}
In the third application, we consider a supervised learning framework where $\mathcal{D}_n=\{(X_1,Y_1),\dots, (X_n, Y_n)\}$ consists of i.i.d. samples drawn from a distribution $F_Z$.
Our goal is to predict the outcome \( Y \in \mathbb{R} \) based on the predictors \( X \in \mathbb{R}^p \) using kernel methods in an RKHS \citep{wahba1990spline}. 
\textcolor{black}{Let \( K(\cdot, \cdot): \mathbb{R}^p \times \mathbb{R}^p \to \mathbb{R} \) be a reproducing kernel function. We consider the model \( Y = f(X) + \epsilon \), where \( f \) belongs to the RKHS defined by \( K \), and \( \epsilon \) represents random error independent of \( X \). Following \citet{rahimi2007random} and \citet{dai2023kernel}, the kernel function \( K(X_i, X_j) \) can be approximated as \( \phi(X_i)^\top \phi(X_j) \) using a feature mapping \( \phi: \mathbb{R}^p \to \mathbb{R}^q \), and \( f(X) \) can be approximated as \( \phi(X)^\top \theta \), where \( \theta \in \mathbb{R}^q \) is a parameter vector, defined as  
\begin{equation}
\label{eqn:thetakrr}
    \theta = \left[\mathbb{E}[\phi(X)\phi(X)^\top]\right]^{-1} \mathbb{E}[\phi(X)Y].
\end{equation}}
Using any $k$ data points $\{(X_1^*, Y_1^*), \dots, (X_k^*, Y_k^*)\}$ resampled from $\mathcal{D}_n$, the kernel ridge regression estimator of $\theta$ is obtained by solving the following optimization problem for a given $\lambda\geq 0$, 
\begin{equation*}
    \argmin_{\theta\in \R^{q}} \left\{\sum_{i=1}^{k} [Y_i^*-\phi(X_i^*)^\top\theta]^2+\lambda||\theta||^2_2\right\}.
\end{equation*}

The solution takes the form of 
\begin{equation}\label{eq41}
\left[ \sum_{i=1}^k  g(X_i^*,Y_i^*)+\lambda\mathbb{I}_p\right]^{-1}\left[  \sum_{i=1}^k f(X_i^*,Y_i^*)\right],
\end{equation} 
where $g(X_i^*, Y_i^*) = \phi(X_i^*)\phi(X_i^*)^\top$ and $f(X_i^*, Y_i^*) = \phi(X_i^*) Y_i^*$.
The setup in \eqref{eqn:thetakrr} and \eqref{eq41} aligns with our framework in \eqref{eqn:defoftheta} and \eqref{new_hkn}, with an added regularization term $\lambda \mathbb{I}_p$.  This term does not impact the derivation of our main results.

The standard computational cost of the kernel ridge regression with $n$ data points is $ O(n^3) $ in time \citep{wahba1990spline}. 
The divide-and-conquer algorithm \citep{zhang2013divide} reduces this cost by dividing the dataset into $m<n$ disjoint subsets, each of $n/m$, and averaging the local solutions across these subsets to construct a global predictor. This approach achieves a trade-off between computational cost and estimation error. 
In contrast, our approach, which incorporates the experience replay method, also averages over subsets but differs fundamentally in how the subsets are constructed. Instead of partitioning the dataset into non-overlapping subsets, we repeatedly draw $ B $ random subsamples, each containing $k$  data points. This resampling allows for overlapping subsets and potential duplication of data points, resulting in a total computational cost of 
$O(Bkq^2+Bq^3)$ in time.
Theorems \ref{new_u_incomplete} ensures that the conditions  $\lim_{n\to \infty}n/(Bk)\to 0$ and $k = o(n)$ are sufficient for variance reduction. By carefully choosing $B$ and $k$, our approach achieves both computational savings and variance reduction, offering a practical and efficient alternative to traditional kernel ridge regression, especially for large-scale problems.  For instance, setting $B= O(n^{13/8}),k = O(n^{1/8})$, $q=O(n^{1/8})$,  satisfies the conditions of Theorems \ref{new_u_incomplete} and reduces the variance. In this setup, the computational cost is further reduced to $O(n^2)$ in time. Additional discussion and examples are provided in Appendix~\ref{complexity}.

%
\section{Numerical Experiments}\label{sec:experiments}

\subsection{Experiments of Policy Evaluation Using LSTD Algorithm}\label{exp:PE}
Firstly, we present the experimental results obtained using LSTD with functions $g$ and $f$ defined in \eqref{g and f}.
We conduct the experiments in a similar setting as described in \citet{zhu2024phibe}, where the state space $\mathbb{S} = [-\pi, \pi]$, 
\textcolor{black}{and the state under policy $\pi$ is driven by the transition distribution $P^{\pi}(s_{j+1}|s_j)$ following the normal distribution with expectation $se^{\lambda/10}$, variance $\frac{\sigma^2}{2\lambda }(e^{\lambda/5}-1)$, where $\lambda = 0.05$ and $\sigma = 1$.}
The reward function is set to be $r^{\pi}_{*}(s) = 0.1*[\cos^3(s)-\lambda s (-3\cos^2(s)\sin(s)) - \frac12\sigma^2(6\cos(s)\sin^2(s) -3\cos^3(s))]$ and the discounted factor $\gamma$ is set to be $e^{-0.1}$. 
We use periodic bases $\{\phi_n(s)\}_{k=1}^{2I+1} = \frac{1}{\sqrt{\pi}}\{\frac{1}{\sqrt{2}}, \cos(is), \sin(is)\}_{i=1}^I$ with $I=4$.
We consider the case $L=2$, where each trajectory has three data points and the state $s_j^l$ in $D_n$ \eqref{D_n in rl} is sampled at time $j/10$ for $j=0,\dots, L$ and $l=1,\dots, n$.  
In each experiment, we draw $n$ independent trajectories $\mathcal{D}_n$ with the initial state $s_0^l$ of each trajectory sampled from a truncated normal distribution over $\mathbb{S}$ with mean $0$ and standard deviation $0.1$.

We check the performance of the three prediction models on $m=50$ test points evenly selected in $\mathbb{S}$, denoted by $\mathcal{S}_{test}=\{s_j^{*}\}_{j=1}^{m}$ with $s_j^{*}=-\pi+2(j-1)*\pi/(m-1)$. 
The experiment is conducted \( M = 50 \) times, and the variance of the estimated outcome for each test state \( s^*_j \), where \( j = 1, \dots, m \), is approximated using the sample variance. Three different estimators are used, resulting in approximate variances denoted by \(\text{Var}(\tilde{V}(s^*_j))\), \(\text{Var}(\hat{V}_U(s^*_j))\) and \(\text{Var}(\hat{V}_V(s^*_j))\). To assess the variance reduction property, we compare these three variances across all test states.

\begin{figure}
\centering
\includegraphics[width=0.33\linewidth]{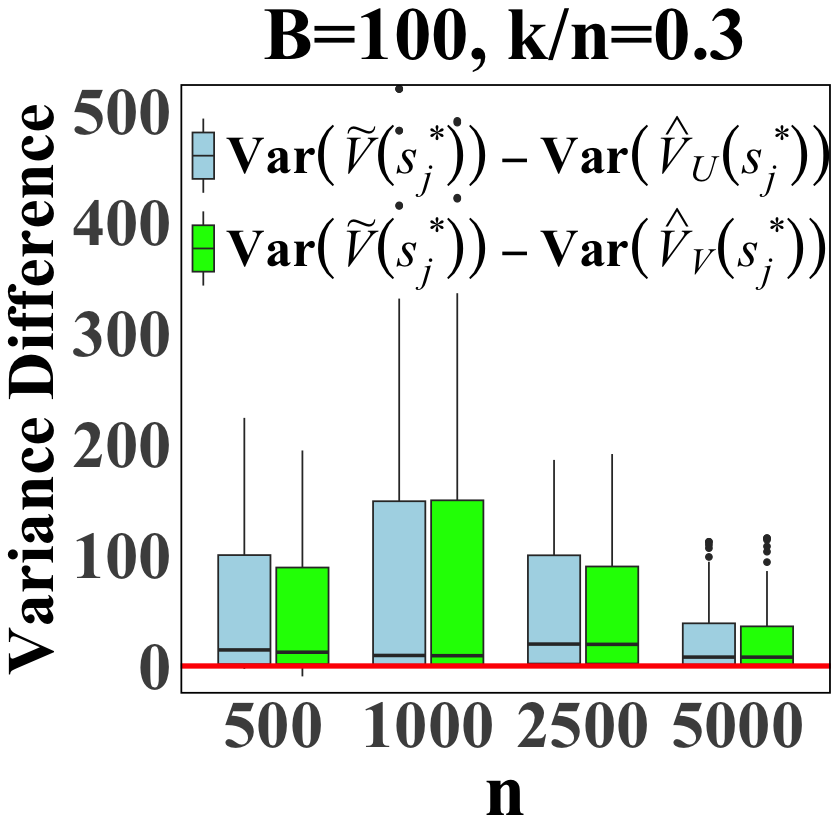}%
\includegraphics[width=0.33\linewidth]{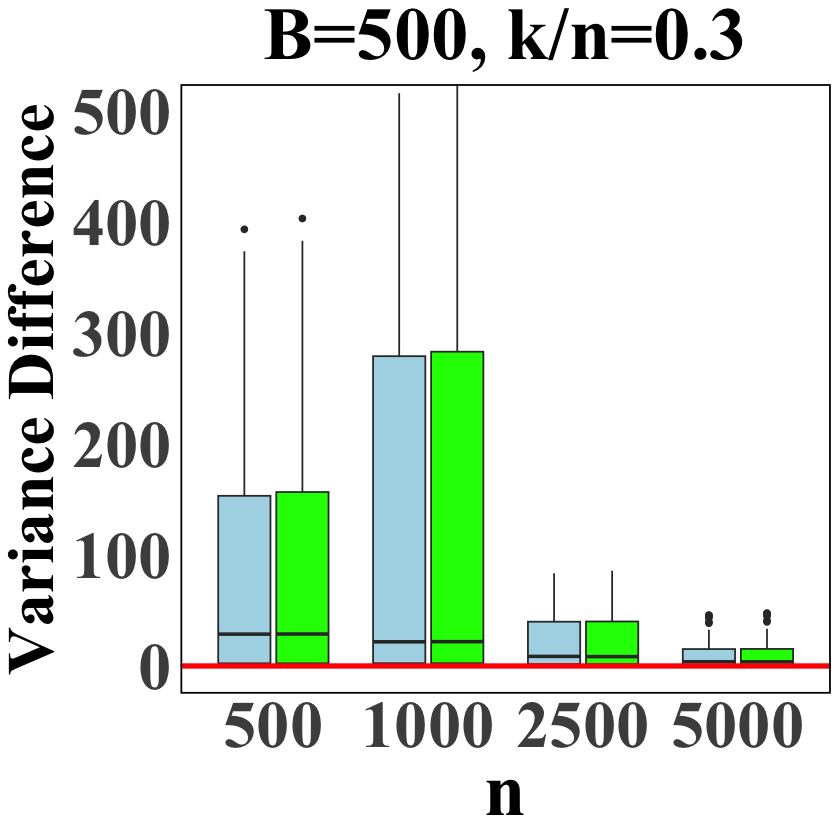}%
\includegraphics[width=0.33\linewidth]{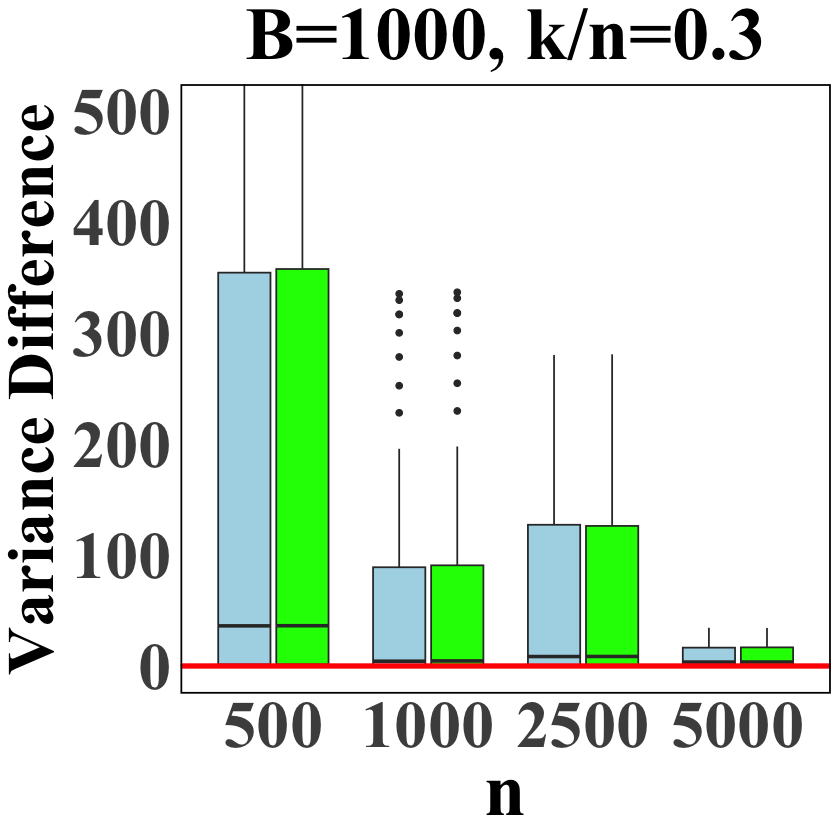}

\caption{ Variance differences among the predicted policy values using the LSTD algorithm 
with $ m = 50 $, $ M = 50 $, and $ k/n =0.3$, evaluated across various values of $ n $ and $ B $. 
\textcolor{black}{$\tilde{V}(s^*_j)$ represents the results without experience replay, while $\hat{V}_{U}(s^*_j)$ and $\hat{V}_{V}(s^*_j)$ represent the results with experience replay.}
The red line represents the baseline where the variance difference is 0.}
\label{fig:RL_LSTD}
\end{figure}

Figure \ref{fig:RL_LSTD} compares the variances using standard boxplots that display the quartile breakdown of the differences $\{\text{Var}(\tilde{V}(s_j^{*}))-\text{Var}(\hat{V}_U(s_j^{*}))\}_{j=1}^m$ and $\{\text{Var}(\tilde{V}(s_j^{*}))-\text{Var}(\hat{V}_V(s_j^{*}))\}_{j=1}^m$, with $n\in \{500, 1000,2500,5000\}$, $B=\{100, 500, 1000\}$, and $k/n=0.3$. The results clearly demonstrate that for all of the different parameters, the variance differences across all test data points are consistently greater than 0 for both $U$- and $V$-statistics-based experience replay methods, particularly in data-scarce settings.
As \( n \) increases, the variance differences become small as all estimation methods exhibit reduced variance; nonetheless, the variance reduction remains substantial. 
To illustrate this, we consider the case where $ n = 5000 $, $ B = 1000 $, and $ k/n = 0.3 $, as shown in Figure \ref{fig:figure1}. 
From the figure, we observe that the resampled methods demonstrate a significant improvement in variance in this large $ n $ scenario. 
Additional experiments with varying choices of \( k/n \) are provided in Appendix \ref{app:lstd}, further confirming the robustness of the approach. 

\subsection{Experiments of Policy Evaluation Using PDE-Based Algorithm}\label{PDE:2nd}

Secondly, we present the experimental results obtained using the second-order PDE-based algorithm with functions $g$ and $f$ defined in \eqref{f and g 2} with $\alpha=2$. 
Similar to \citet{zhu2024phibe}, we consider an experimental setting where the state dynamics are governed by the Ornstein–Uhlenbeck (OU) process 
$ds(t) = \lambda sdt + \sigma dB_t$ with $\lambda = 0.05, \sigma = 1$. The reward function is set to be $r^{\pi}(s) = \beta\cos^3(s)-\lambda s (-3\cos^2(s)\sin(s)) - 0.5\sigma^2(6\cos(s)\sin^2(s) -3\cos^3(s))$ with the discounted coefficient $\beta=0.1$. 
\textcolor{black}{For the OU process, the transition distribution $P^{\pi}(s'|s)$ from time $t$ to $t+\dt$ follows a normal distribution with mean $se^{\lambda \dt}$ and variance $\frac{\sigma^2}{2\lambda}(e^{2\lambda \dt} - 1)$. We set $\dt = 0.1$, and under this setting, $D_n$ in Section \ref{exp:PE} follows the same transition distribution, allowing us to use the same simulated trajectory data. Additionally, we employ the same periodic basis functions as described in Section \ref{exp:PE}.} 
The true value function $V^{\pi}(s)$ then can be exactly obtained from \eqref{def of value}, $V^{\pi}(s)=\text{cos}^3(s)$. 

\textcolor{black}{Note that the experiments using LSTD in Section \ref{exp:PE} can be considered as a way for estimating $ V^{\pi}(s) $ by discretizing it as a MDP. This approach uses the relationships $ r^{\pi}_{*}(s) = r^{\pi}(s)\dt $ and $ \gamma = e^{-\beta \dt} $, which hold in the given setting. However, as observed in Figure \ref{fig:three_figures}, when the original methods are used, the PDE-based approach generally shows greater accuracy with narrower confidence bands.  }

\begin{figure}
\centering
\includegraphics[width=0.33\linewidth]{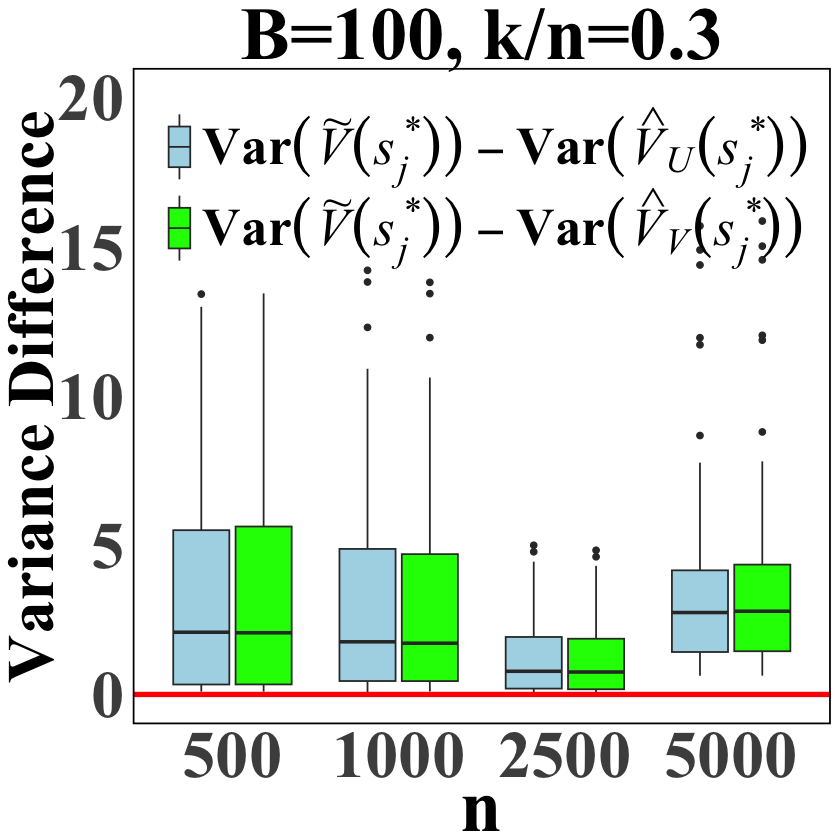}%
\includegraphics[width=0.33\linewidth]{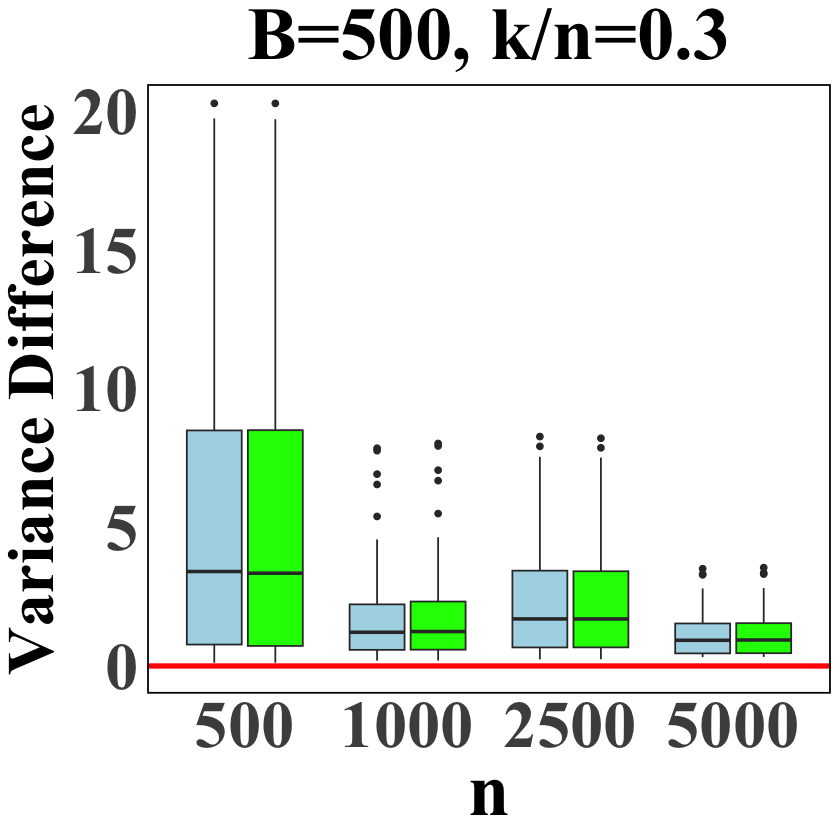}%
\includegraphics[width=0.33\linewidth]{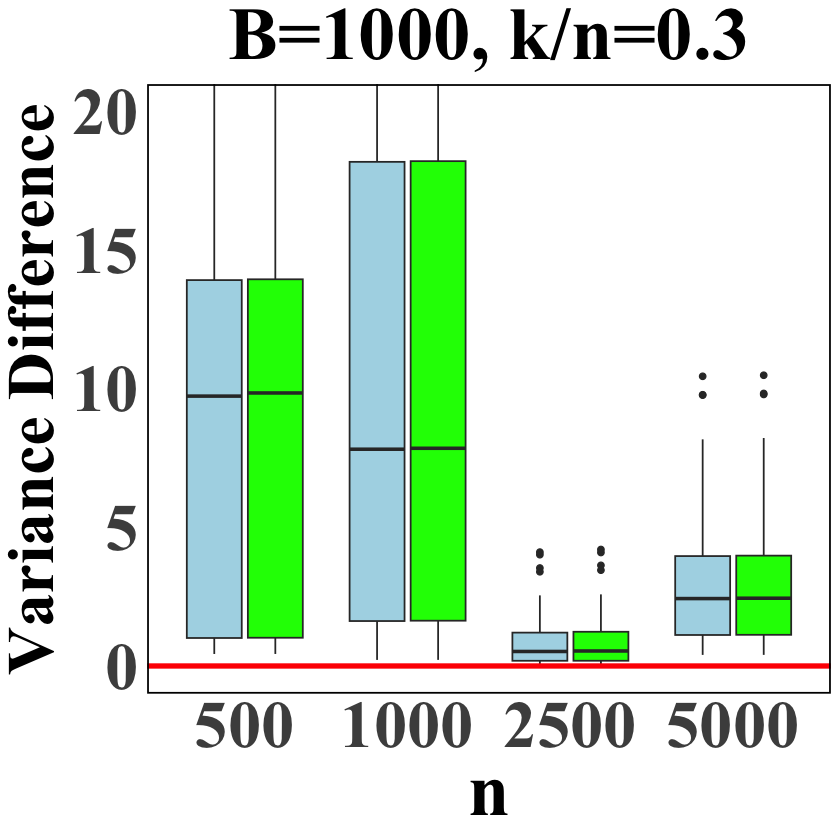}

\caption{ 
Variance differences among the predicted policy values using the second-order PDE-based algorithm with $ m = 50 $, $ M = 50 $, and $k/n=0.3$, evaluated across various values of $ n $ and $ B $. \textcolor{black}{$\tilde{V}(s^*_j)$ represents the results without experience replay, while $\hat{V}_{U}(s^*_j)$ and $\hat{V}_{V}(s^*_j)$ represent the results with experience replay.} The red line represents the baseline where the variance difference is 0.
}
\label{fig:RL_2nd}
\end{figure}

We evaluate the performance of the three prediction models using the same way as in Section \ref{exp:PE}. 
Figure \ref{fig:RL_2nd} clearly demonstrates that for all of the different parameters, the variance differences across all test data points are consistently greater than 0 for both $U$- and $V$-statistics-based experience replay methods.
Figure \ref{fig:figure3} illustrates  the large $n$ case where $ n = 5000 $, $ B = 500 $, and $ k/n = 0.3 $.

We present additional experiments with different choices of $k/n$, along with first-order results in Appendix~\ref{app:PE 1st}.
With the use of experience replay, the second-order method achieves a greater percentage reduction in variance compared to the LSTD method. Intuitively, the second-order method accounts for two future steps, introducing more stochasticity, which provides greater potential for variance reduction.
Moreover, we compare the root mean squared error (RMSE) of the proposed methods with the original method over the $ m $ test points across all $ M $ experiments for both the LSTD and PDE-based methods in Appendix \ref{rmse pe}. The results demonstrate that the combination of experience replay, regardless of the specific resampling method used, not only reduces variance but also tends to achieve smaller prediction errors, further highlighting its effectiveness.

\subsection{Experiments of Kernel Ridge Regression}\label{sec:kernel}

Thirdly, we consider a regression setting where for each $(X, Y) \sim F_Z$, the predictor $X = (X_{(1)}, X_{(2)}) \in \mathbb{R}^2$ is generated with $X_{(1)}, X_{(2)} \sim \text{Unif}(0, 1)$, and the response is given by $Y=e^{10(-(X_{(1)} - 0.25)^2 - (X_{(2)} - 0.25)^2)} + 0.5 \cdot e^{14(-(X_{(1)} - 0.7)^2 - (X_{(2)} - 0.7)^2)}  + \epsilon, $ where $\epsilon \sim \mathcal{N}(0, 0.25)$ is independent of $X$. 
This setting is widely used in the study of kernel ridge regression and generalized regression models \citep[see,][]{hainmueller2014kernel, wood2003thin}.

For each experiment, we independently draw $n$ data points from $F_Z$ to form the training dataset $\mathcal{D}_n$. 
We use the \texttt{krls} function in \texttt{R} to fit the kernel ridge regression model with a Gaussian kernel. The $\lambda$ is chosen as $n^{-2/3}$.
We evaluate the performance of these models on $m=100$ test points independently drawn from $F_Z$, denoted by $\mathcal{D}_{test}=\{ (x_j,y_j)\}_{j=1}^m$. The experiment is repeated \( M = 100 \) times, and the variances of the predicted outcomes $\tilde{y}_j, \hat{y}_{j,U}$, and $\hat{y}_{j,V}$ for each test predictor \( x_j \), where \( j = 1, \dots, m \), are approximated using the sample variances, denoted by \(\text{Var}(\tilde{y}_j)\), \(\text{Var}(\hat{y}_{j,U})\), and \(\text{Var}(\hat{y}_{j,V})\). 
\textcolor{black}{As stated in \citet{dai2023kernel}, the predictions \(\tilde{y}_j\), \(\hat{y}_{j,U}\), and \(\hat{y}_{j,V}\) are approximately equal to \(\phi(x_j)^{\top}\tilde{\theta}_n\), \(\phi(x_j)^{\top}\hat{\theta}_{U}\), and \(\phi(x_j)^{\top}\hat{\theta}_{V}\) when \(q\) is  large. 
Consequently, \(\text{Var}(\tilde{y}_j)\), \(\text{Var}(\hat{y}_{j,U})\), and \(\text{Var}(\hat{y}_{j,V})\) serve as estimates for \(\phi(x_j)^{\top}\text{Var}(\tilde{\theta}_n)\phi(x_j)\), \(\phi(x_j)^{\top}\text{Var}(\hat{\theta}_{U})\phi(x_j)\), and \(\phi(x_j)^{\top}\text{Var}(\hat{\theta}_{V})\phi(x_j)\), respectively. Therefore, the reduction in the variance of the estimators of \(\theta\) can be directly assessed by evaluating the reduction in the variance of these predictions.} We compare the variances \(\text{Var}(\tilde{y}_j)\), \(\text{Var}(\hat{y}_{j,U})\), and \(\text{Var}(\hat{y}_{j,V})\) across all test points.

\begin{figure}[htb]
\centering
\includegraphics[width=0.15\textwidth]{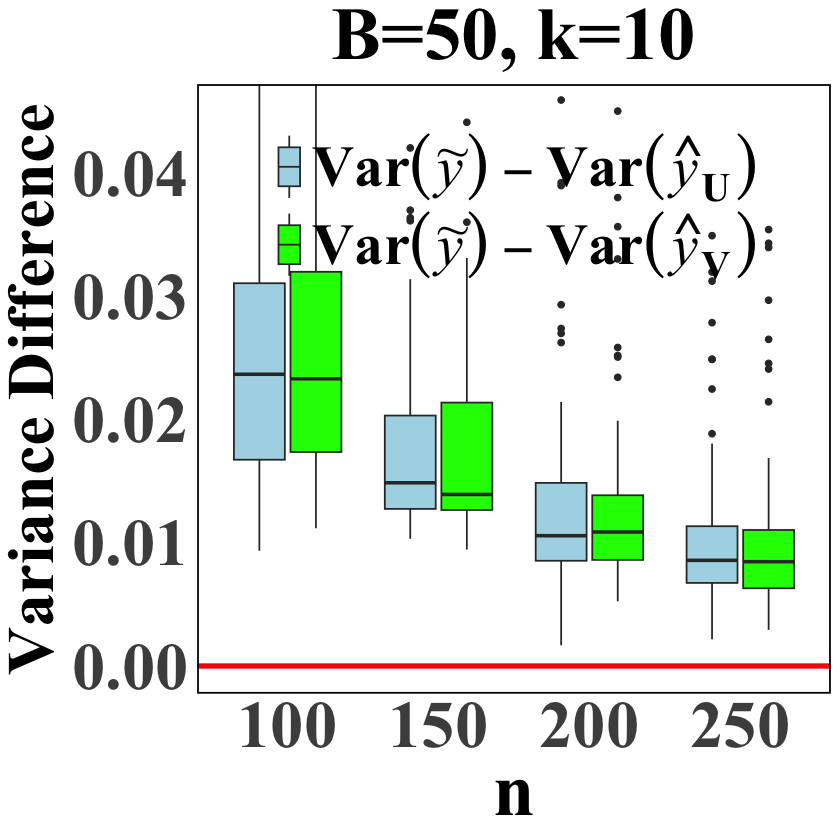} 
\includegraphics[width=0.15\textwidth]{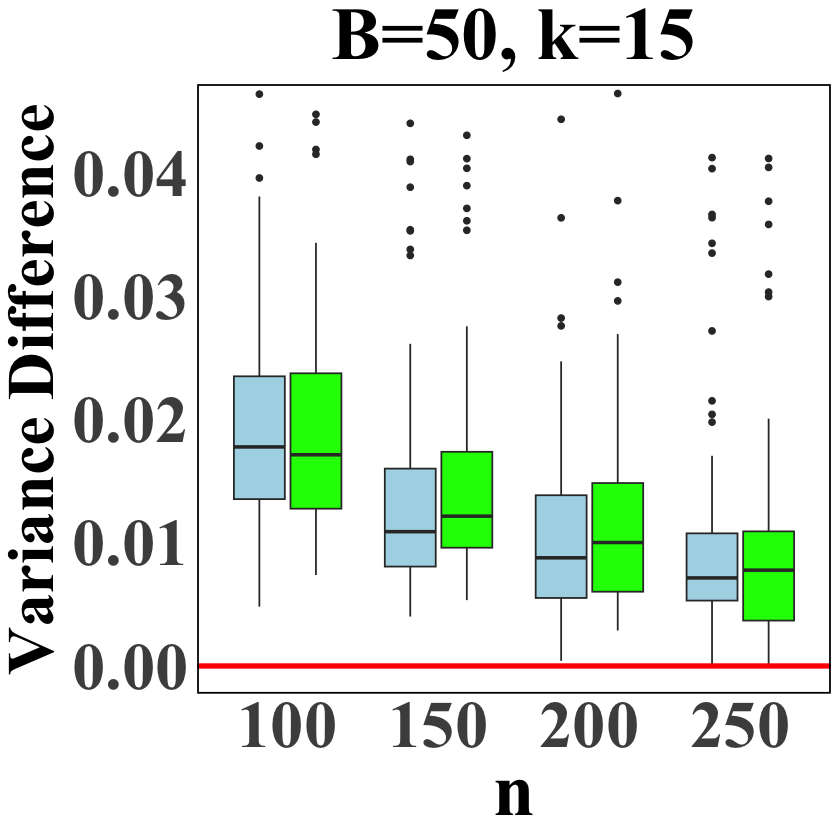} 
\includegraphics[width=0.15\textwidth]{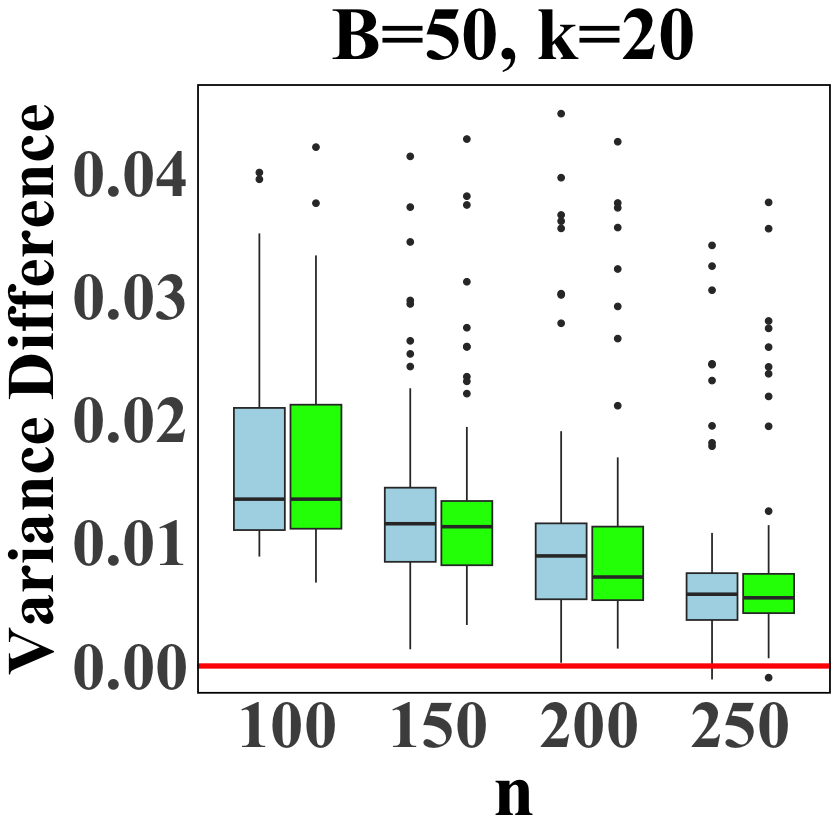} 

\caption{ Variance differences in predicted outcomes using kernel ridge regression on the simulated data with $M=100, m= 100 $ and $ B = 50 $, evaluated across various values of  $ n $ and $ k$. 
\textcolor{black}{$\tilde{y}$ represents the results without experience replay, while $\hat{y}_{U}$ and $\hat{y}_{V}$ represent the results with experience replay.
} The red line represents the baseline where the variance difference is $0$.}
\label{fig:kr}
\end{figure}

Figure \ref{fig:kr} shows the variance differences across test points by plotting the standard quartile breakdown boxplots of $\{\text{Var}(\tilde{y}_j )-\text{Var}(\hat{y}_{j,U} )\}^m_{j=1}$ and $\{\text{Var}(\tilde{y}_j )-\text{Var}(\hat{y}_{j,V} )\}^m_{j=1}$ with $B=50$,  $n\in \{100,150,200,250\}$, and $k\in\{10, 15, 20\}$. The results confirm that the variance reduction property holds across all settings for both $ U $- and $ V $-statistics-based experience replay methods. 
Appendix \ref{app:krls} includes additional experiments with different \( B \) values and evaluations on a real-world dataset from the U.S. Census Bureau on Boston housing, further demonstrating effectiveness.

\begin{table}[ht!]
\centering
\fontsize{9}{10}\selectfont
\renewcommand{\arraystretch}{0.8}  
{%
\begin{tabular}{lllllll}
\toprule 
& \multicolumn{2}{c}{\( k = 10 \)} & \multicolumn{2}{c}{\( k = 15 \)} & \multicolumn{2}{c}{\( k = 20 \)} \\ \cmidrule(l{0.75em}r{0.75em}){2-3}
\cmidrule(l{0.75em}r{0.75em}){4-5}
\cmidrule(l{0.75em}r{0.75em}){6-7}
\( n \) & \( {t} - {t}_U \) & \( {t} - {t}_V \) & \( {t} - {t}_U \) & \( {t} - {t}_V \) & \( {t} - {t}_U \) & \( {t} - {t}_V \) \\  \midrule
200  & 0.369 & 0.323 & 0.335 & 0.272 & 0.108 & 0.110 \\  
250  & 3.005 & 2.905 & 2.850 & 2.804 & 2.954 & 2.848 \\ \bottomrule 
\end{tabular}%
}
\caption{Time cost reduction achieved by experience replay methods (measured in seconds) with \( B=50 \) for different values of \( k \) and \( n \).}
\label{tab:1} 
\end{table}

Table \ref{tab:1} presents the time cost reduction achieved by the experience replay methods with $B=50$, $k \in \{10, 15, 20\}$, and $n \in \{200, 250\}$. 
Here, ${t}$ represents the total time cost across all experiments without experience replay, while ${t}_U$ and ${t}_V$ represent the total time costs with experience replay based on resampled $U$- and $V$-statistics, respectively. Time cost was measured as wall-clock time on a single core without parallelization on a laptop with an Apple M2 Pro and 16 GB of RAM.
The results demonstrate that, for a fixed $B$, the experience replay method reduces the computational cost in time, particularly when $k$ is small and $n$ is large.  We also compare the RMSE of the proposed methods with the original method in Appendix \ref{rmse simu}.
The results indicate that incorporating experience replay, regardless of the specific resampling method used, not only reduces variance and time cost but also decreases prediction errors for all settings, especially in data-scarce scenarios. 

While our theoretical results apply to both $U$- and $V$-statistics, empirical results show no major differences between them. In practice, $V$-statistics are often preferable due to their GPU-friendliness, ease of parallelization, and compatibility with modern machine learning frameworks.

\section{Conclusion}\label{sec:conclusion}
Experience replay improves stability and efficiency in reinforcement learning, but its theoretical properties are still underexplored. This paper presents a theoretical framework that models experience replay using resampled $U$- and $V$-statistics, enabling us to establish variance reduction guarantees across policy evaluation and supervised learning tasks.
We applied this framework to two policy evaluation algorithms—the LSTD method and a PDE-based model-free algorithm—demonstrating notable improvements in stability and accuracy, particularly in data-scarce settings. Additionally, we applied the framework to kernel ridge regression, achieving both significant computational savings and variance reduction. Future research could extend experience replay to federated and active learning settings. For example, using replay to improve communication efficiency and model personalization in federated learning, or selecting informative data subsets for replay in active learning, may address distributed data challenges.

\section*{Acknowledgments}

We would like to thank the area chair, senior program committee, and five anonymous referees for constructive suggestions that improve the paper. Dai's research was supported in part by NIH grant R01DK142026, a Merck Research
Award, and a Hellman Fellowship Award. Zhu's research was supported in part by NSF grant DMS-2529107 and a Hellman Fellowship Award.





{
\bibliography{subsample}
}

\newpage
\setcounter{secnumdepth}{2}
\appendix
\section*{Appendix}
\textcolor{black}{
Appendix \ref{extend} discusses the extension to the importance-sampling setting.
Appendix \ref{relax} presents a relaxation of the i.i.d. assumption employed in the paper.
Appendix \ref{dis_condition} offers additional discussion of the condition $\lim_{n \to \infty} k^2 \zeta_{1,k} > 0$.
Appendix \ref{insight} presents theoretical insights into experience replay for Q-Learning.
Appendix \ref{complexity} provides details of computational complexity analysis.}
Appendix \ref{app:proofs} provides the proofs of theoretical results in this paper. Appendix \ref{app:add exp} includes supplementary experiments that further demonstrate the variance reduction properties of the proposed $U$- and $V$-statistics-based experience replay methods. Appendix \ref{sec:appaddexp} presents numerical results comparing the RMSE of the proposed methods with the original method.

\section{Extension to Importance Sampling}\label{extend}
Our paper establishes theoretical guarantees under uniform sampling, which is the most commonly used and computationally efficient strategy \citep{zhang2017deeper}. However, our framework extends naturally to non-uniform (importance) sampling. In particular, the asymptotic normality of resampled
$U$- and $V$- statistics continue to hold under non-uniform sampling. 

Let indices $i$ be drawn with probability $P(i)$ and define importance weights $w_i = (\binom{n}{k}P(i))^{-1}$ or a self-normalized form. Based on standard results for weighted $U$- and $V$- statistics \citep[e.g.,][]{csorgHo2013asymptotics}, the only change to the asymptotic distribution is that the variance term $\zeta_{k,k}$ is replaced by $$\zeta_W = E[\text{Var}(w_i h(Z_{i_1},\dots,Z_{i_k}))|Z_1,\dots,Z_n].$$ For well-chosen proposals $P(i)$ (e.g., $P(i)\propto |h(Z_{i_1},\dots,Z_{i_k})|$ in Horvitz–Thompson form), we have $\zeta_W \le \zeta_{k,k}$, implying a smaller asymptotic variance. Hence, our main theorem extends directly to importance sampling and provides a theoretical variance-reduction guarantee for such replay-based methods.

\section{Relaxation of Independence and Identical Distribution (i.i.d.) Assumption}\label{relax}
This i.i.d. assumption has been adopted in numerous prior works in reinforcement learning to conduct theoretical analysis \citep[e.g.,][]{antos2008learning,fan2020theoretical} and also widely used in supervised learning tasks \citep[e.g.,][]{hainmueller2014kernel, wood2003thin}. In our case, this assumption is used to establish the asymptotic normality of the proposed estimators via the Central Limit Theorem (CLT). In fact, this assumption can be relaxed in two directions, and the results in the paper still hold.

Firstly, the requirement for independence across data points can be weakened under standard mixing or dependence conditions that still guarantee asymptotic normality. Specifically, the CLT holds if the sequence of data $ Z_1, \dots, Z_n $ satisfies one of the following:
\begin{itemize}
    \item It forms a stationary and ergodic Markov chain that is aperiodic, irreducible, and positive recurrent, with finite second moment $ \mathbb{E}[Z^2] < \infty $ \citep{meyn2012markov}.
    \item It forms a stationary, $\beta$-mixing process with summable coefficients, i.e., 
    $ \sum_{k=1}^\infty \beta(k) < \infty $, where  
     {\scriptsize\[ \beta(k) := \sup_t \mathbb{E}\left[ \sup_{A \in \sigma(Z_{t+k}, Z_{t+k+1}, \ldots)} \left| \mathbb{P}(A \mid Z_1, \ldots, Z_t) - \mathbb{P}(A) \right| \right],
     \]}
  and has finite second moment $ \mathbb{E}[Z^2] < \infty $
     \citep{bradley2005basic}.
     \item  It is an $ m $-dependent stationary process, i.e., random vectors $ (Z_1, \ldots, Z_i) $ and $ (Z_j, Z_{j+1}, \ldots) $ are independent whenever $ j - i > m $ \citep{hoeffding1948central}.
\end{itemize}
   In the RL literature, such mixing conditions are often satisfied when trajectories are collected sequentially in a Markovian environment under a fixed policy and have been used to analyze non-i.i.d. data (e.g., \citealp{thodoroff2018temporal}).

   Secondly, in the RL examples, the trajectory dataset defined in \eqref{D_n in rl} can comprise sequences of varying length, which our method naturally accommodates. A long trajectory can be decomposed into a sequence of shorter trajectories of fixed length without affecting the results. For instance, a trajectory of length $ L+1 $, denoted $ (s_0^l, s_1^l, \dots, s_L^l) $, can be partitioned into $ L $ sub-trajectories of length two: $(s_0^l, s_1^l), (s_1^l, s_2^l), \dots, (s_{L-1}^l, s_L^l).$ This decomposition does not affect the computation of our estimator in either the LSTD case (Eq. \ref{g and f}) or the first-order PDE-based algorithm (Eq. \ref{f and g 2} with $\alpha = 1$), as each term in the functions $g$ and $f$ depends on at most two consecutive states, and each sub-trajectory contributes separately to the empirical average. For the second-order PDE-based algorithm (Eq. \ref{f and g 2} with $\alpha = 2$), the same result can be obtained by partitioning each trajectory into $L - 1$ sub-trajectories of length three: $(s_0^l, s_1^l, s_2^l),\ (s_1^l, s_2^l, s_3^l),\ \dots,\ (s_{L-2}^l, s_{L-1}^l, s_L^l).$ By doing so, all trajectories are converted into a collection of uniformly short trajectories, enabling the consistent application of our theoretical framework.

\section{Further Discussion on the Condition $\lim_{n \to \infty} k^2 \zeta_{1,k} > 0$}\label{dis_condition}
The condition $ \lim_{n\to\infty}k^2\zeta_{1,k}>0$ holds for many base learners, where \[\zeta_{1,k} = \text{Var}(\mathbb{E}[h_k(Z_1, \dots, Z_k)|Z_1]) \] defined in Section \ref{sec:background}. For example, consider a neural network of the form $h_k(Z_1,\dots,Z_k)=\frac{1}{k}\sum_{i=1}^k\phi(Z_i),$ where $\phi$ is a shallow feedforward sub-network with bounded variance $\mathrm{Var}(\phi(Z)) = \sigma^2$. In this case, $\mathbb{E}[h_k(Z_1, \dots, Z_k)|Z_1] = \frac{1}{k}\phi(Z_1) +\frac{k-1}{k} \mathbb{E}[\phi(Z)],$ implying that $k^2\zeta_{1,k} = \sigma^2>0$. Other base learners, including trees and $k$-nearest neighbors, also satisfy this condition; see \citet{peng2019asymptotic}.

The class of algorithms considered in our paper of the form \[h_k(Z_1, \dots, Z_k) = [\sum_{i=1}^kg(Z_i)]^{-1}[\sum_{i=1}^kf(Z_i)]\] also satisfies this condition. Let $\mu_g=\mathbb{E}[g(Z)]$ and $\mu_f=\mathbb{E}[f(Z)]$. Applying the central limit theory and delta method, we obtain \[\mathbb{E}[h_k(Z_1, \dots, Z_k)|Z_1]=\mu_g^{-1} \mu_f+\frac{1}{k}H(Z_1)+o(k^{-1}),\] where \[H(Z_1)=\mu_g^{-1}(f(Z_1)-\mu_f)-\mu_g^{-1}(g(Z_1)-\mu_g) \mu_g^{-1}\mu_f.\] Hence, it holds that $\lim_{n \to \infty} k^2 \zeta_{1,k}=\text{Var}(H(Z_1)) > 0$.

\section{Theoretical Insights into Experience Replay for Q-Learning}\label{insight}
While prior work \citep{zhang2017deeper,fedus2020revisiting} primarily investigates experience replay heuristics empirically in the context of Q-learning by treating replay as a black box, our results provide theoretical insights that align with their findings—specifically, that both small and large replay buffer capacities $n$ can degrade performance under a fixed replay ratio $B$.
This aligns with our theoretical results in Theorem \ref{new_u_incomplete} and Theorem \ref{the_2}, which show that variance reduction is achieved only when $\lim_{n \to \infty} n / (Bk) \to 0$, implying that $n$ cannot be too large for a fixed $B$ in order for the replay mechanism to be effective. At the same time, $n$ cannot be too small, as a minimum buffer size is required to ensure the asymptotic variance reduction.

\section{Detailed Analysis of Computational Complexity}\label{complexity}

The replay-based methods can provide lower computational cost, even when $ Bk \gg n $, since only $ k $ samples are processed per iteration over $ B $ rounds. For example, in the case of standard kernel ridge regression, the cost of processing $ k $ points is $ O(k^3) $ \citep{wahba1990spline}, resulting in a total cost of $ O(Bk^3) $ across all iterations. In contrast, directly applying kernel ridge regression to all $ n $ data points incurs a cost of $ O(n^3) $. Notably, $ O(Bk^3) < O(n^3) $ can still hold even when $ Bk \gg n $; for instance, when $ B = n^{13/8} $ and $ k = n^{1/8} $, the experience replay-based method reduces the computational cost from the traditional  $O(n^3)$ in time to as low as $O(n^2)$ in time.

{Using feature mapping as an approximation of the kernel can also reduce the computational cost of standard kernel ridge regression. Specifically, when operating on $k$ data points and using a feature mapping with target dimension $q$, the computational complexity of evaluating an expression of the form $\left[\sum_{i=1}^k g(Z_i^*)\right]^{-1} \left[\sum_{i=1}^k f(Z_i^*)\right]$ is $O(kq^2 + q^3)$, where each $g(Z_i^*)$ is a $q \times q$ matrix and each $f(Z_i^*)$ is a $q$-dimensional vector. Over $B$ iterations, the total computational cost is therefore $O(Bkq^2 + Bq^3)$. 
Importantly, compared with the $ O(n^3) $ complexity of the standard kernel ridge regression \citep{wahba1990spline}, this replay-based method can offer both substantial computational savings and variance reduction when parameters are appropriately chosen, as demonstrated in Examples 1 and 2.  }

{\textit{Example} 1. Setting $B = O(n^{2\delta}),k=O(n^{1-\delta}),q = O\left(n^{\frac{2s}{2s+p}(1-\delta)}\log n\right)$, where $s\geq p/2$ is the smoothness parameter of the kernel $K$, and $\delta>0$ is a constant, satisfies the conditions of Theorems \ref{new_u_incomplete} and reduces the variance.
Under this setup, the computational cost is reduced to $O\left(n^{1+\frac{4s}{2s+p}}\cdot n^{\delta[1-\frac{4s}{2s+p}]}(\log n)^2\right)$ in time, offering significant savings compared to the standard $O(n^3)$ complexity of kernel ridge regression.  
Moreover, the method achieves a convergence rate  of $O\left(n^{-\frac{2s}{2s+p}(1-\delta)}\right)$ for the true function in the
RKHS  corresponding to kernel $K$ \citep[see,][]{rudi2017generalization, dai2024nonparametric}. This rate approaches the minimax optimal rate $O\left(n^{-\frac{2s}{2s+p}}\right)$ as $\delta\to 0$ \citep{wahba1990spline}. }

\textit{Example} 2.
Setting $B = O(n),k=O(n^{1/2}),q = O(n^{1/2})$, satisfies the conditions of Theorems \ref{new_u_incomplete} and reduces the variance. Under this setup, the computational cost is reduced to $O(n^{5/2})$ in time.


\section{Proofs of Main Results}\label{app:proofs}
\subsection{Proof of Lemma \ref{new_T1}}\label{A1.0}
\begin{proof}~
\noindent
We begin by applying the central limit theory 
to multivariate i.i.d. random variables to obtain:
   {\small{ \begin{equation}\label{new_11}
        \sqrt{n} \left[ \begin{pmatrix} \frac{1}{n}\sum_{j=1}^n f(Z_j) \\  \text{vec}(\frac{1}{n} \sum_{j=1}^n g(Z_j))\end{pmatrix} - \begin{pmatrix} \E[f(Z)] \\ \text{vec}(\E[g(Z)]) \end{pmatrix} \right]\overset{d}{\rightarrow} N(0, \Sigma_{0}),
    \end{equation}}}
    where the covariance matrix $\Sigma_{0}$ is defined as:
\begin{equation*}
    \Sigma_{0} = \begin{pmatrix}
    \text{Var}(f(Z)) &  \text{Cov}(f(Z),\text{vec}( g(Z))) \\
     \text{Cov}(f(Z),\text{vec}( g(Z))) & \text{Var}(\text{vec}( g(Z))
    \end{pmatrix}.
\end{equation*}
Next, we aim to find the asymptotic distribution of $\tilde{\theta}_n$. Consider the function $g(X,\text{vec}(Y))=Y^{-1}X$,  which is continuous for any $X\in\R^q$ and invertible matrix $Y\in\R^q\times \R^q$. Note that  
    {{\begin{equation*}
    \begin{split}
        &g\left(\frac{1}{n}\sum_{j=1}^n f(Z_j), \text{vec}\Big(\frac{1}{n} \sum_{j=1}^n g(Z_j)\Big)\right)\\
        =&\Big[\sum_{j=1}^n g(Z_j)\Big]^{-1}\Big[\sum_{j=1}^n f(Z_j)\Big]=\tilde{\theta}_n 
    \end{split}
     \end{equation*}}}
     and 
     \begin{equation*}
         g( \E[f(Z)],\text{vec}(\E[g(Z)]))=\big[\E[g(Z)]\big]^{-1}\big[\E[f(Z)]\big]=\theta.
     \end{equation*}
To apply the delta method, we define the Jacobian matrix $G$ as follows,
     \begin{equation*}
     \begin{split}
          G&:=\dot{g}(\E[f(Z)],\text{vec}(\E[g(Z)]))\\
          &=\left([\E[g(Z)]]^{-1},-\theta^\top\otimes[\E[g(Z)]]^{-1}  \right).
     \end{split}
     \end{equation*}
where $\otimes$ denotes the Kronecker product. Applying the delta method to \eqref{new_11}, we obtain:
    \begin{equation*}
\sqrt{n}\left[ \tilde{\theta}_n-\theta\right]
      \xrightarrow{d}N(0,G\Sigma_{0}[G]^\top)
      =N(0,\Sigma_{}),
    \end{equation*}
where $\Sigma $ is defined as:
\begin{equation}\label{n_Sigma}
    G\begin{pmatrix}

    \text{Var}(f(Z)) &  \text{Cov}(f(Z),\text{vec}( g(Z))) \\
     \text{Cov}(f(Z),\text{vec}( g(Z))) & \text{Var}(\text{vec}( g(Z))

    \end{pmatrix}G^\top.
\end{equation}

This completes the proof of Lemma \ref{new_T1}.
\end{proof}

\subsection{Proof of Theorem \ref{new_u_incomplete}}\label{A3.0}
\begin{proof}
From the definition of $\zeta_{1,k}$, we have that 
\begin{equation*}
    \zeta_{1,k}:= \text{Cov}\Big(h_k(Z_1,Z_2\dots,Z_{k}), h_k(Z_1,Z_2', \dots, Z_{k}^{'} )\Big),
\end{equation*}
where $Z_{2}^{'},\dots,Z_{k}^{'}$ are i.i.d. copies from  $F_Z$, independent of the original data set $\mathcal{D}_n$.

As a direct result of Lemma \ref{new_T1}, we can obtain the following corollary.
\begin{corollary}\label{new_c1}
Note that $h_{k}(Z_{1},\dots,Z_{{k}}) = \tilde{\theta}_k $,  where $h_k$ is defined in
\eqref{new_hkn}. Since $\zeta_{k,k}$ defined in Section \ref{sec:background} represents the variance of $h_{k}(Z_{1},\dots,Z_{{k}})$, 
we have     
\begin{equation}\label{new_40}
    \zeta_{k,k}=\frac{\Sigma_{}}{k}+o\Big(\frac{\Sigma_{}}{k}\Big),
\end{equation}
where $\Sigma_{}$ is defined in \eqref{n_Sigma}.
\end{corollary}

To analyze $\zeta_{1,k}$,  we use the following lemma.
\begin{lemma}\label{new_L1}
Let $ Z_1, Z_2, \ldots, Z_n \stackrel{\text{iid}}{\sim} F_Z $, with $ h_k $ defined in \eqref{new_hkn} and $ \Sigma $ defined in \eqref{n_Sigma}. Then,  $ k^2 \zeta_{1,k} < \Sigma_{} + o(\Sigma_{}) $.
\end{lemma}
\begin{proof}~
By \citet{lee2019u},  it follows that $\zeta_{1,k}< \frac{1}{k}\zeta_{k,k}$ when $k>1$. 
From Corollary \ref{new_c1}, we have that,
    \begin{equation*}
        \zeta_{k,k}=\Sigma_{}/k+o(\Sigma/k).
    \end{equation*}
Hence,
    \begin{equation*}
        k^2\zeta_{1,k} < k\zeta_{k,k}=  \Sigma_{}+o(\Sigma_{}).
    \end{equation*}
This completes the proof of Lemma \ref{new_L1}.
\end{proof}

Next, let $U_{n,k}$ denote the complete $U$-statistics with kernel $h_k$, defined as:
    \begin{equation*}
         U_{n,k}:=\frac{1}{\binom{n}{k}}\sum_{i}h_{k}(Z_{i_1},\dots, Z_{i_{k}}),
    \end{equation*}
where $\{Z_{i_1},\dots, Z_{i_{k}}\}$ represents a subsample of $k$ distinct elements from the original dataset $\mathcal{D}_n$, and the sum is taken over all $\binom{n}{k}$ possible subsamples of size $k$. The asymptotic normality of complete $U$-statistics has been studied in the literature  \citet{hoeffding1948class, peng2019asymptotic}. We extend these results to the case of matrix-valued kernels. Specifically, applying Theorem 1 in \citet{peng2019asymptotic} with a constant randomization term, if
\begin{equation*}
    \lim_{n\to \infty}\frac{1}{n}\zeta_{k,k}[\zeta_{1,k}]^{-1}\to 0,
\end{equation*} then it holds that
\begin{equation*}
   \sqrt{n} [{  U_{n,k}-\E h_{k}(Z_{1},\dots,Z_{{k}})}]\xrightarrow{d}  N(0,k^2\zeta_{1,k}).
\end{equation*}
From this result, we derive the asymptotic variance of $U_{n,k}$ as:
\begin{equation}\label{eq50}
    \text{Var}(U_{n,k})=\frac{k^2}{n}\zeta_{1,k}+o(\frac{k^2}{n}\zeta_{1,k}).
\end{equation}

For incomplete $ U $-statistics, \citet{blom1976some} established that the variance of an incomplete $ U $-statistic $ U_{n,k,B} $, constructed from $ B $ subsamples selected uniformly at random with replacement, is given by: 
\begin{equation}\label{eq51}
     \text{Var}(U_{n,k,B}) = \left( 1 - \frac{1}{B} \right) \text{Var}(U_{n,k}) + \frac{1}{B} \zeta_{k,k}.   
\end{equation}
This result holds even when both $ k $ and $ B $ vary with $ n $.

To finish the proof, we will use the following lemma.
\begin{lemma}\label{new_l1}
    $o(a_n)+o(B)=o(a_n+B)$, provided $a_n>0$ and $B>0$. 
\end{lemma}
\begin{proof}
  $|\frac{o(a_n)+o(B)}{a_n+B}|=|\frac{o(a_n)/a_n}{1+(B/a_n)}+\frac{o(B)/B}{(a_n/B)+1}|<|o(a_n)/a_n|+|o(B)/B|\to 0$. 
\end{proof}

Now we can prove the Theorem \ref{new_u_incomplete}. 
By \eqref{eq50}, \eqref{eq51}, \eqref{new_40}, Lemma \ref{new_L1}, and Lemma \ref{new_l1}, the variance of the estimator $\hat{\theta}_U$ can be expressed as:
{\small{\begin{equation}\label{new_49-1}
\begin{split}
    \text{Var}(\hat{\theta}_U)&=  \left( 1 - \frac{1}{B} \right)\left(\frac{k^2}{n}\zeta_{1,k}+o\Big(\frac{k^2}{n}\zeta_{1,k}\Big)\right) + \frac{1}{B} \left(\frac{\Sigma_{}}{k}+o\Big(\frac{\Sigma_{}}{k}\Big) \right)\\
    &<  \left( 1 - \frac{1}{B} \right)\left(\frac{\Sigma}{n}+o\Big(\frac{\Sigma}{n}\Big)\right) + \frac{1}{B} \left(\frac{\Sigma_{}}{k}+o\Big(\frac{\Sigma_{}}{k}\Big) \right)\\
    &=\frac{\Sigma}{n}\left(1+\frac{1}{B}\Big(\frac{n}{k}-1\Big)\right)+o\left(\frac{\Sigma}{n}+\frac{\Sigma}{Bk}\right).
\end{split}
\end{equation}}}
By Lemma \ref{new_T1}, we have $\Sigma_{}=n\text{Var}( \tilde{\theta}_n)+o(\Sigma_{})$. Substituting this into \eqref{new_49-1} and using 
 Lemma \ref{new_l1}, we have
{\small{\begin{equation*}
        \begin{split}
            \text{Var}(\hat{\theta}_U)&< \frac{1}{n}\left(1+\frac{1}{B}\Big(\frac{n}{k}-1\Big)\right)(n\text{Var}( \tilde{\theta}_n)+o(\Sigma_{}))+o\left(\frac{\Sigma}{n}+\frac{\Sigma}{Bk}\right)\\
            &=\left(1+\frac{1}{B}\Big(\frac{n}{k}-1\Big)\right)\text{Var}( \tilde{\theta}_n)+o\left(\frac{\Sigma_{}}{n}+\frac{\Sigma_{}}{Bk}\right).
        \end{split}
    \end{equation*}}}
Therefore, when $\lim_{n\to \infty}n/(Bk)\to 0$, it follows that $\text{Var}(\hat{\theta}_U)< \text{Var}( \tilde{\theta}_n)+o(1)$. In other words,
  \begin{equation*}
    \liminf_{n\to \infty}[\text{Var}( \tilde{\theta}_n) - \text{Var}(\hat{\theta}_U)] \geq 0.
\end{equation*}

This completes the proof of Theorem \ref{new_u_incomplete}.
\end{proof}

\subsection{Proof of Theorem \ref{the_2}}\label{A4.0}
\begin{proof}
We begin by extending Theorem 10 from \citet{zhou2021v} to the matrix-valued setting under the assumptions that $k=o(n^{1/4})$,  $h_{k}\in \mathcal{H}$, and $\lim_{n\to\infty} k^2\zeta_{1,k}>0$.
It follows that the estimator $\hat{\theta}_V$ satisfies,
\begin{equation}\label{new_v_as}
    { \hat{\theta}_V-\E h_{k}(Z_{1},\dots,Z_{{k}})}{}\xrightarrow{d} N\left(0,{\frac{k^2}{n}\zeta_{1,k}+\frac{1}{B}\zeta_{k,k}}\right).
\end{equation}
Next, we show that the assumption
\begin{equation}
\label{eqn:asspzetakkn}
    \lim_{n\to \infty}\frac{1}{n}\zeta_{k,k}[ \zeta_{1,k}]^{-1} \to 0
\end{equation}
in Theorem 10 of \citet{zhou2021v} is redundant. To see this, consider:
\begin{equation*}
    \frac{1}{n}\zeta_{k,k}[ \zeta_{1,k}]^{-1} 
    = \frac{k^2}{n}\frac{\zeta_{k,k}}{ k^2\zeta_{1,k}}.
\end{equation*}
Since $\lim_{n\to \infty} k^2\zeta_{1,k} > 0$, $\lim_{n\to \infty}\zeta_{k,k} > 0$  (by the assumption that $h_{k} \in \mathcal{H}$), and because $k = o(n^{1/4})$ implies $k^2/n \to 0$, it follows that:
\begin{equation*}
    \lim_{n\to \infty} \frac{1}{n}\zeta_{k,k}[ \zeta_{1,k}]^{-1} = 0.
\end{equation*}
Thus, we do not require the assumption \eqref{eqn:asspzetakkn} for the proof.

From equation \eqref{new_v_as}, we have:
\begin{equation*}
\begin{split}
    \text{Var}(\hat{\theta}_V)&=\frac{k^2}{n}\zeta_{1,k}+\frac{1}{B}\zeta_{k,k}+o\left(\frac{k^2}{n}\zeta_{1,k}+\frac{1}{B}\zeta_{k,k}\right).
\end{split}
\end{equation*}
Using Corollary \ref{new_c1}, Lemma \ref{new_L1}, and Lemma \ref{new_l1}, we derive the following inequality,
\begin{equation}\label{new_49}
\begin{split}
    \text{Var}(\hat{\theta}_V)&< \frac{\Sigma_{}}{n}+o\Big(\frac{\Sigma_{}}{n}\Big)+\frac{1}{B}\left(\frac{\Sigma_{}}{k}+o\Big(\frac{\Sigma_{}}{k}\Big)\right)+o\left(\frac{\Sigma_{}}{n}+\frac{\Sigma_{}}{Bk}\right)\\
    &=\left(\frac{1}{n}+\frac{1}{Bk}\right)\Sigma_{}+o\left(\frac{\Sigma_{}}{n}+\frac{\Sigma_{}}{Bk}\right).
\end{split}
\end{equation}
From Lemma \ref{new_T1}, we know that $\Sigma_{}=n\text{Var}( \tilde{\theta}_n)+o(\Sigma_{})$.
Substituting this into equation \eqref{new_49}, and by Lemma \ref{new_l1}, we obtain,
    \begin{equation*}
        \begin{split}
            \text{Var}(\hat{\theta}_V)&<\left(\frac{1}{n}+\frac{1}{Bk}\right)(n\text{Var}( \tilde{\theta}_n)+o(\Sigma_{}))+o\left(\frac{\Sigma_{}}{n}+\frac{\Sigma_{}}{Bk}\right)\\
            &=\left(1+\frac{n}{Bk}\right)\text{Var}( \tilde{\theta}_n)+o\left(\frac{\Sigma_{}}{n}+\frac{\Sigma_{}}{Bk}\right).
        \end{split}
    \end{equation*}
Therefore, when $\lim_{n\to \infty}n/(Bk)\to 0$, it follows that $     \text{Var}(\hat{\theta}_V)< \text{Var}( \tilde{\theta}_n)+o(1)$. In other words,  
  \begin{equation*}
    \liminf_{n\to \infty}[\text{Var}( \tilde{\theta}_n) - \text{Var}(\hat{\theta}_V)] \geq 0.
  \end{equation*}

  This completes the proof of Theorem \ref{the_2}.
\end{proof}

\section{Additional Numerical Experiments on Variance Reduction}\label{app:add exp}
We conducted all experiments on a personal laptop equipped with an Apple M2 Pro and 16GB of memory. All experiments were completed within three hours.

\subsection{Reinforcement Leaning Policy Evaluation}

\subsubsection{Additional Experiments Using LSTD Algorithm}\label{app:lstd}

\begin{figure}[htb]
\centering
\includegraphics[width=0.15\textwidth]{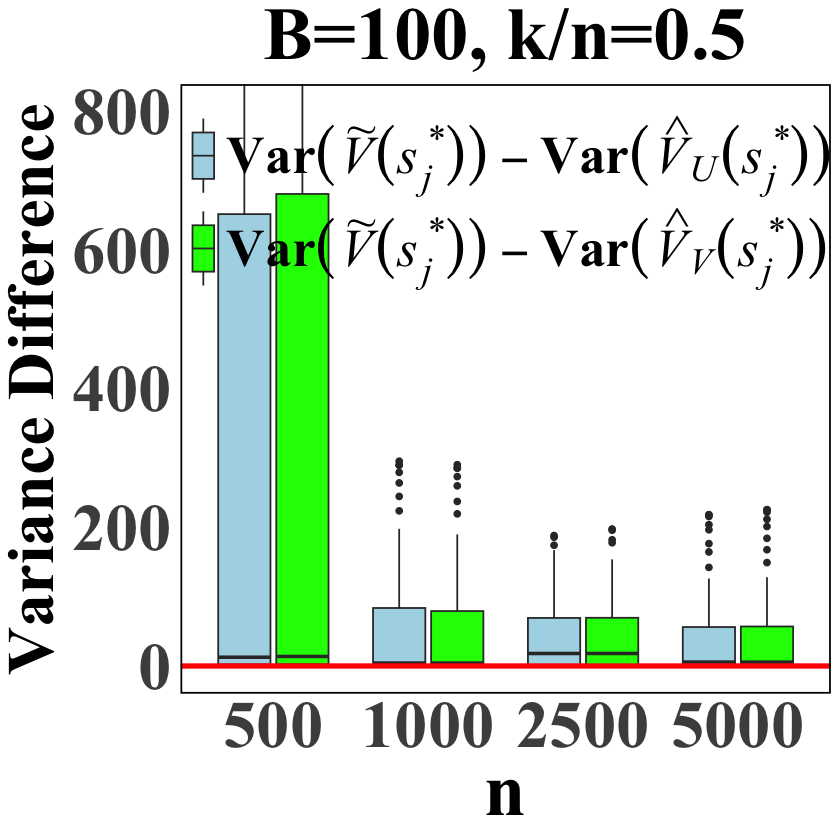} 
\includegraphics[width=0.15\textwidth]{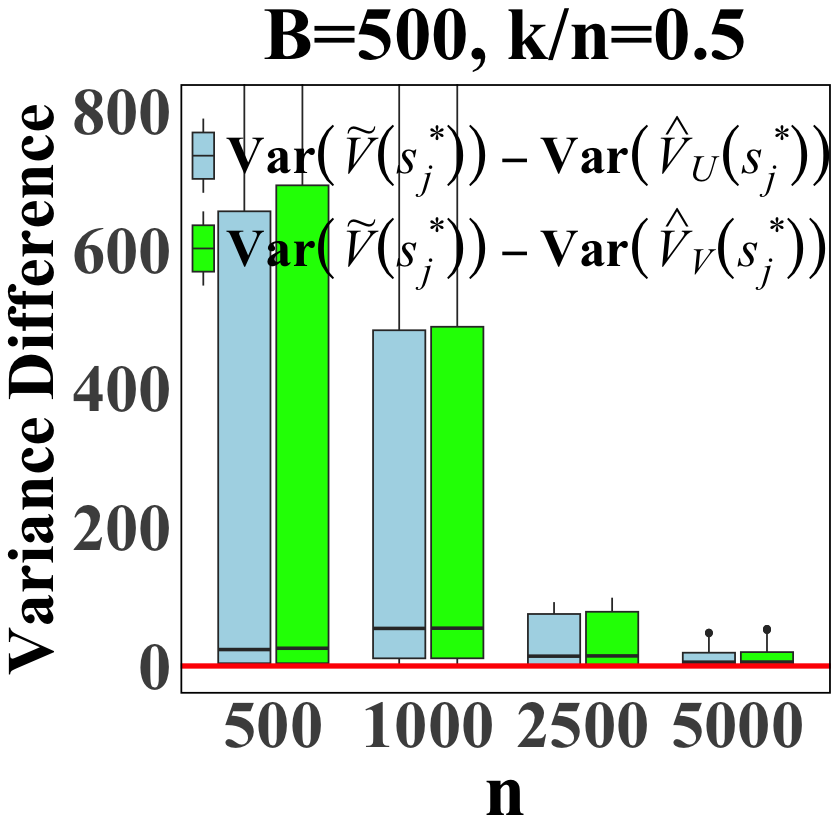} 
\includegraphics[width=0.15\textwidth]{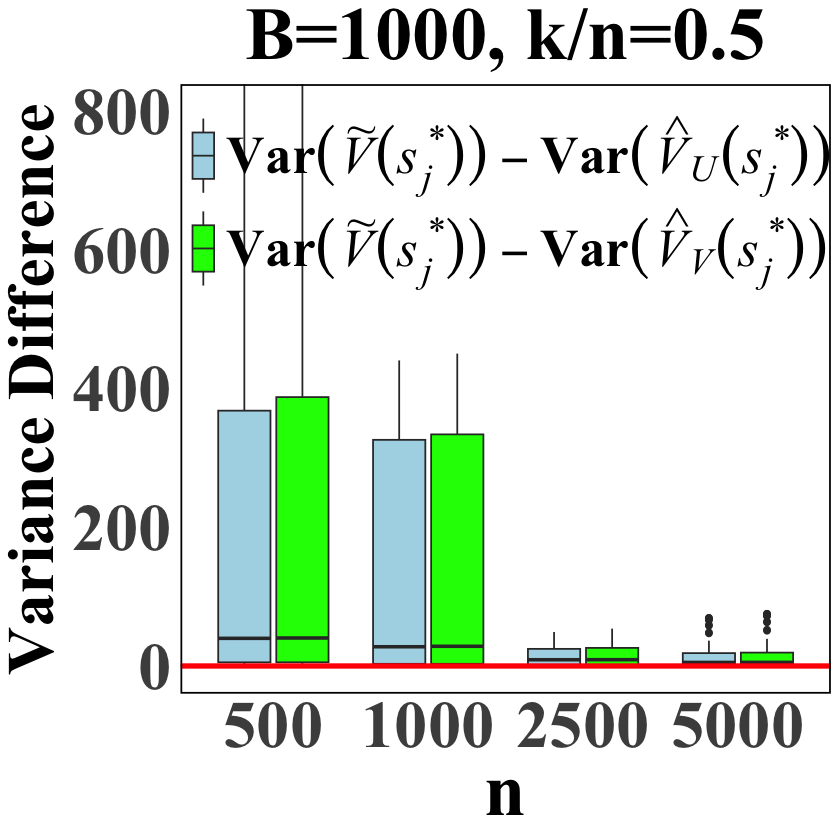} \\
\includegraphics[width=0.15\textwidth]{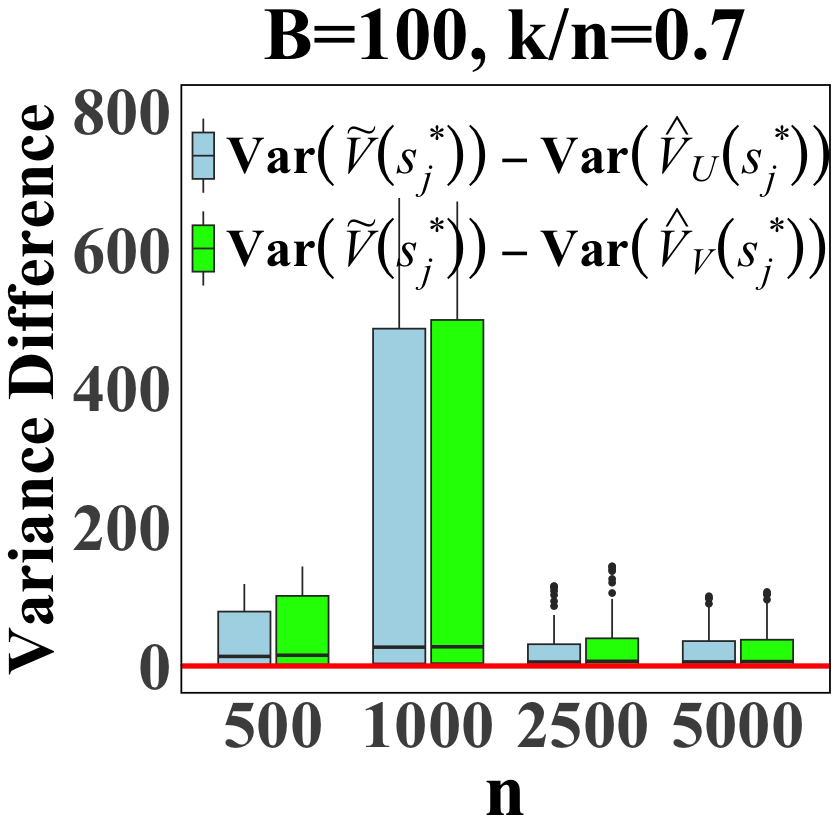} 
\includegraphics[width=0.15\textwidth]{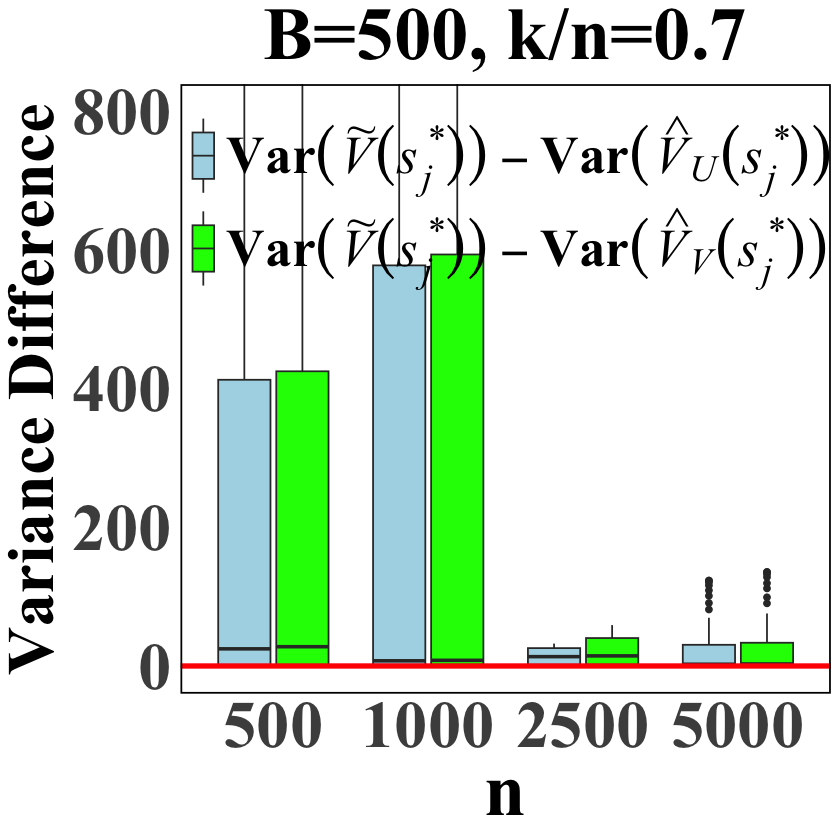} 
\includegraphics[width=0.15\textwidth]{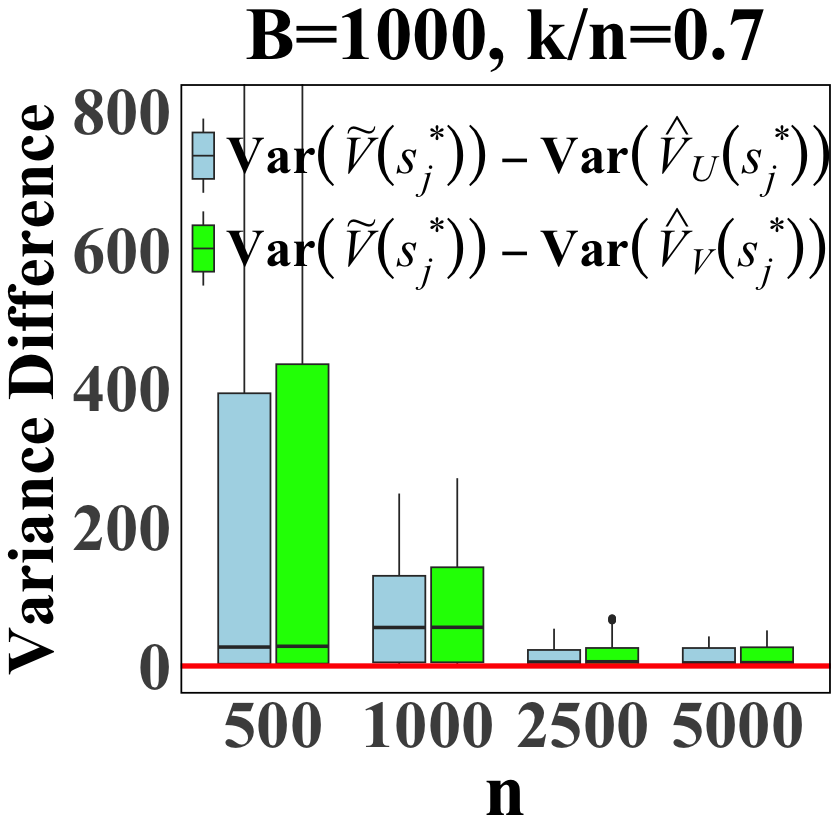} 

\caption{ Variance differences among the predicted policy values using the LSTD algorithm with $ m = 50 $ and $ M = 50 $, evaluated across various values of $ n $, $ B $, and $ k/n $. 
\textcolor{black}{$\tilde{V}(s^*_j)$ represents the results without experience replay, while $\hat{V}_{U}(s^*_j)$ and $\hat{V}_{V}(s^*_j)$ represent the results with experience replay.}
The red line represents the baseline where the variance difference is 0.}
\label{app:RL_LSTD}
\end{figure}

As a supplement to Figure \ref{fig:RL_LSTD} in Section \ref{exp:PE}, Figure \ref{app:RL_LSTD} compares the variances by presenting boxplots of the differences \(\{\text{Var}(\tilde{V}(s_j^{*})) - \text{Var}(\hat{V}_U(s_j^{*}))\}_{j=1}^m\) and \(\{\text{Var}(\tilde{V}(s_j^{*})) - \text{Var}(\hat{V}_V(s_j^{*}))\}_{j=1}^m\), with \( n \in \{500, 1000, 2500, 5000\} \) and \( k/n \in \{0.5, 0.7\} \). The results clearly show that, for all parameter settings, the variance differences across all test data points remain consistently greater than 0 for both resampling methods, further validating the effectiveness and robustness of our approach.

\subsubsection{Additional Experiments Using Second-Order PDE-Based Algorithm}\label{app:PE 2nd}

\begin{figure}[ht]
\centering
\includegraphics[width=0.15\textwidth]{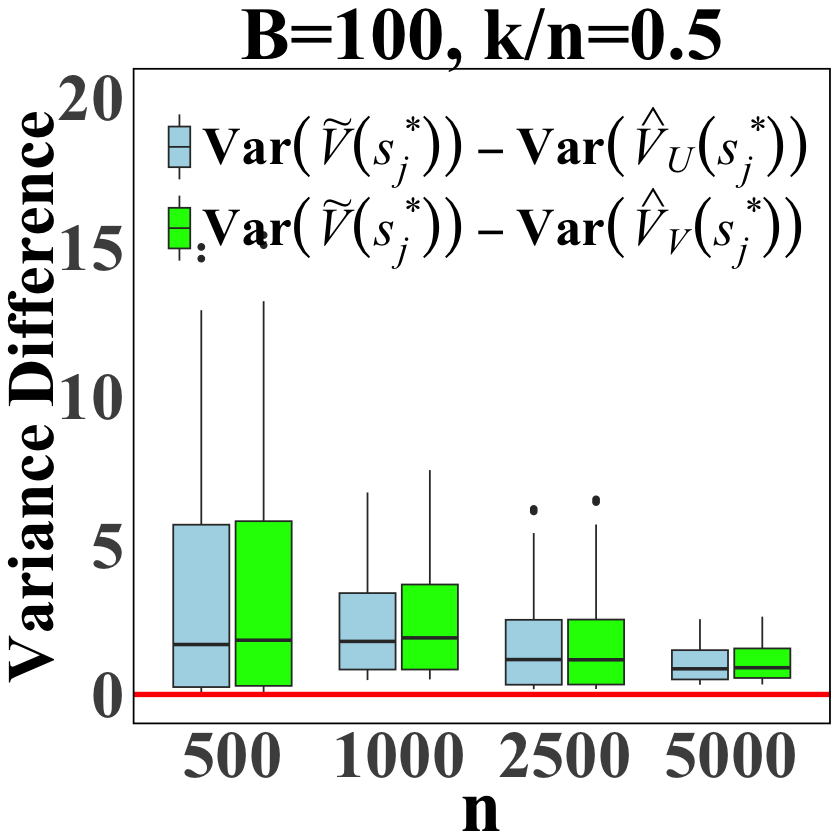} 
\includegraphics[width=0.15\textwidth]{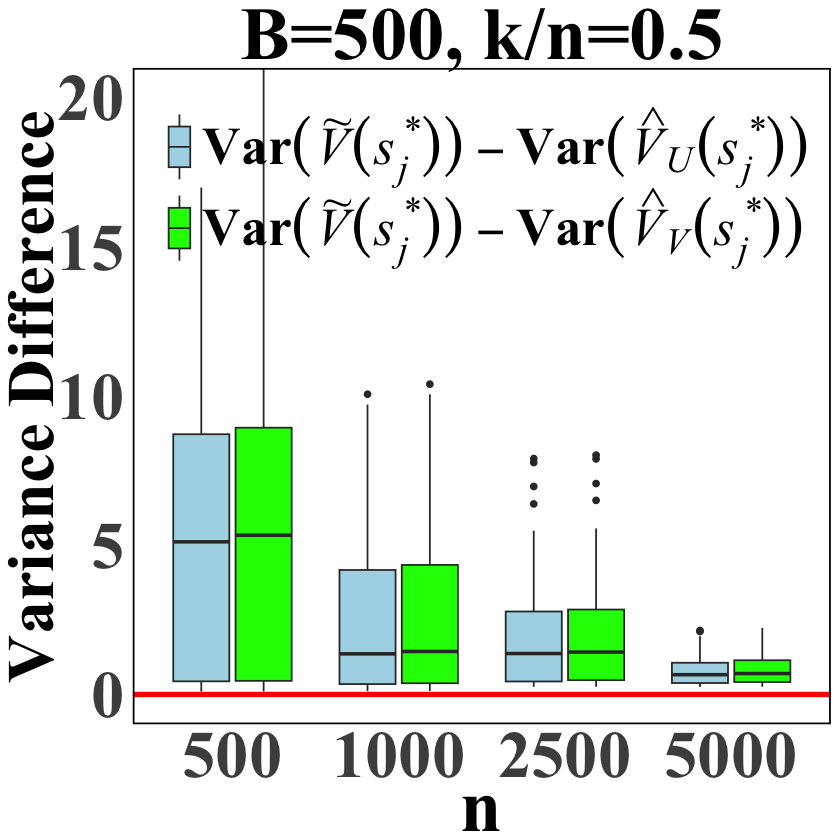} 
\includegraphics[width=0.15\textwidth]{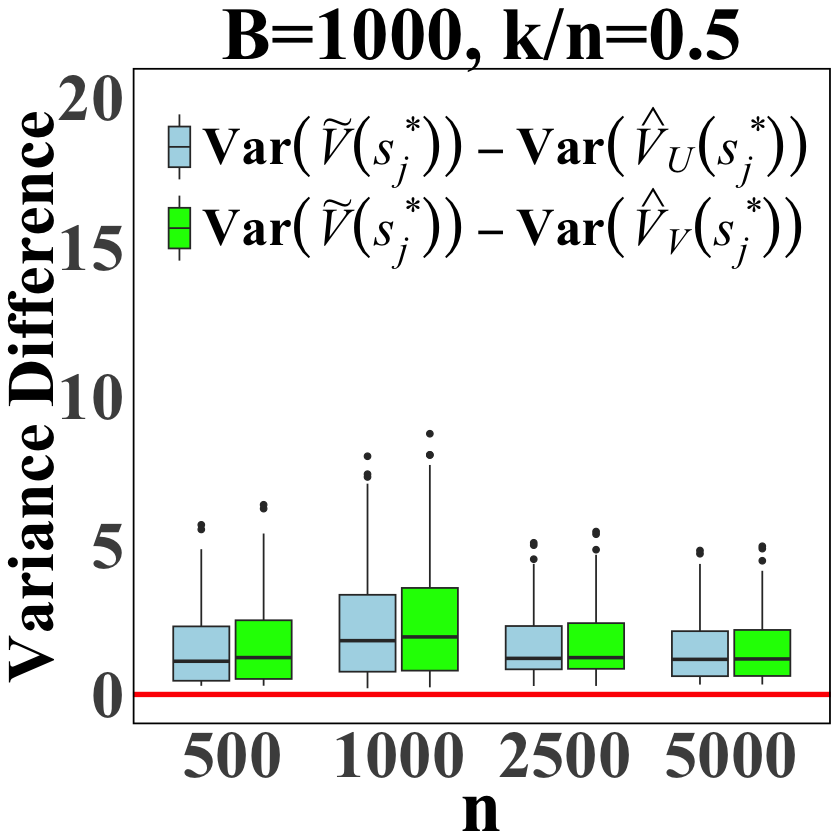} \\
\includegraphics[width=0.15\textwidth]{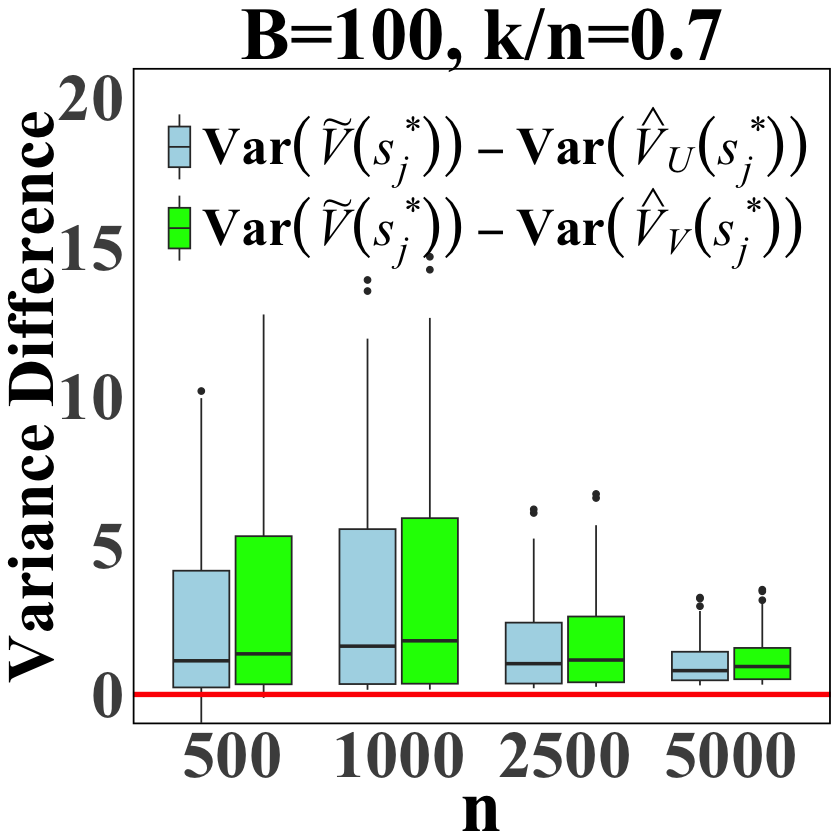} 
\includegraphics[width=0.15\textwidth]{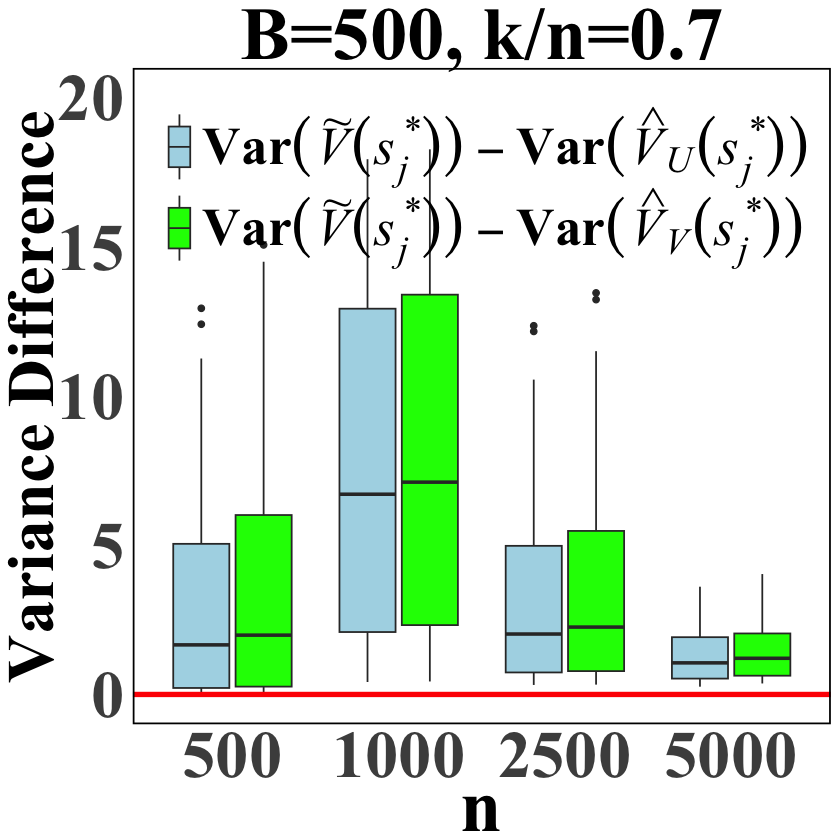} 
\includegraphics[width=0.15\textwidth]{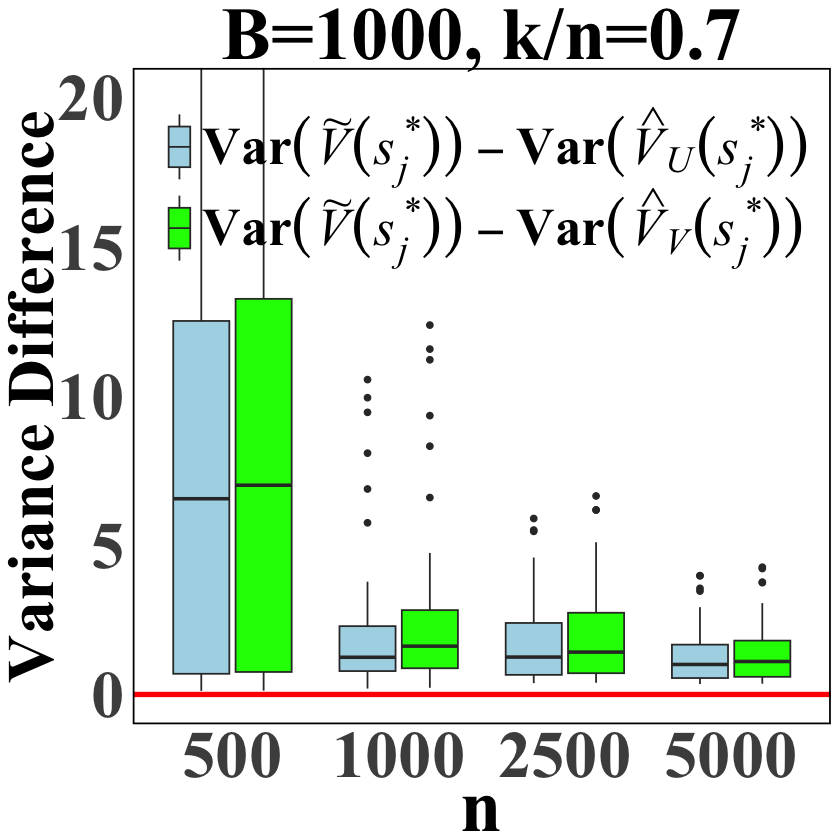} 
\caption{ Variance differences among the predicted policy values using the second-order PDE-based algorithm with $ m = 50 $ and $ M = 50 $, evaluated across various values of $ n $, $ B $, and $ k/n $. \textcolor{black}{$\tilde{V}(s^*_j)$ represents the results without experience replay, while $\hat{V}_{U}(s^*_j)$ and $\hat{V}_{V}(s^*_j)$ represent the results with experience replay.} The red line represents the baseline where the variance difference is 0.}
\label{app:RL_2nd}
\end{figure}

Figure \ref{app:RL_2nd} complements Figure \ref{fig:RL_2nd} in Section \ref{PDE:2nd} by providing boxplots of the variance differences, \(\{\text{Var}(\tilde{V}(s_j^{*})) - \text{Var}(\hat{V}_U(s_j^{*}))\}_{j=1}^m\) and \(\{\text{Var}(\tilde{V}(s_j^{*})) - \text{Var}(\hat{V}_V(s_j^{*}))\}_{j=1}^m\), under \( n \in \{500, 1000, 2500, 5000\} \) and \( k/n \in \{0.5, 0.7\} \). These results demonstrate that, across all parameter configurations, the variance differences for both resampling methods consistently remain positive for all test data points, further confirming the robustness and efficiency of our approach.

\subsubsection{Experiments Using First-Order PDE-Based Algorithm}\label{app:PE 1st}

We use the same experiment setting in Section \ref{PDE:2nd} and use the first-order PDE-based approach in the continuous-time case with functions $g$ and $f$ defined in \eqref{f and g 2} with $\alpha=1$.

\begin{figure}[htb]
\centering
\includegraphics[width=0.15\textwidth]{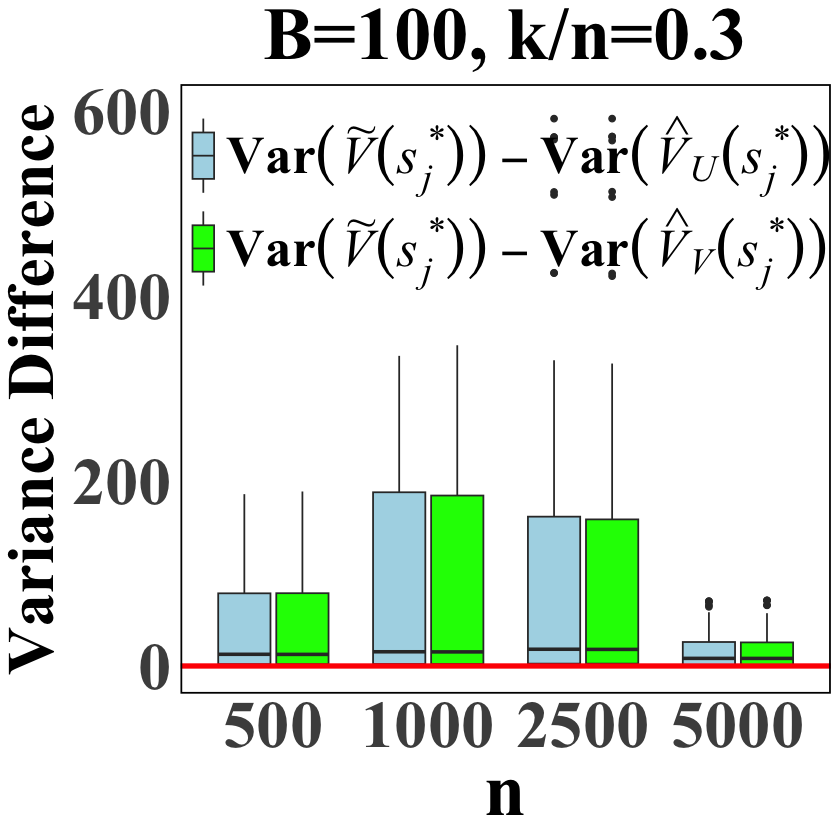} 
\includegraphics[width=0.15\textwidth]{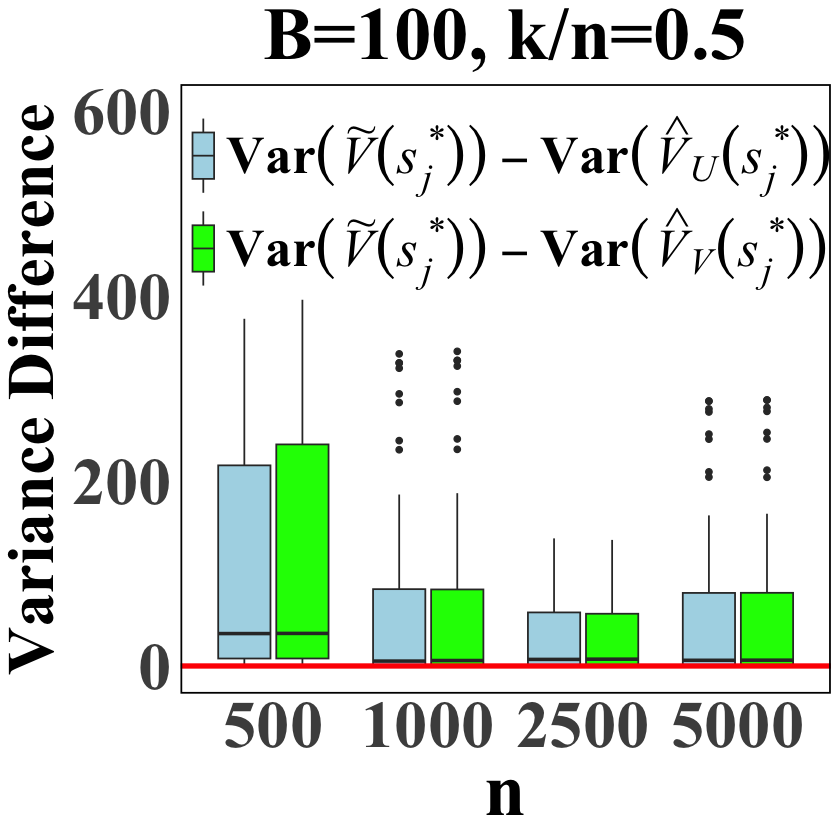} 
\includegraphics[width=0.15\textwidth]{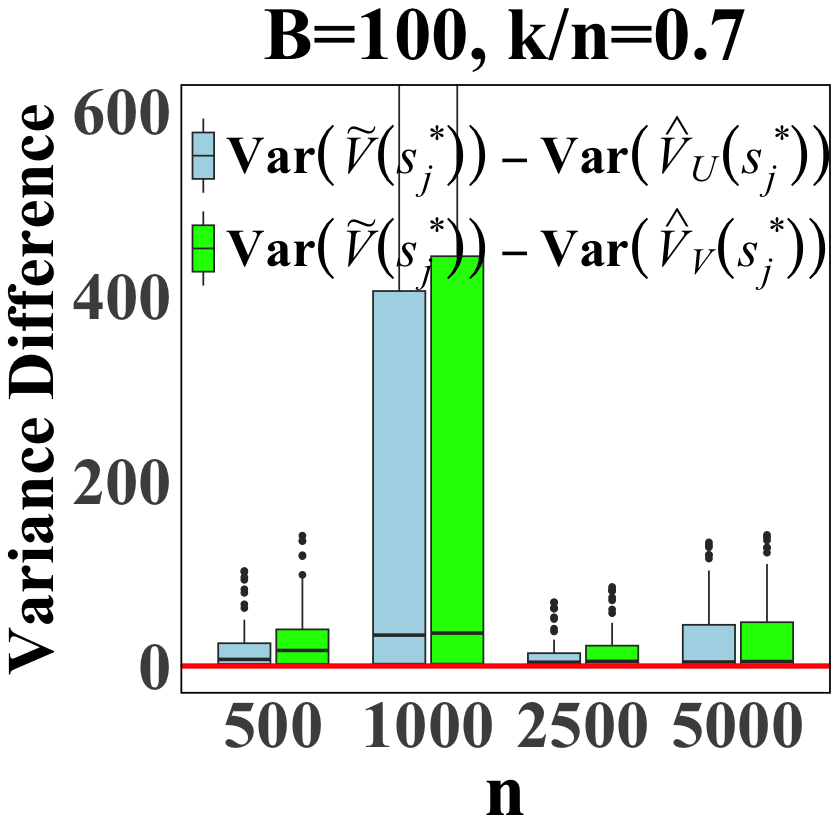} \\
\includegraphics[width=0.15\textwidth]{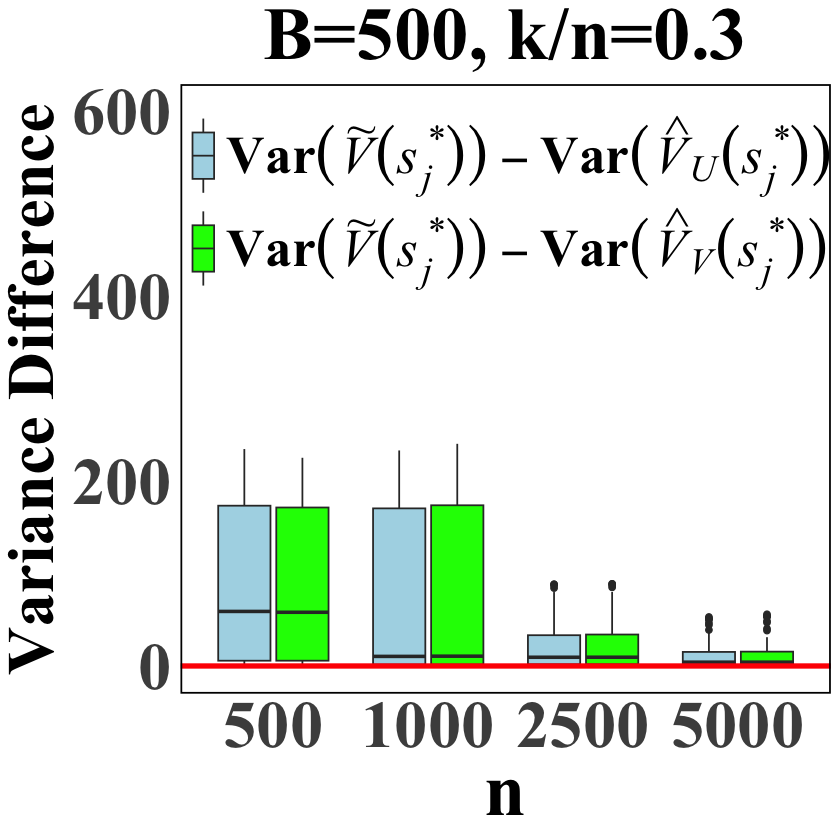} 
\includegraphics[width=0.15\textwidth]{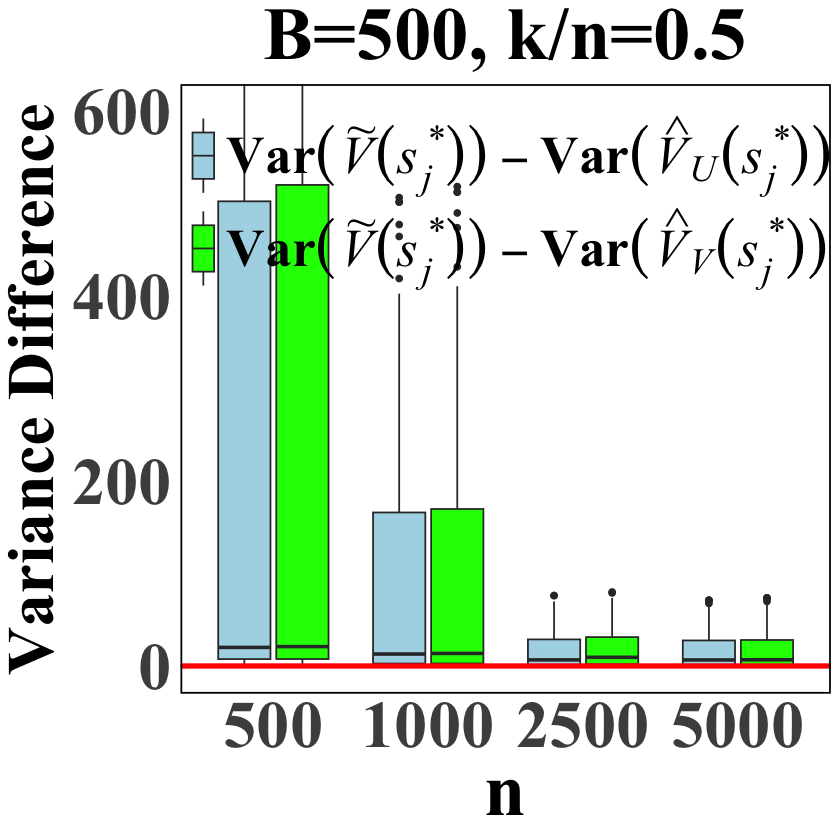} 
\includegraphics[width=0.15\textwidth]{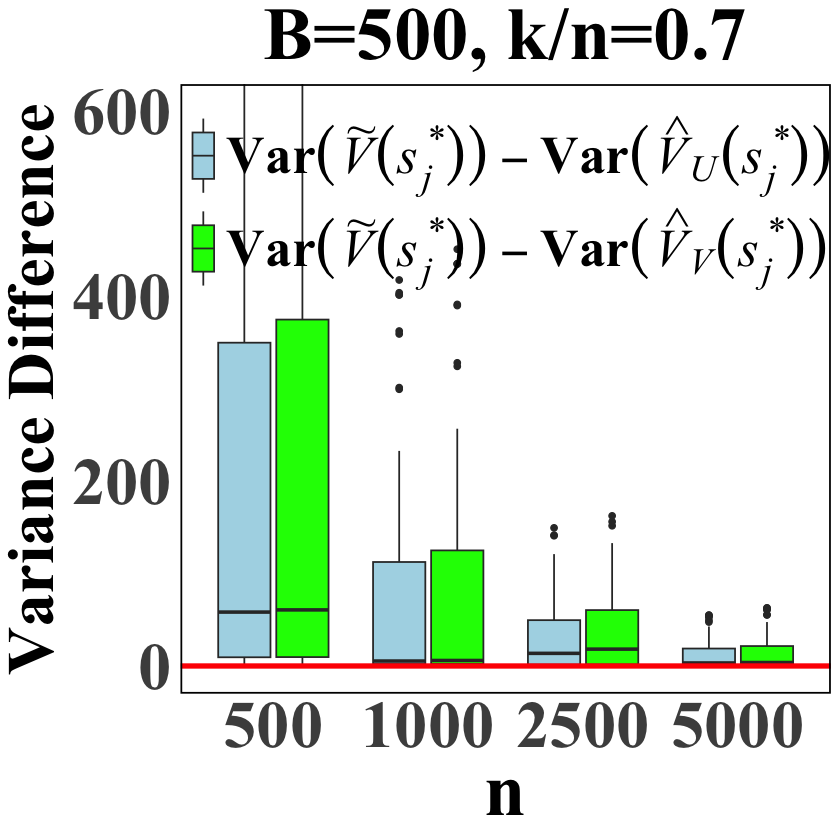} \\
\includegraphics[width=0.15\textwidth]{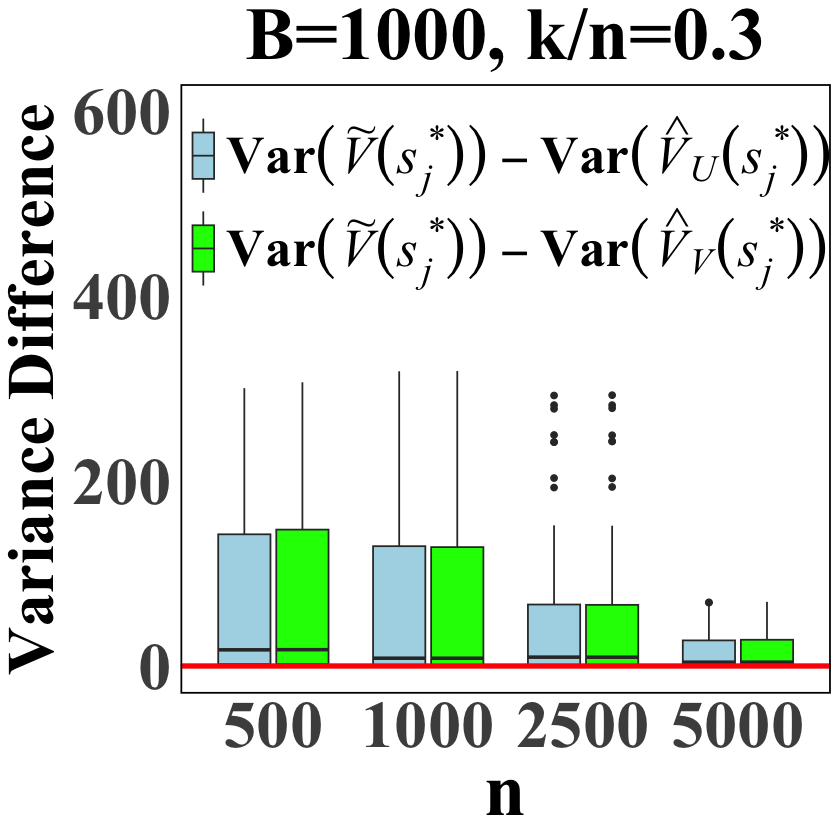} 
\includegraphics[width=0.15\textwidth]{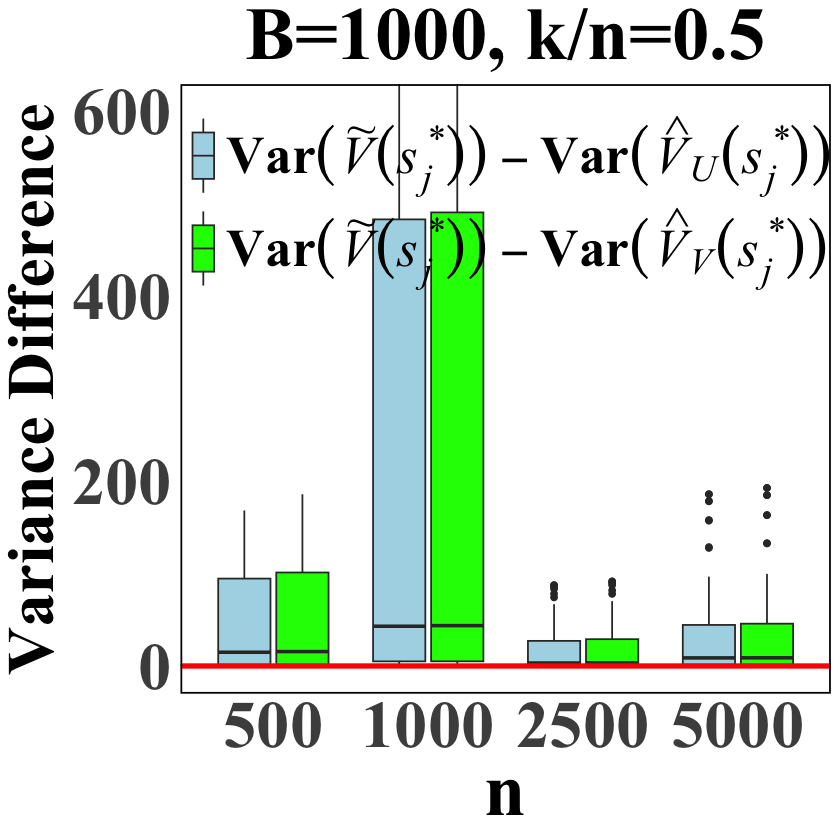} 
\includegraphics[width=0.15\textwidth]{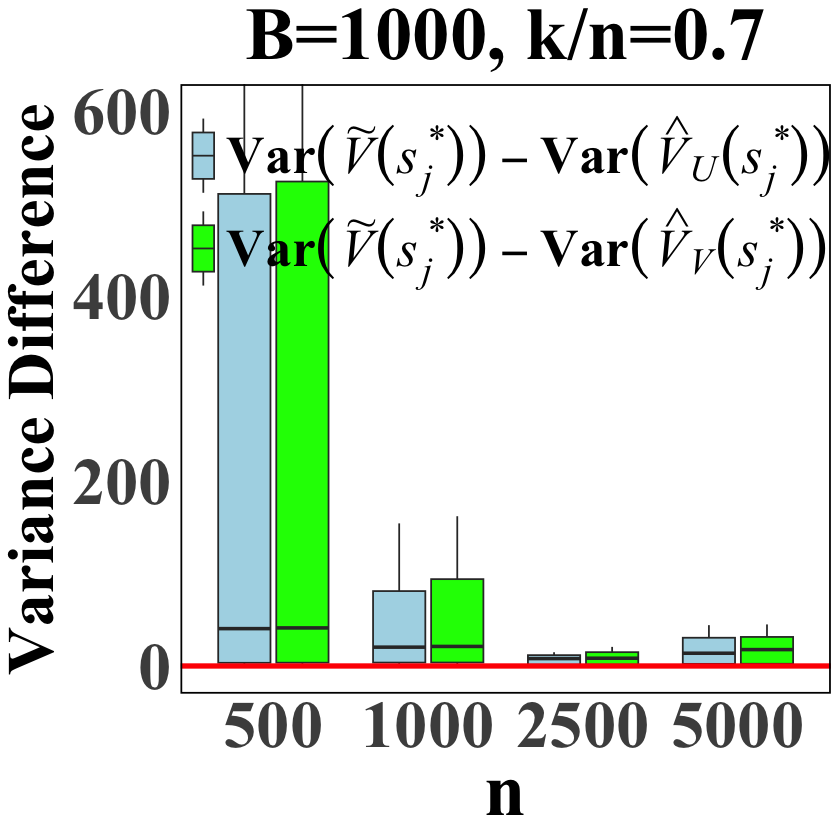} 
\caption{ Variance differences among the predicted policy values using the first-order PDE-based method with $ m = 50 $ and $ M = 50 $, evaluated across various values of $ n $, $ B $, and $ k/n $. \textcolor{black}{$\tilde{V}(s^*_j)$ represents the results without experience replay, while $\hat{V}_{U}(s^*_j)$ and $\hat{V}_{V}(s^*_j)$ represent the results with experience replay.} The red line represents the baseline where the variance difference is 0.}
\label{fig:RL_1ST}
\end{figure}

\begin{figure}[htbp]
    \centering
    \includegraphics[width=0.23\textwidth]{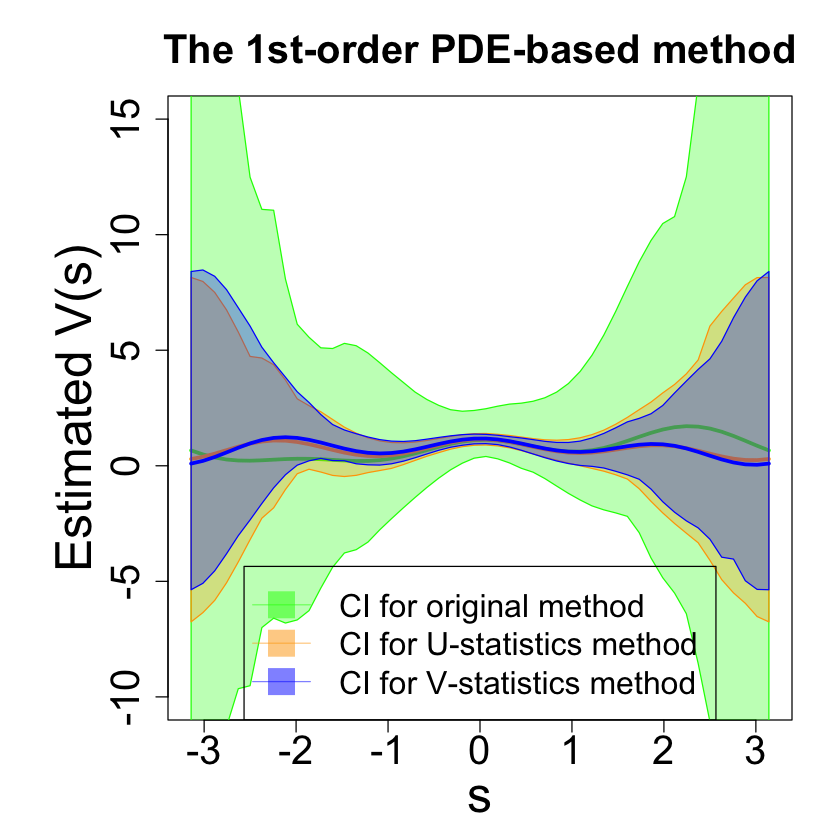} 
    \caption{ Variance reduction achieved by experience replay in policy evaluation using the first-order PDE-based approach. The solid lines represent the mean, and the shaded areas show $95\%$ confidence intervals based on 50 replications.}
    \label{fig:figure2}
\end{figure}

Figure \ref{fig:RL_1ST} compares the variances by drawing the standard quartile breakdown boxplots of the differences $\{\text{Var}(\tilde{V}(s_j^{*}))-\text{Var}(\hat{V}_U(s_j^{*}))\}_{j=1}^m$ and $\{\text{Var}(\tilde{V}(s_j^{*}))-\text{Var}(\hat{V}_V(s_j^{*}))\}_{j=1}^m$, with regard to different $n$, $B$, and the ratio $k/n$. We choose the $n\in \{500, 1000,2500,5000\}$, $B\in\{100, 500, 1000\}$, and $k/n\in\{0.3, 0.5, 0.7\}$. The results clearly demonstrate that for all of the different parameters, the variance differences across all test data points are consistently greater than 0 for both $ U $- and $ V $-statistics-based experience replay methods. 
As $ n $ increases, the variance differences tend to diminish because all three methods exhibit reduced variance, resulting in correspondingly smaller differences; however, the reduction in variance remains significant. To illustrate this, we consider the case where $ n = 5000 $, $ B = 500 $, and $ k/n = 0.5 $, and draw the Figure \ref{fig:figure2} similar to the Figure \ref{fig:three_figures}. From Figure \ref{fig:figure2}, we observe that the resampled methods demonstrate a significant improvement in variance in this larger $ n $ scenario.
With the use of experience
replay, the second-order method achieves a greater
percentage reduction in variance compared to the LSTD and the
first-order PDE-based method. Intuitively, the second-order
method accounts for two future steps, introducing more
stochasticity, which provides greater potential for variance
reduction.

We compare the RMSE of the proposed methods
with the original method across the $ m $ test points for all $ M $ experiments. The detailed results are presented in Appendix \ref{rmse 1st}, demonstrating that the combination of experience replay, regardless of the specific resampling method used, not only reduces variance but also tends to achieve smaller prediction errors, further highlighting its superiority.


\subsection{Kernel Ridge Regression}

\subsubsection{Additional Simulation Experiments}\label{app:krls}

\begin{figure}[ht]
\centering
\includegraphics[width=0.15\textwidth]{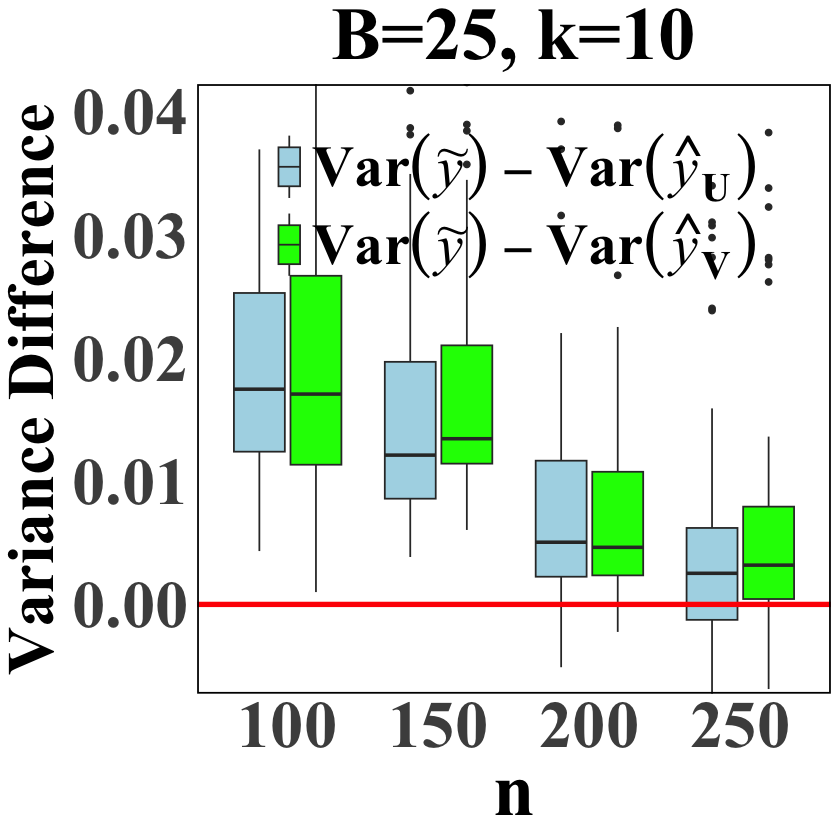} 
\includegraphics[width=0.15\textwidth]{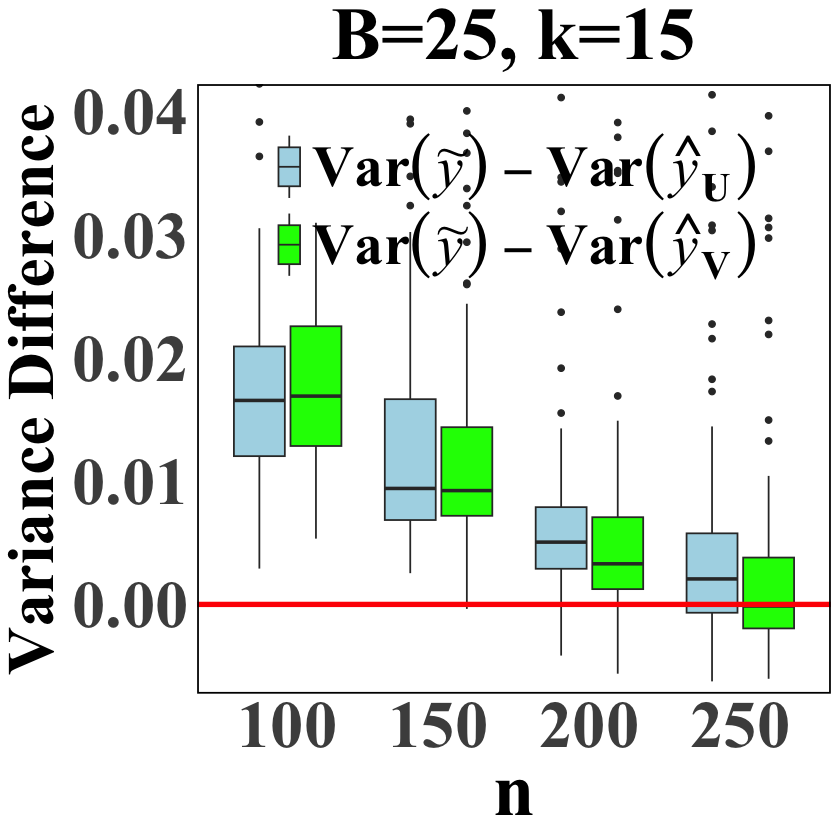} 
\includegraphics[width=0.15\textwidth]{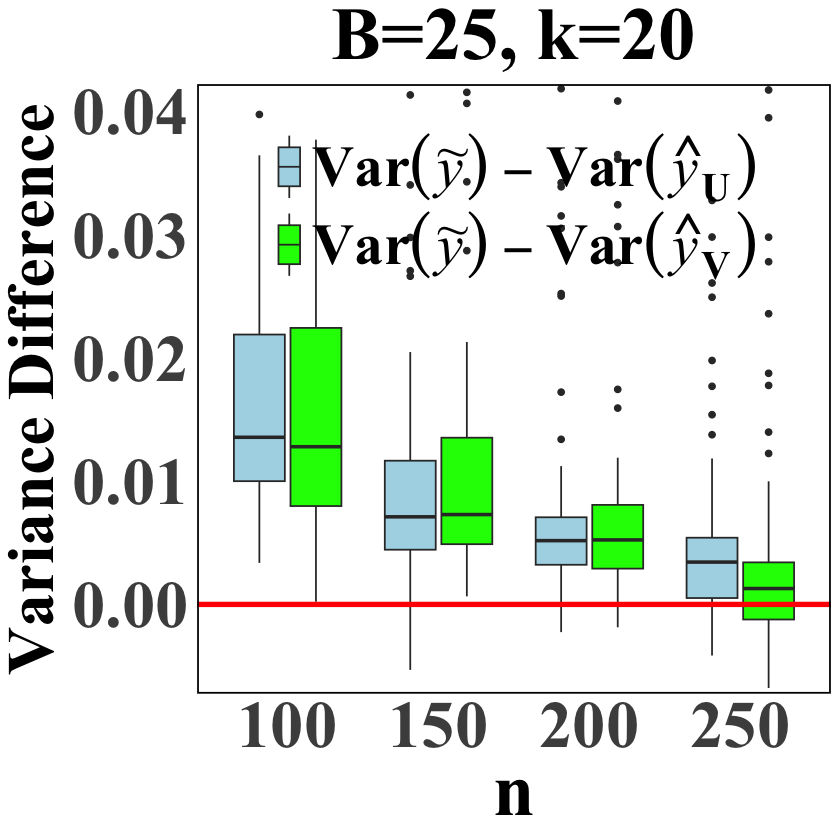} \\
\includegraphics[width=0.15\textwidth]{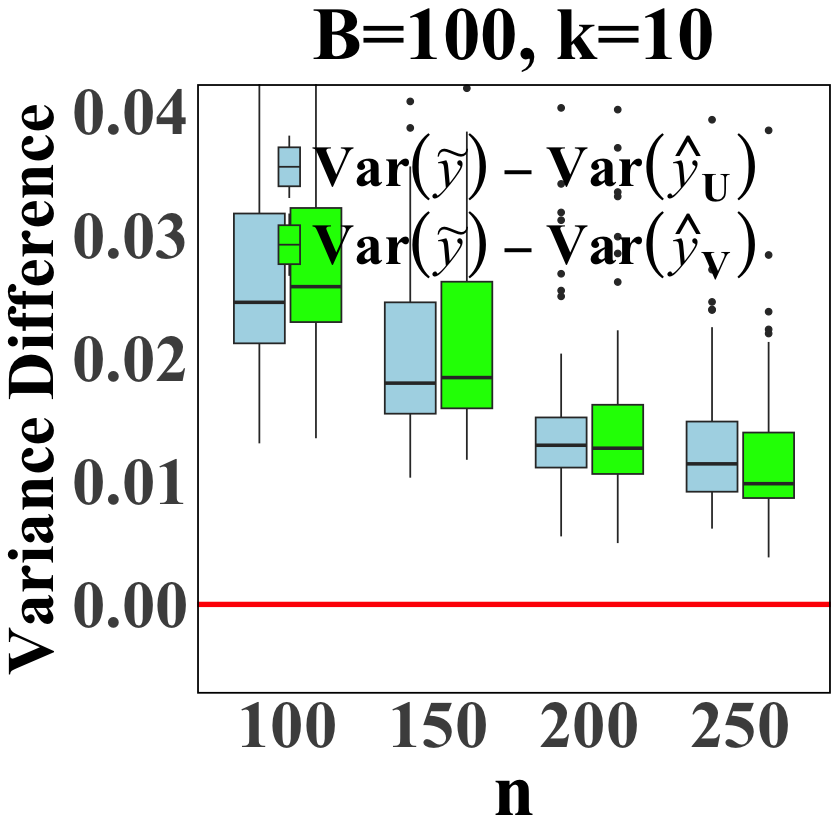} 
\includegraphics[width=0.15\textwidth]{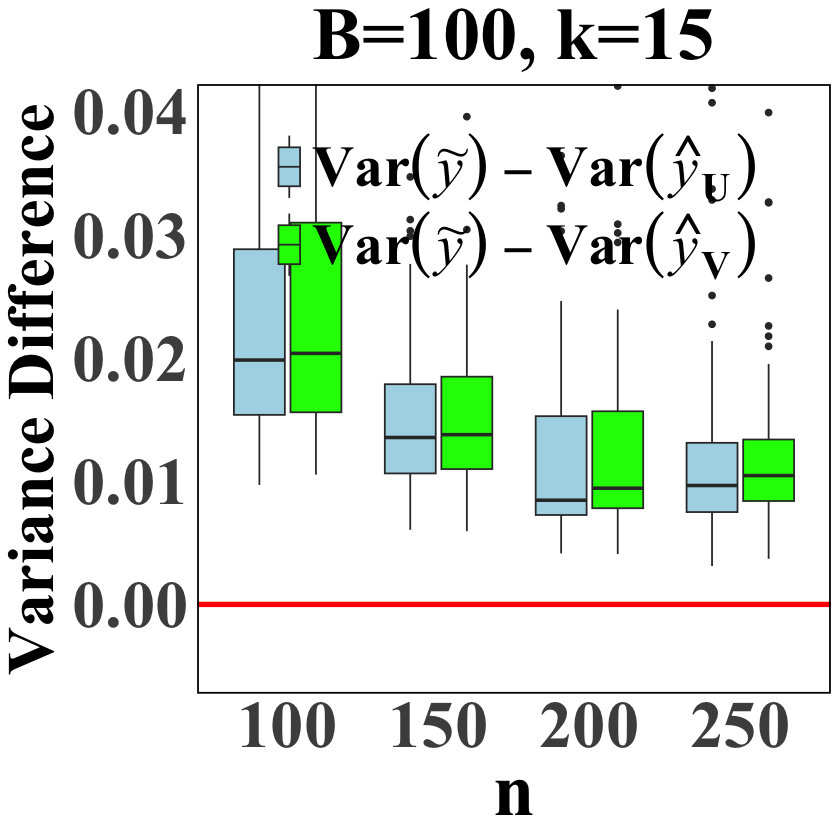} 
\includegraphics[width=0.15\textwidth]{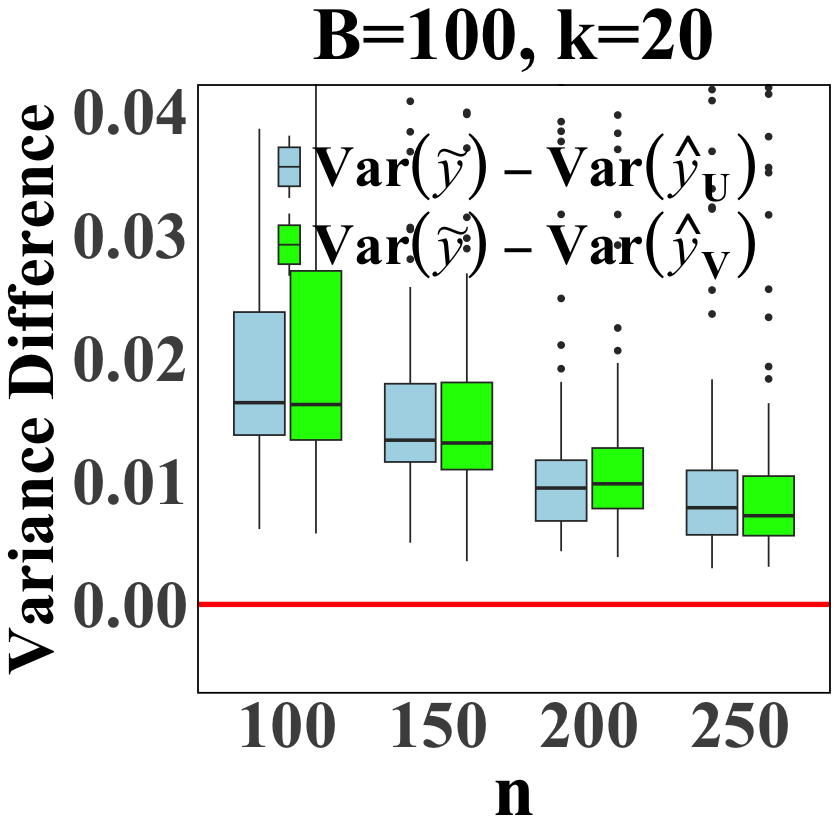} 
\caption{ Variance differences in predicted outcomes using kernel ridge regression on the simulated data with $B=50, m= 100 $ and $ M = 100 $, evaluated across various values of  $ n $ and $ k$. \textcolor{black}{$\tilde{y}$ represents the results without experience replay, while $\hat{y}_{U}$ and $\hat{y}_{V}$ represent the results with experience replay.
} The red line represents the baseline where the variance difference is $0$.}
\label{app:kr}
\end{figure}

As a supplement to Figure \ref{fig:kr} in Section \ref{sec:kernel}, Figure \ref{app:kr} shows the variance differences across test points by plotting the boxplots of $\{\text{Var}(\tilde{y}_j )-\text{Var}(\hat{y}_{j,U} )\}^m_{j=1}$ and $\{\text{Var}(\tilde{y}_j )-\text{Var}(\hat{y}_{j,V} )\}^m_{j=1}$ with $B=\{25, 100\}$,  $n\in \{100,150,200,250\}$, and $k\in\{10, 15, 20\}$. The results confirm that the variance reduction property generally holds for both $ U $- and $ V $-statistics-based experience replay methods, further validating the effectiveness and robustness of our approach.

\begin{table}[ht!]
\centering
\fontsize{9}{10}\selectfont
\renewcommand{\arraystretch}{0.8} 
\begin{tabular}{lllllll}
\toprule 
& \multicolumn{2}{c}{\( k = 10 \)} & \multicolumn{2}{c}{\( k = 15 \)} & \multicolumn{2}{c}{\( k = 20 \)} \\ \cmidrule(l{0.75em}r{0.75em}){2-3}
\cmidrule(l{0.75em}r{0.75em}){4-5}
\cmidrule(l{0.75em}r{0.75em}){6-7}
\( n \) & \( {t} - {t}_U \) & \( {t} - {t}_V \) & \( {t} - {t}_U \) & \( {t} - {t}_V \) & \( {t} - {t}_U \) & \( {t} - {t}_V \) \\ \midrule
200  & 1.279 & 1.118 & 1.139 & 1.105 & 0.968 & 1.018 \\  
250  & 3.792 & 3.811 & 3.658 & 3.620 & 3.473 & 3.449 \\ \bottomrule 
\end{tabular}%
\caption{ Time cost reduction achieved by experience replay methods (measured in seconds) with \( B=25 \) for different values of \( k \) and \( n \).}
\label{tab:2} 

\end{table}

Table \ref{tab:2} presents the time cost reduction achieved by the experience replay methods with $B=25$, $k \in \{10, 15, 20\}$, and $n \in \{200, 250\}$. Here, ${t}$ represents the total time cost across all experiments without experience replay, while ${t}_U$ and ${t}_V$ represent the total time costs with experience replay based on resampled $U$- and $V$-statistics, respectively.
The results demonstrate that, for a fixed $B$, the experience replay method reduces the computational cost in time, particularly when $k$ is small and $n$ is large.

\subsubsection{Real Data Analysis}\label{app:real:data}

We study the \texttt{Boston} dataset from the \texttt{R} package \texttt{MASS}, which contains information collected by the U.S. Census Bureau regarding housing in the Boston area. The task is to predict the median value of owner-occupied homes.
We randomly sample $ m = 100 $ observations from the dataset as the test set $ \mathcal{D}_{\text{test}}$. For each experiment, we randomly draw $ n $ observations from the remaining data to form the training dataset $ \mathcal{D}_n $. Following the same procedure as in the simulation study, we conduct $ M = 100 $ experiments and calculate $ \text{Var}(\tilde{y}_j) $, $ \text{Var}(\hat{y}_{j,U}) $, and $ \text{Var}(\hat{y}_{j,V}) $ for each test point $x_j$. 

\begin{figure}[ht!]
\centering
\includegraphics[width=0.15\textwidth]{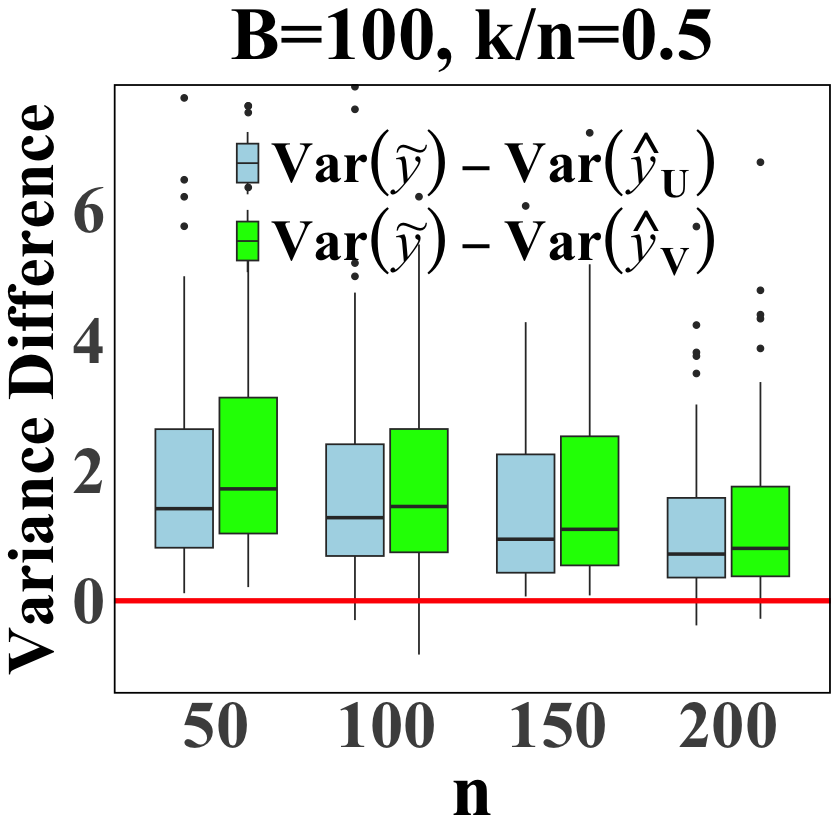} 
\includegraphics[width=0.15\textwidth]{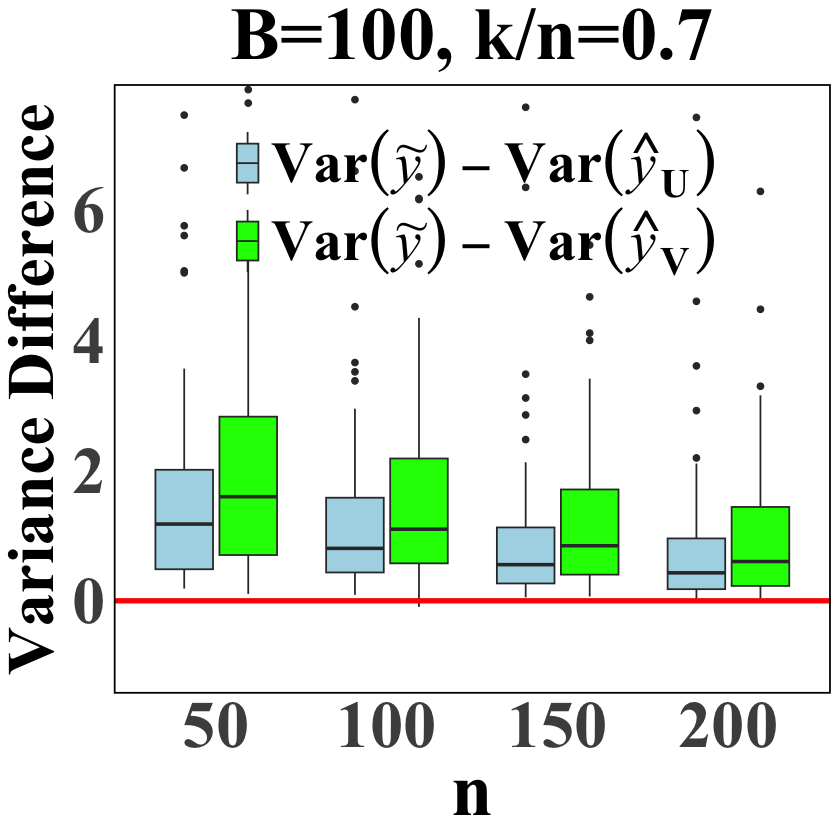} 
\includegraphics[width=0.15\textwidth]{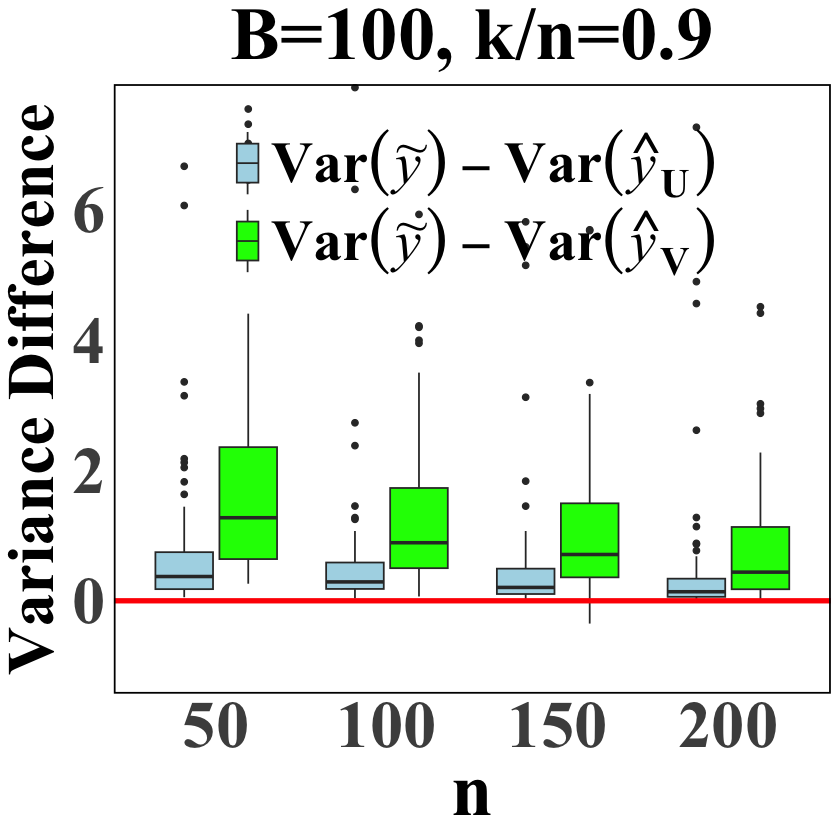} \\
\includegraphics[width=0.15\textwidth]{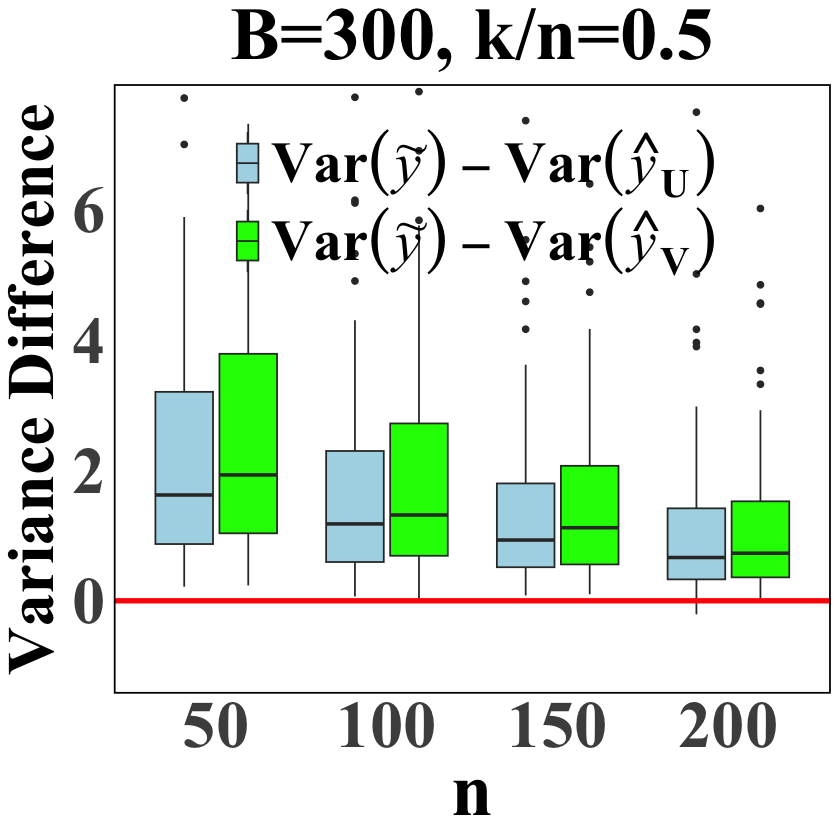} 
\includegraphics[width=0.15\textwidth]{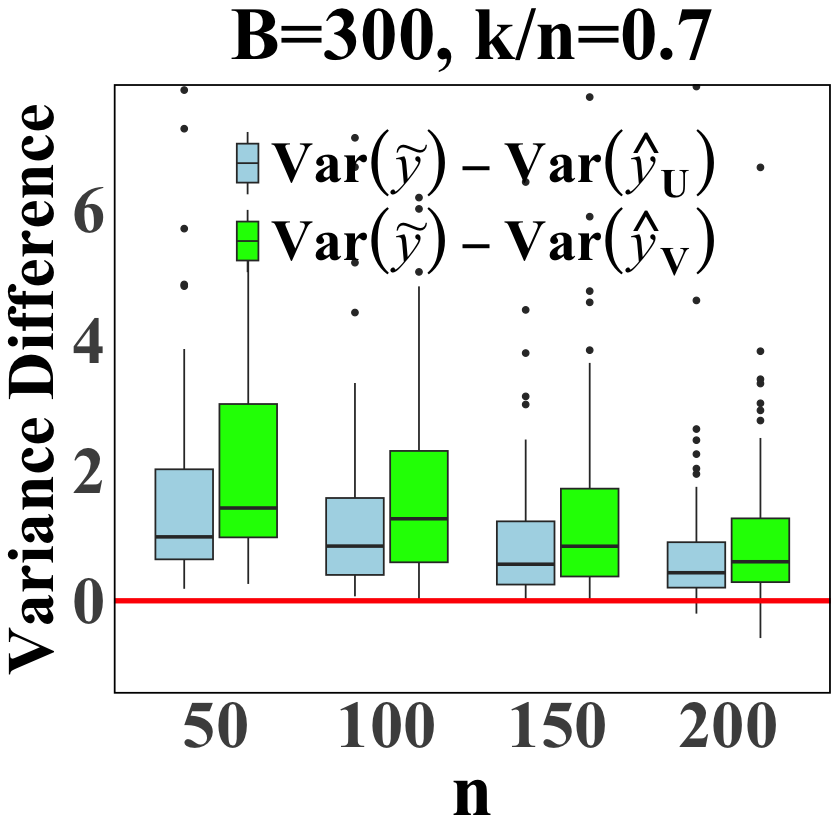} 
\includegraphics[width=0.15\textwidth]{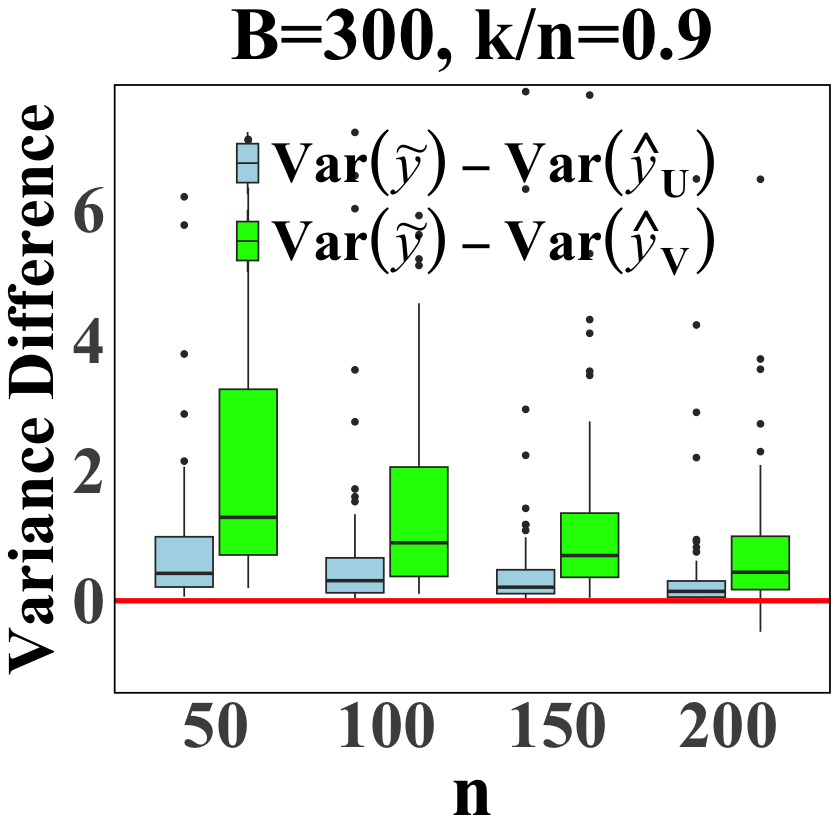} \\
\includegraphics[width=0.15\textwidth]{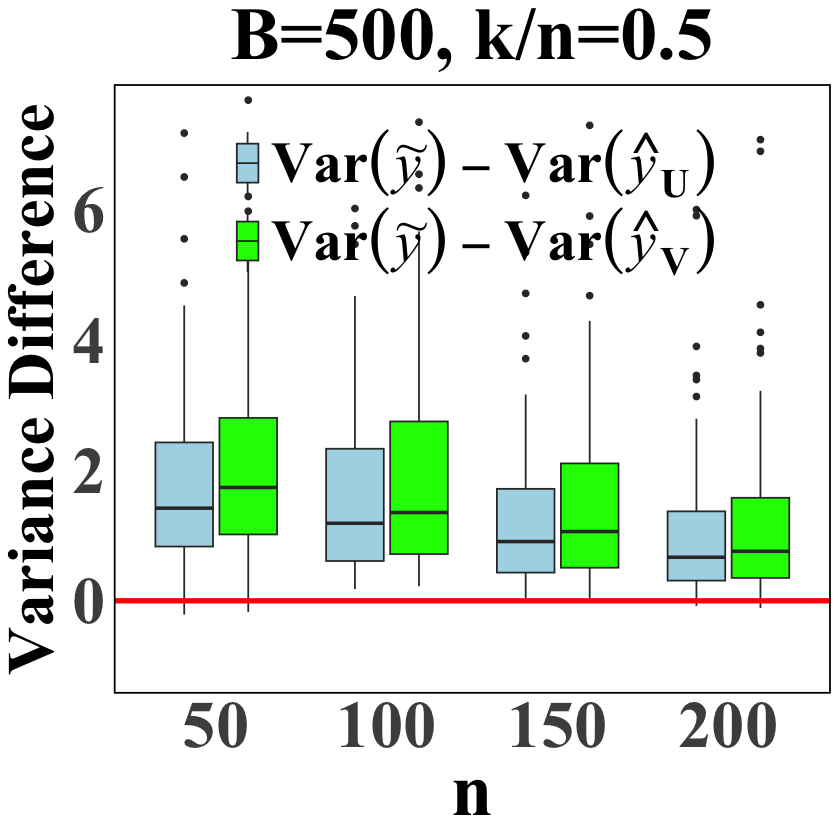} 
\includegraphics[width=0.15\textwidth]{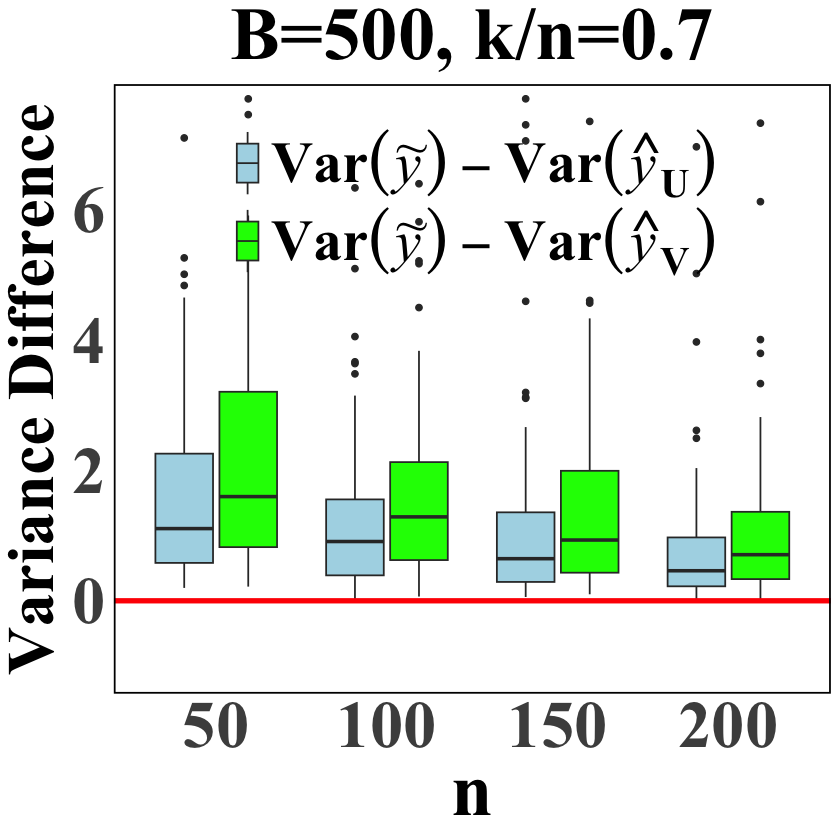} 
\includegraphics[width=0.15\textwidth]{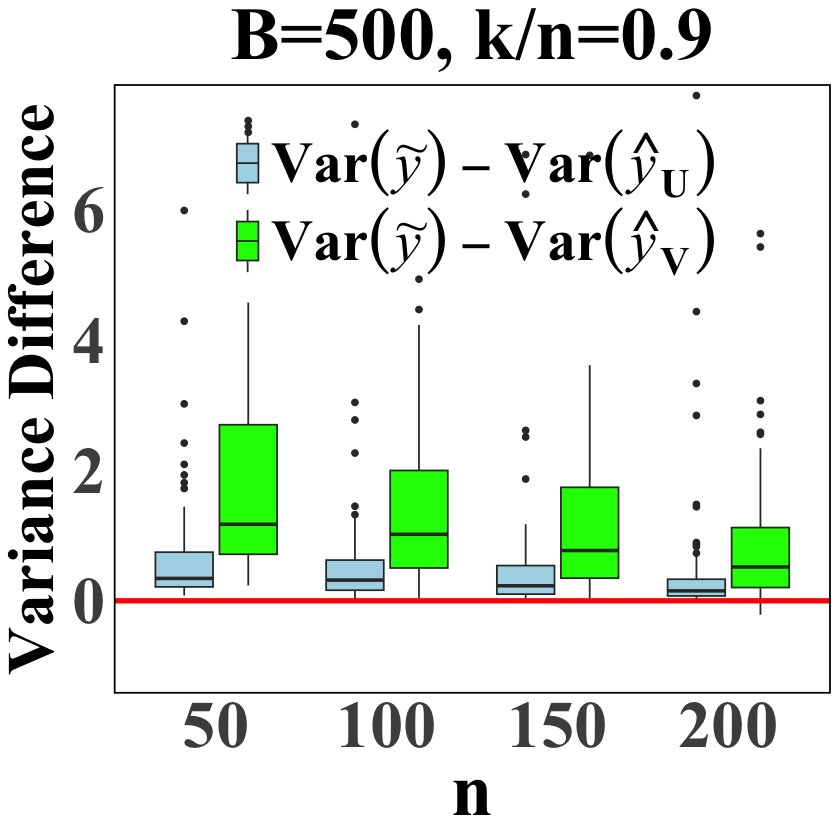} 
\caption{ Variance differences in predicted outcomes using kernel ridge regression on the \texttt{Boston} dataset, with $ m = 100 $, $ M = 100 $, evaluated across various values of $ n $, $ B $, and $ k/n $. \textcolor{black}{$\tilde{y}$ represents the results without experience replay, while $\hat{y}_{U}$ and $\hat{y}_{V}$ represent the results with experience replay.
} The red line represents the baseline where the variance difference is $0$.}
\label{fig:kr_real}
\end{figure}

Figure \ref{fig:kr_real} presents the boxplots of  $\{\text{Var}(\tilde{y}_j )-\text{Var}(\hat{y}_{j,U} )\}^m_{j=1}$ and $\{\text{Var}(\tilde{y}_j )-\text{Var}(\hat{y}_{j,V} )\}^m_{j=1}$ for different values of $n$, $B$, and $k/n$. We choose  $n\in \{50, 100,150,200\}$, $B\in\{100, 300, 500\}$, and $k/n\in\{0.5, 0.7, 0.9\}$. The results confirm that the variance reduction property holds across all settings for both $ U $- and $ V $-statistics-based experience replay methods.

\setcounter{secnumdepth}{2}
\section{RMSE Comparison}\label{sec:appaddexp}
In addition to checking the variance reduction property, we also compare the root mean squared error (RMSE) of the proposed methods with the original methods over the $ m $ test points across all $ M $ experiments.

\subsection{Reinforcement Leaning Policy Evaluation}\label{rmse pe}
In the experiment setting of  Section \ref{exp:PE}, for each experiment $i=1,2,\dots, M$, we define the following RMSE over the $m$ test points
\begin{equation*}
  \tilde{R}_i=\sqrt{\frac{1}{m}\sum_{j=1}^m (\tilde{V}^i(s_j^*)-V(s_j^*))^2}, \quad 
\end{equation*}
\begin{equation*}
    \hat{R}_{i, U}=\sqrt{\frac{1}{m}\sum_{j=1}^m (\hat{V}_{U}^i(s_j^*)-V(s_j^*))^2}, 
\end{equation*}
\begin{equation*}
    \hat{R}_{i, V}=\sqrt{\frac{1}{m}\sum_{j=1}^m (\hat{V}_{V}^i(s_j^*)-V(s_j^*))^2},
\end{equation*}
where \(\tilde{V}^i(s_j^*)\), \(\hat{V}_{U}^i(s_j^*)\), and \(\hat{V}_{V}^i(s_j^*)\) denote the predicted values of \(s_j^*\) using the three methods in the \(i\)-th experiment, and $V(s^*_j)=\text{cos}^3(s^*_j)$.

We compare the prediction errors by comparing $\tilde{{R}}_i, \hat{{R}}_{i, U}$, and $\hat{{R}}_{i, V}$ for $i=1,\dots, M$.

\subsubsection{LSTD Algorithm}\label{rmse lstd}

\begin{figure}[htb]
\centering
\includegraphics[width=0.15\textwidth]{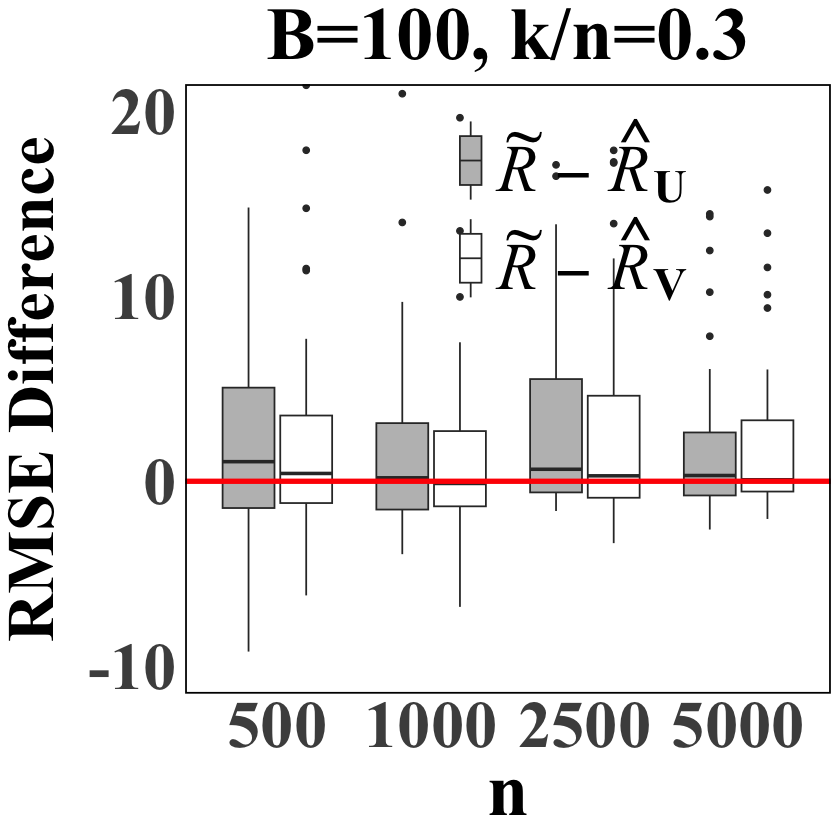} 
\includegraphics[width=0.15\textwidth]{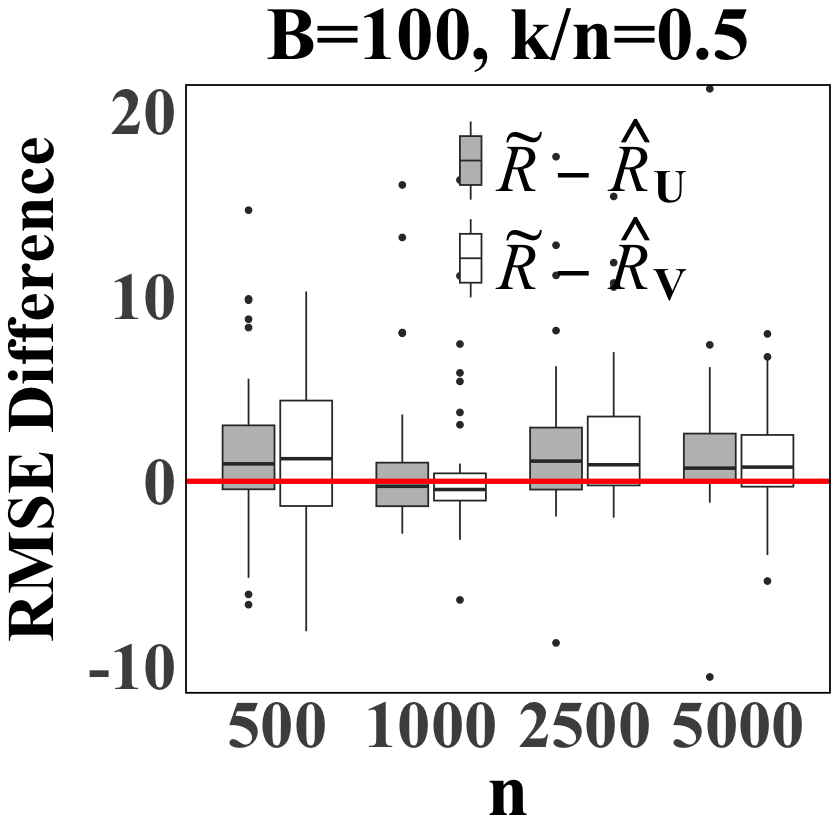} 
\includegraphics[width=0.15\textwidth]{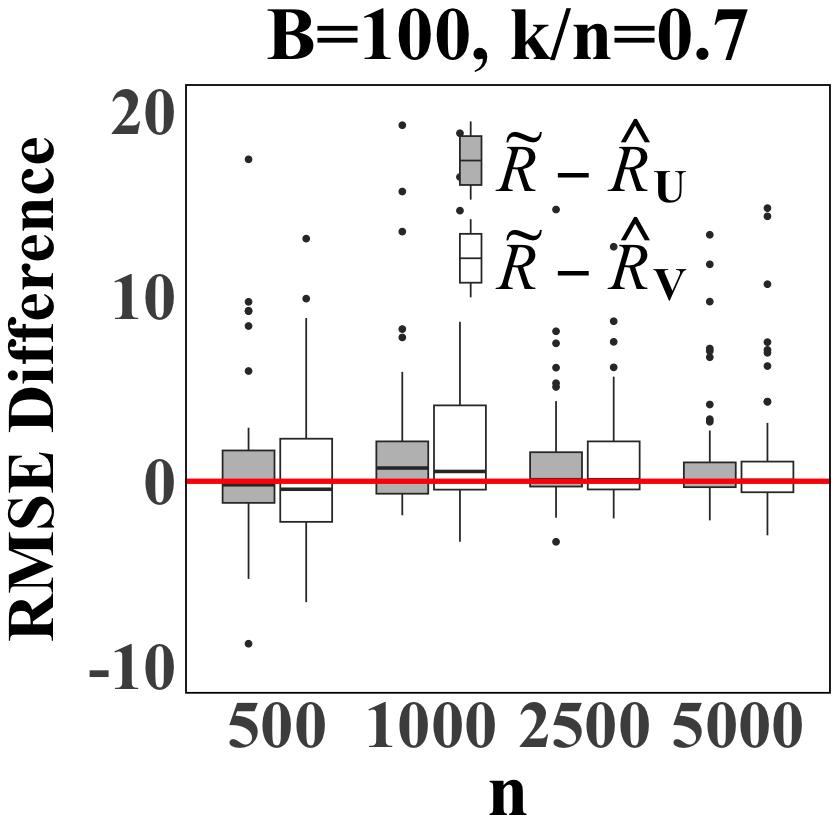} \\
\includegraphics[width=0.15\textwidth]{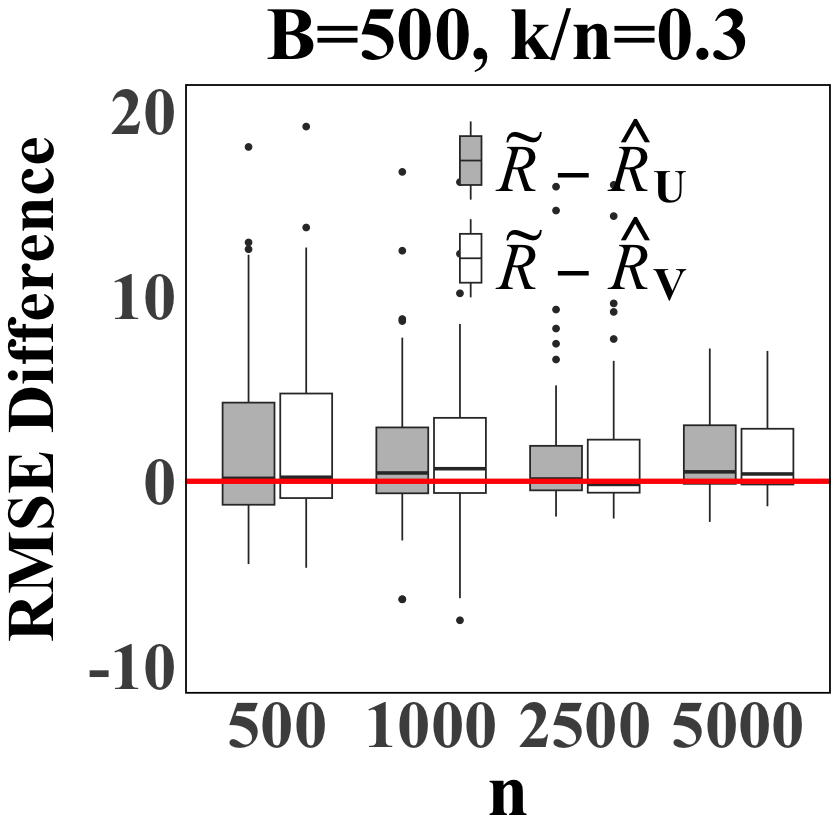} 
\includegraphics[width=0.15\textwidth]{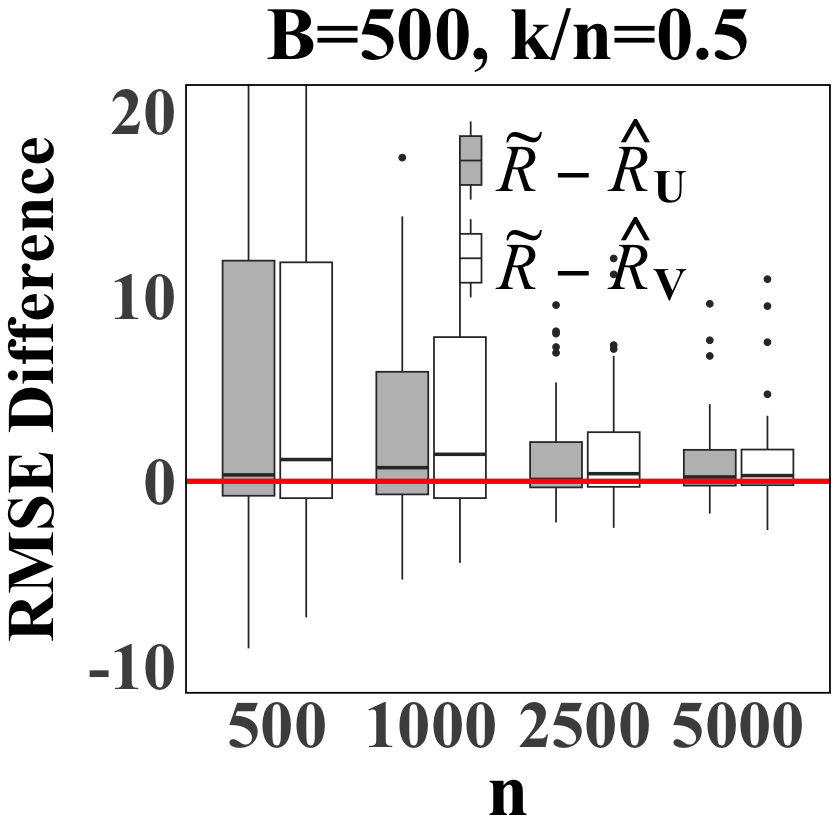} 
\includegraphics[width=0.15\textwidth]{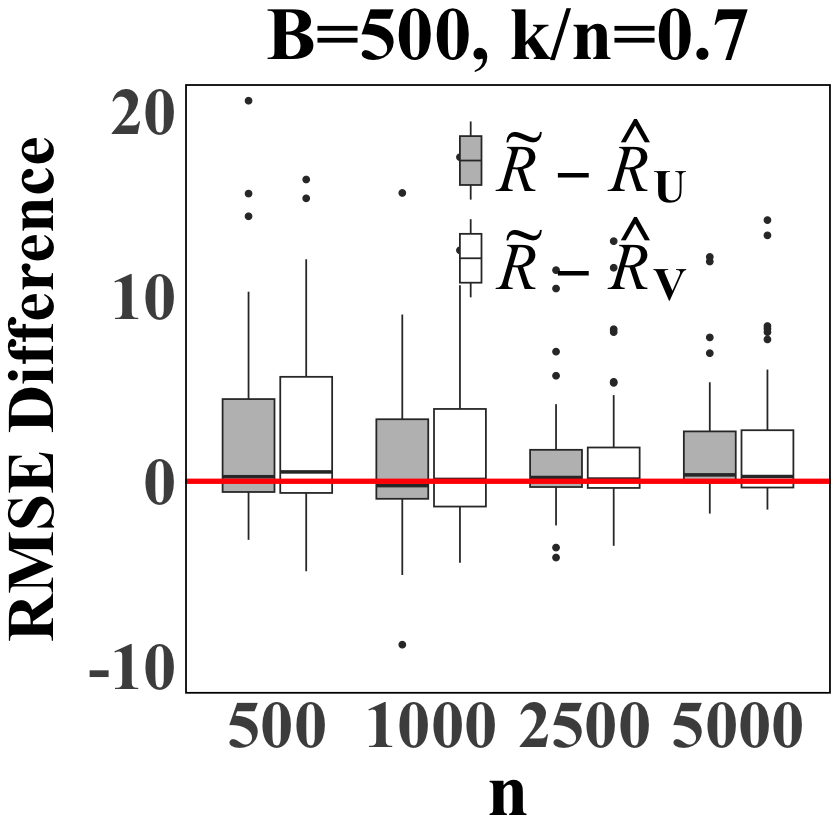} \\
\includegraphics[width=0.15\textwidth]{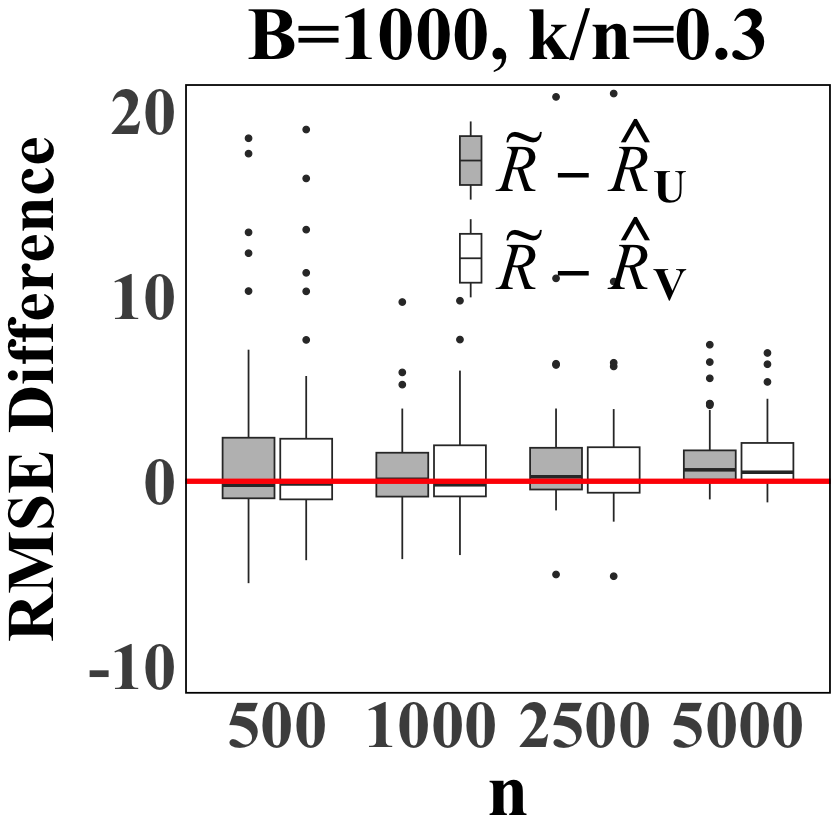} 
\includegraphics[width=0.15\textwidth]{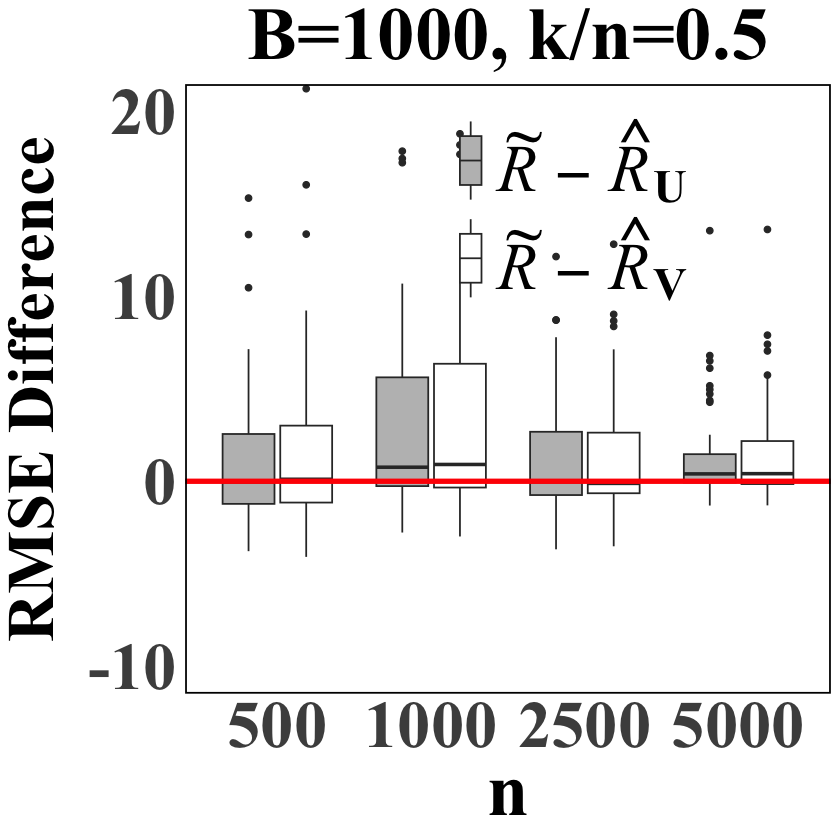} 
\includegraphics[width=0.15\textwidth]{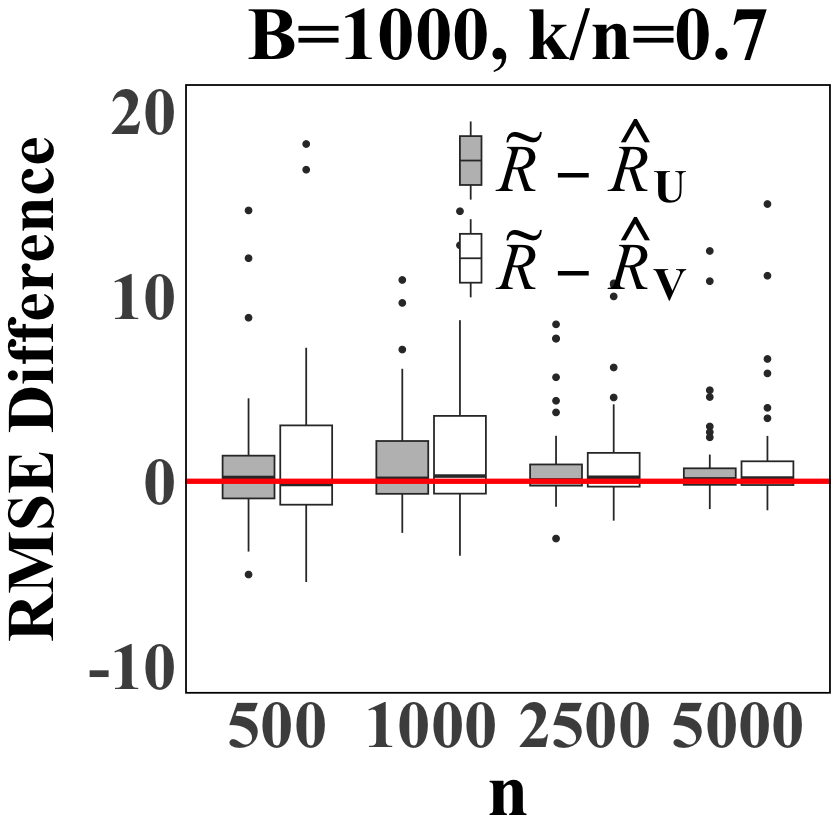} 
\caption{ RMSE differences among the predicted policy values using the LSTD algorithm with $ m = 50 $ and $ M = 50 $, evaluated across various values of $ n $, $ B $, and $ k/n $. \textcolor{black}{$\tilde{R}$ denotes the RMSE without experience relay, while $\hat{R}_{U}$ and $\hat{R}_{V}$ represent the RMSE with experience replay.}
The red line represents the baseline where the variance difference is 0.}
\label{fig:RL_LSTDMSE}
\end{figure}

Figure \ref{fig:RL_LSTDMSE} compares the RMSE of the LSTD algorithm by drawing the boxplots of the differences $\{\tilde{R}_i-\hat{R}_{i, U}\}_{i=1}^M$ and $\{\tilde{R}_i-\hat{R}_{i, V}\}_{i=1}^M$, with regard to different $n, B$, and the ratio $k/n$. We choose the $n\in\{500, 1000, 2500, 5000\}$, $B\in\{100, 500, 1000\}$, and $k/n\in\{0.3,0.5,0.7\}$. The results demonstrate that the combination of experience replay with the LSTD algorithm, regardless of the specific resampling method used, not only reduces variance but also tends to achieve smaller prediction errors, further highlighting its effectiveness.

\subsubsection{Second-Order PED-Based Algorithm}\label{rmse 2nd}
\begin{figure}[htb!]
\centering
\includegraphics[width=0.15\textwidth]{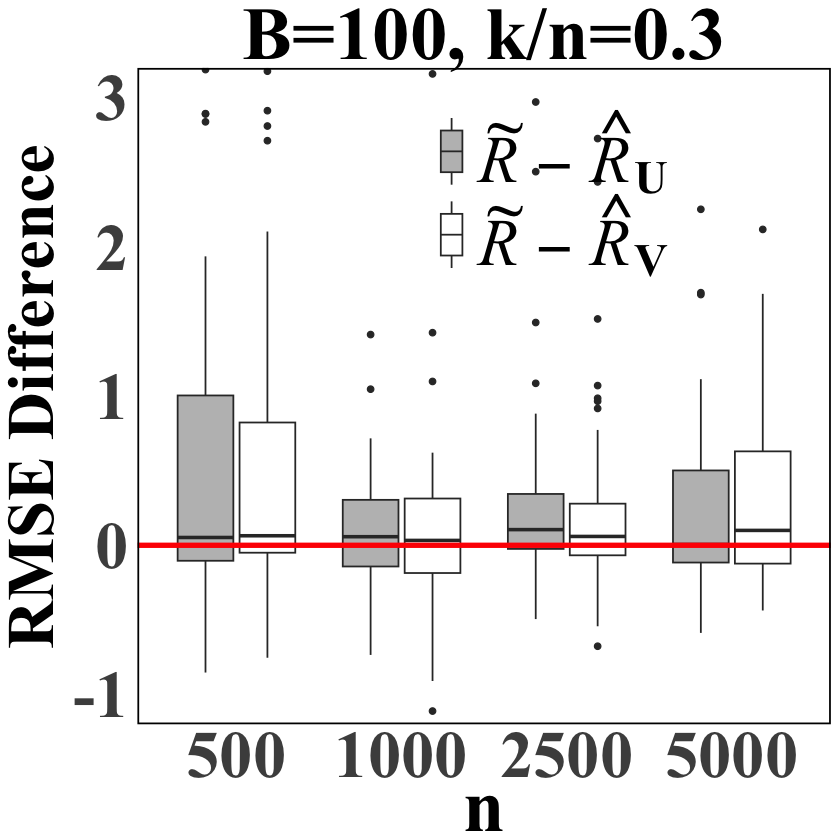} 
\includegraphics[width=0.15\textwidth]{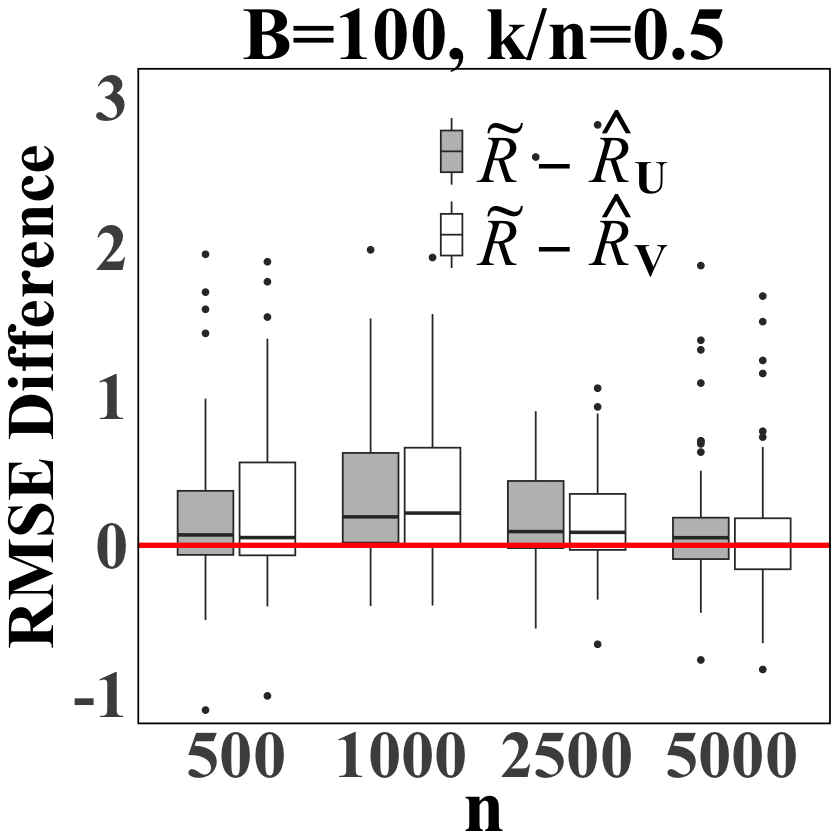} 
\includegraphics[width=0.15\textwidth]{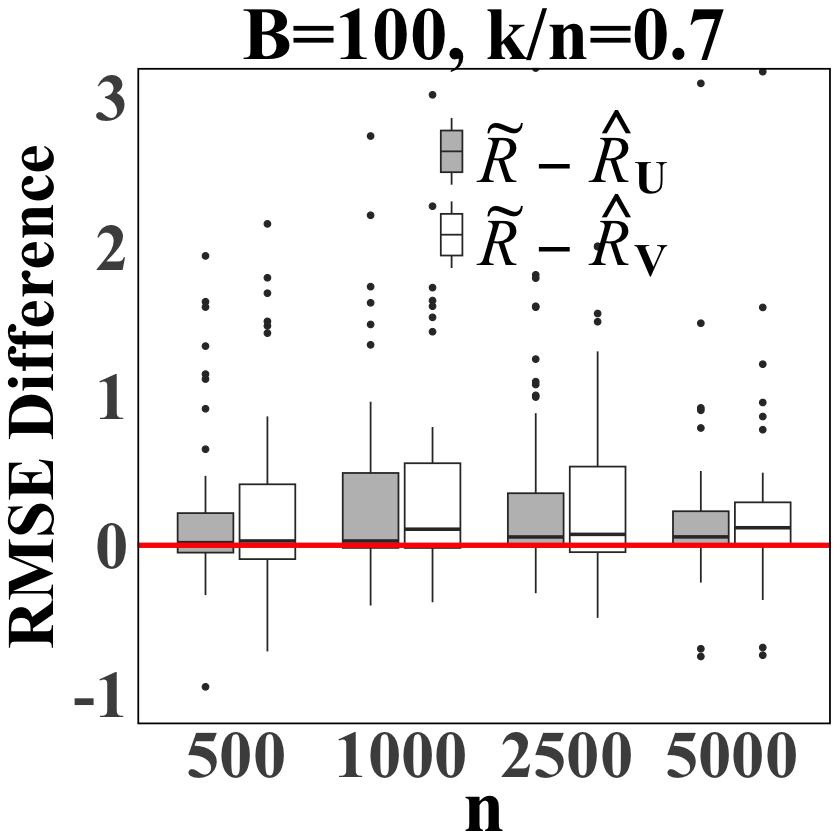} \\
\includegraphics[width=0.15\textwidth]{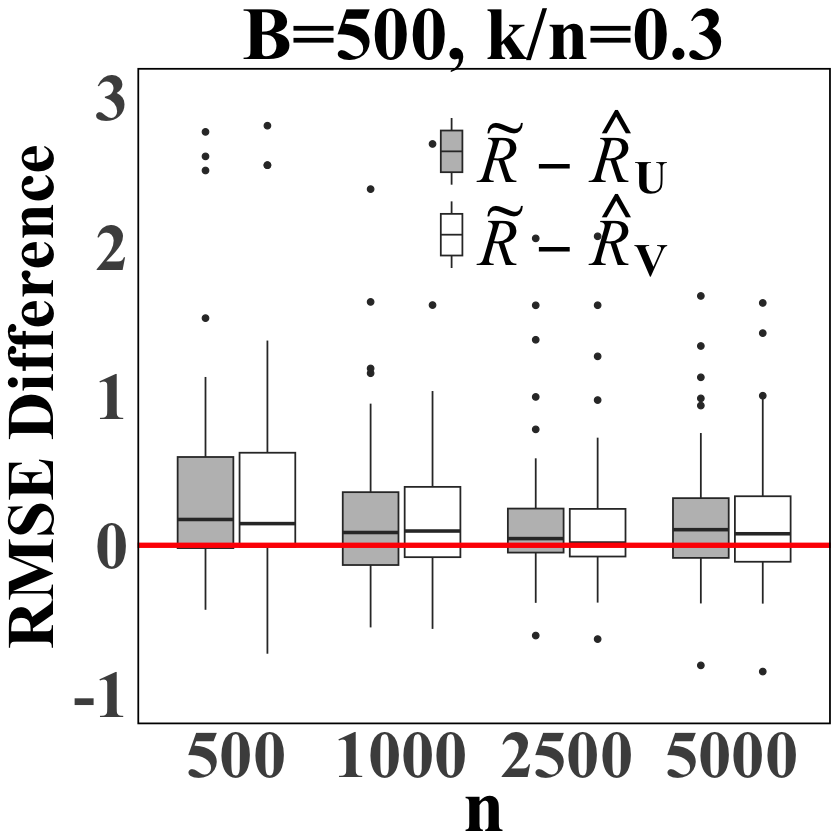} 
\includegraphics[width=0.15\textwidth]{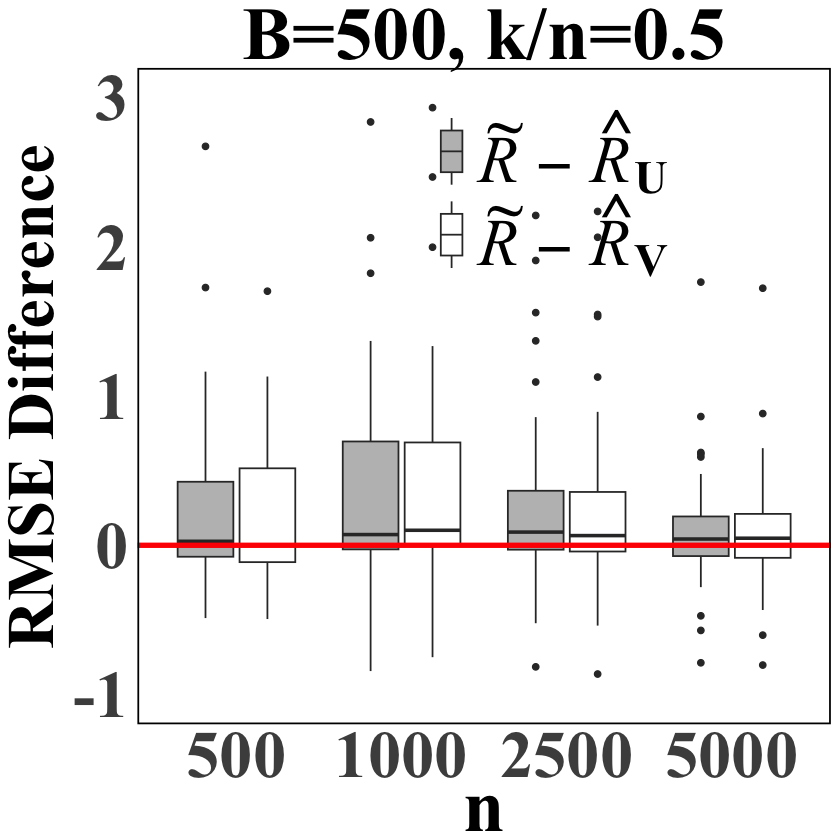} 
\includegraphics[width=0.15\textwidth]{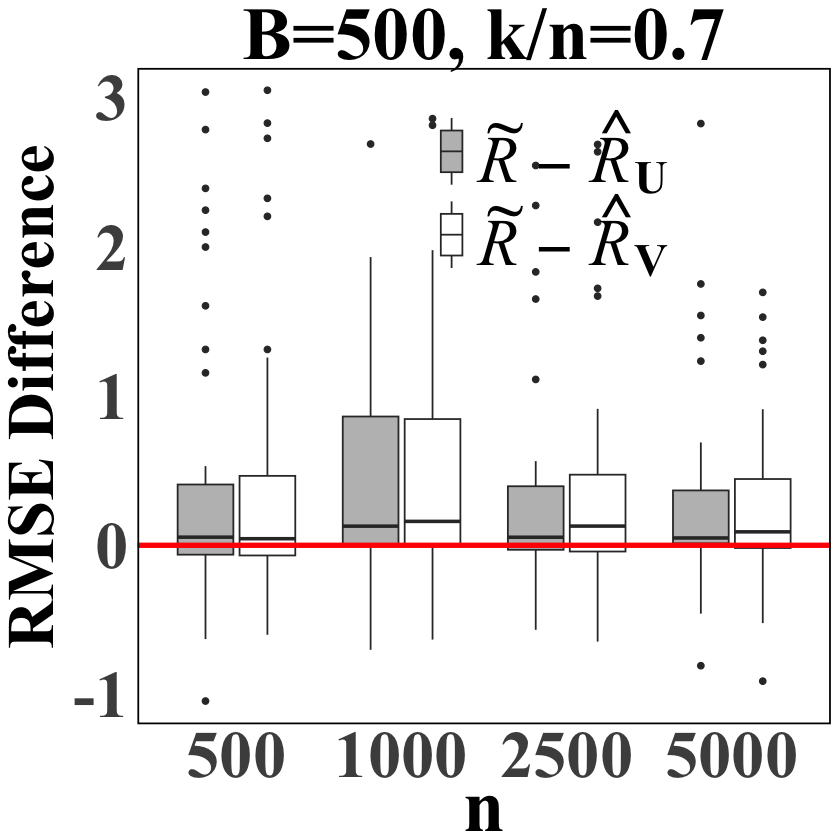} \\
\includegraphics[width=0.15\textwidth]{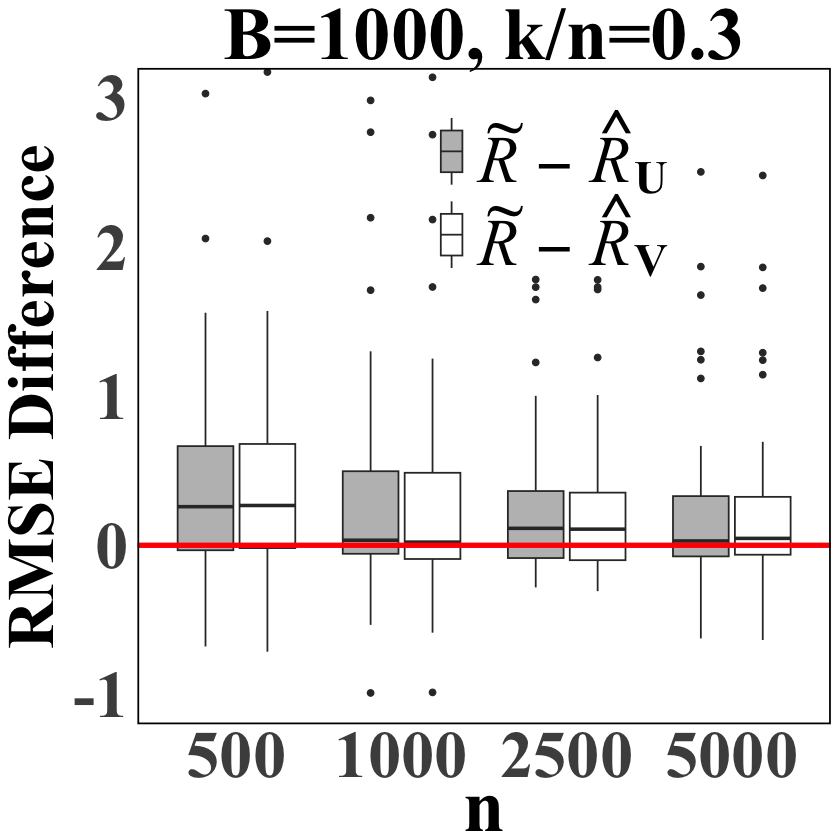} 
\includegraphics[width=0.15\textwidth]{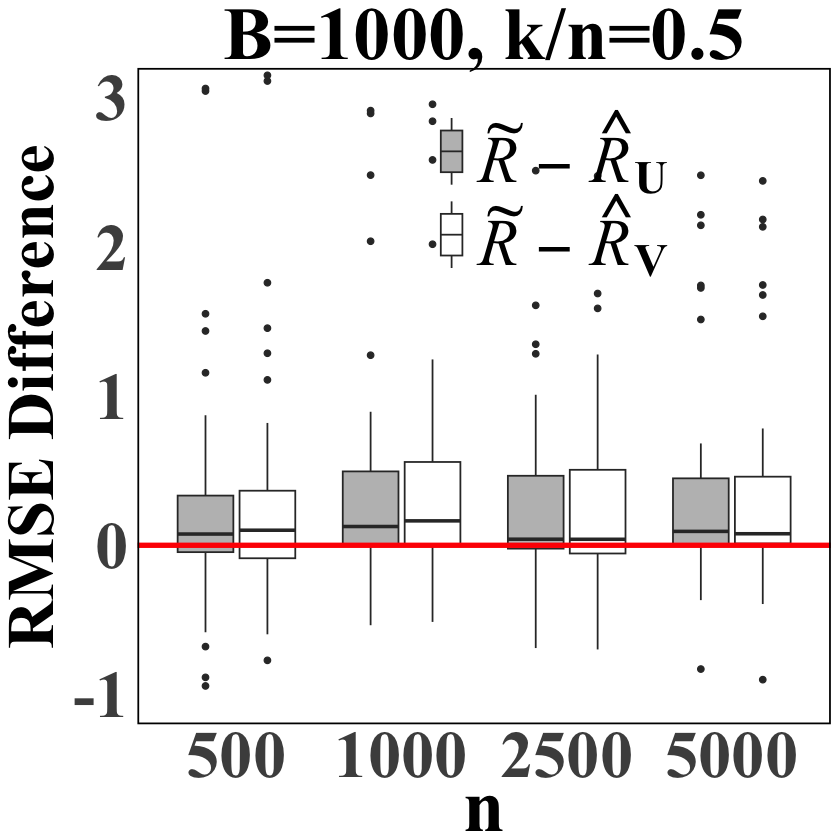} 
\includegraphics[width=0.15\textwidth]{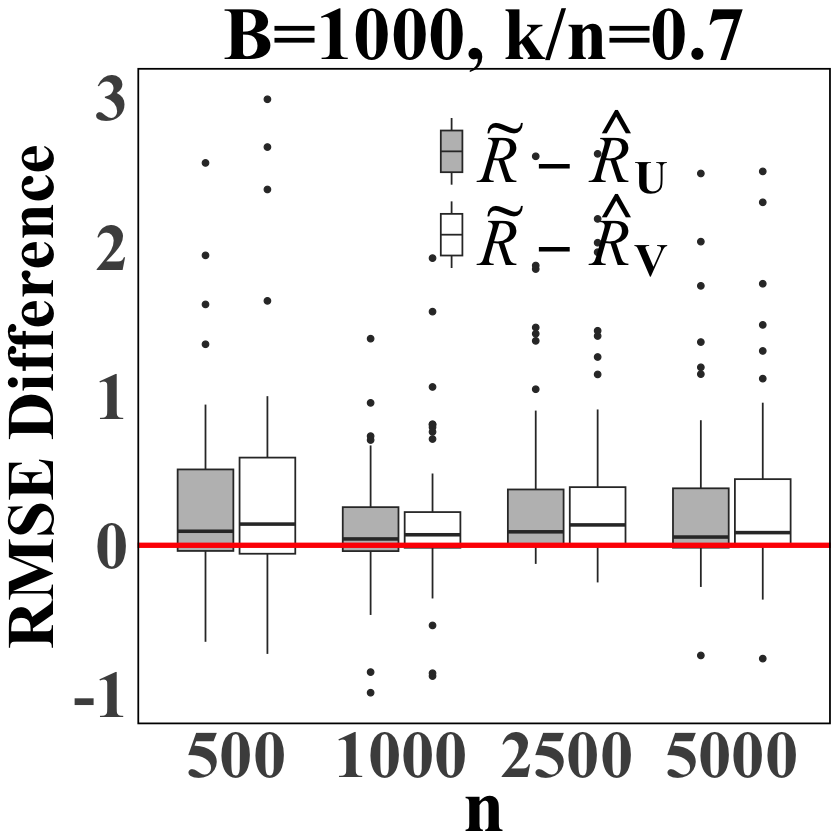} 
\caption{ RMSE differences among the predicted policy values using the second-order PDE-based algorithm with $ m = 50 $ and $ M = 50 $, evaluated across various values of $ n $, $ B $, and $ k/n $. \textcolor{black}{$\tilde{R}$ denotes the RMSE without experience relay, while $\hat{R}_{U}$ and $\hat{R}_{V}$ represent the RMSE with experience replay.} The red line represents the baseline where the variance difference is 0.}
\label{fig:RL_2ndrmse}
\end{figure}

Figure \ref{fig:RL_2ndrmse} compares the RMSE of the second-order PDE-based algorithm by drawing the boxplots of the differences $\{\tilde{{R}}_i-\hat{{R}}_{i, U}\}_{i=1}^M$ and $\{\tilde{{R}}_i-\hat{{R}}_{i, V}\}_{i=1}^M$, with regard to different $n, B$, and the ratio $k/n$. We choose the $n\in\{500, 1000, 2500, 5000\}$, $B\in\{100, 500, 1000\}$, and $k/n\in\{0.3,0.5,0.7\}$. The results show that incorporating experience replay into the PDE-based algorithm not only reduces variances, but consistently reduces prediction errors, demonstrating its effectiveness regardless of resampling method.

\subsubsection{First-Order PED-Based Algorithm}\label{rmse 1st}

\begin{figure}[htb]
\centering
\includegraphics[width=0.15\textwidth]{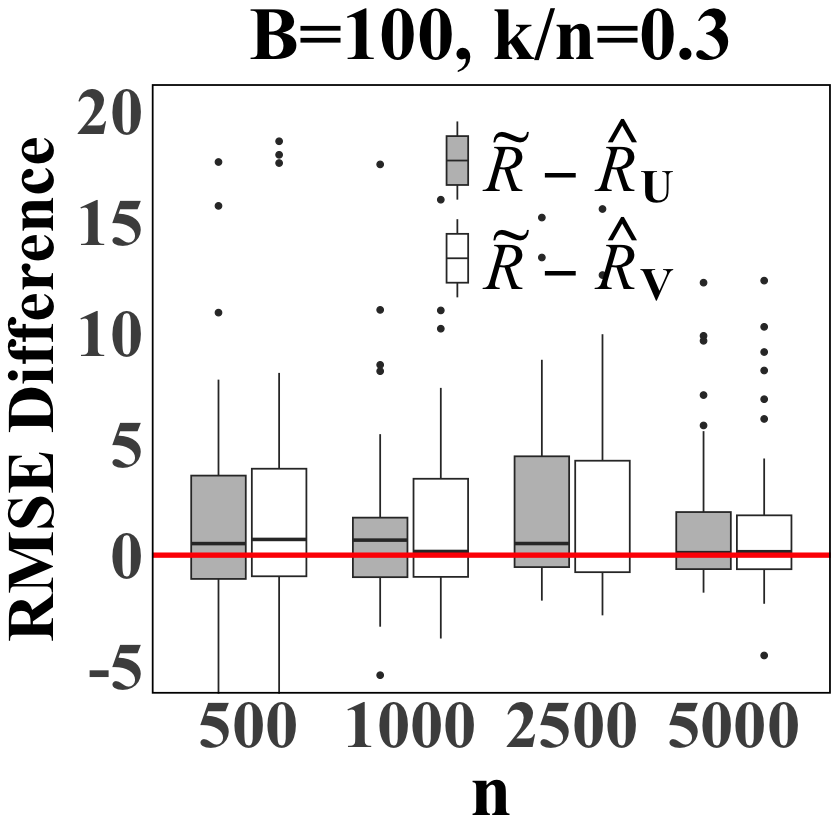} 
\includegraphics[width=0.15\textwidth]{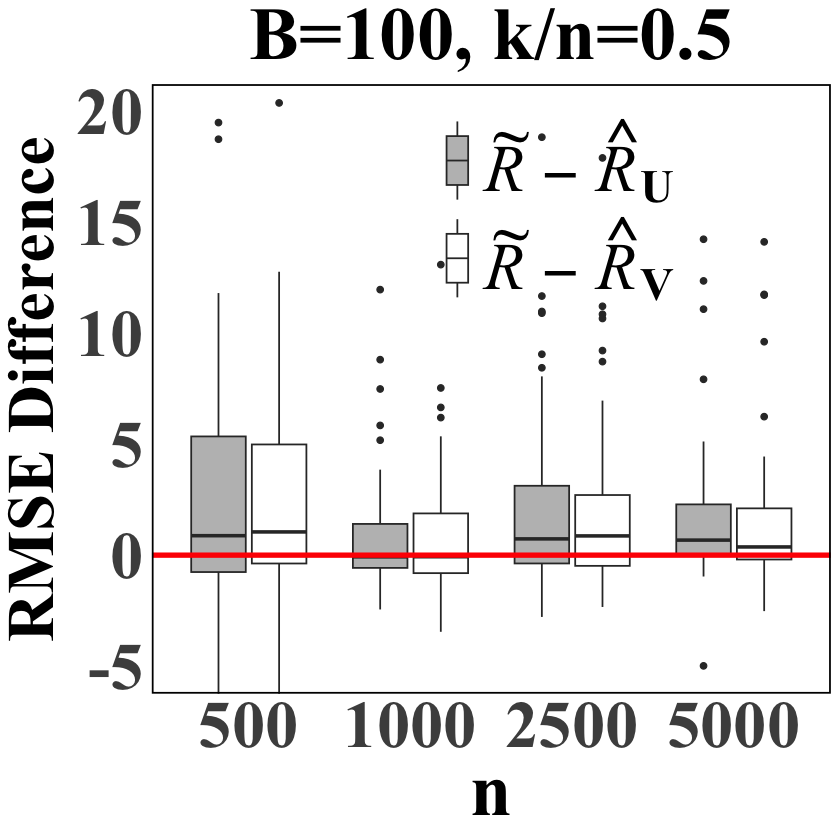} 
\includegraphics[width=0.15\textwidth]{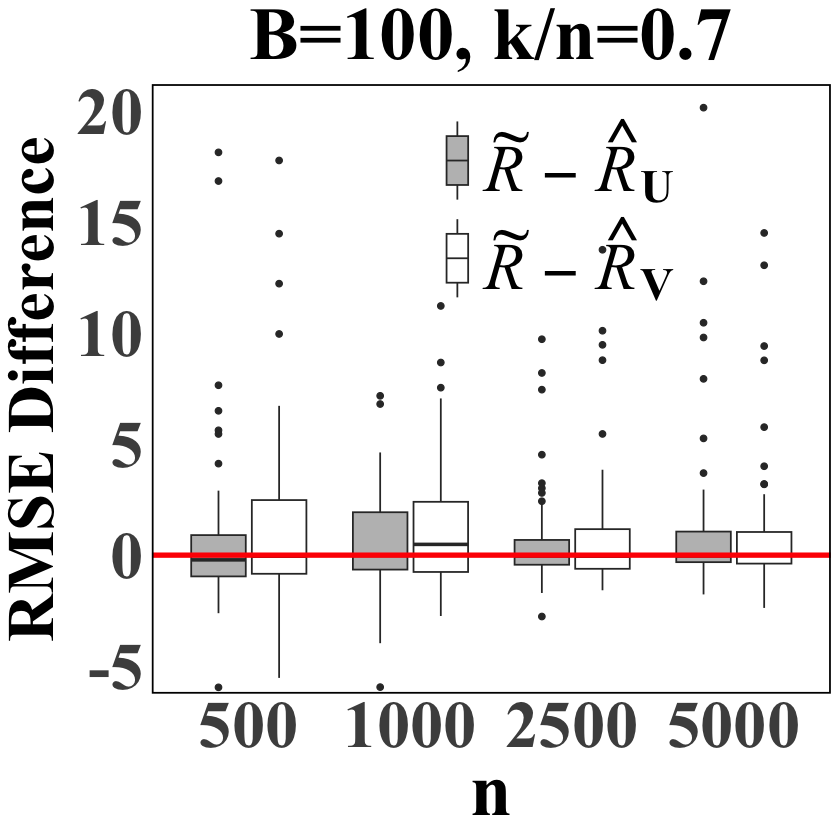} \\
\includegraphics[width=0.15\textwidth]{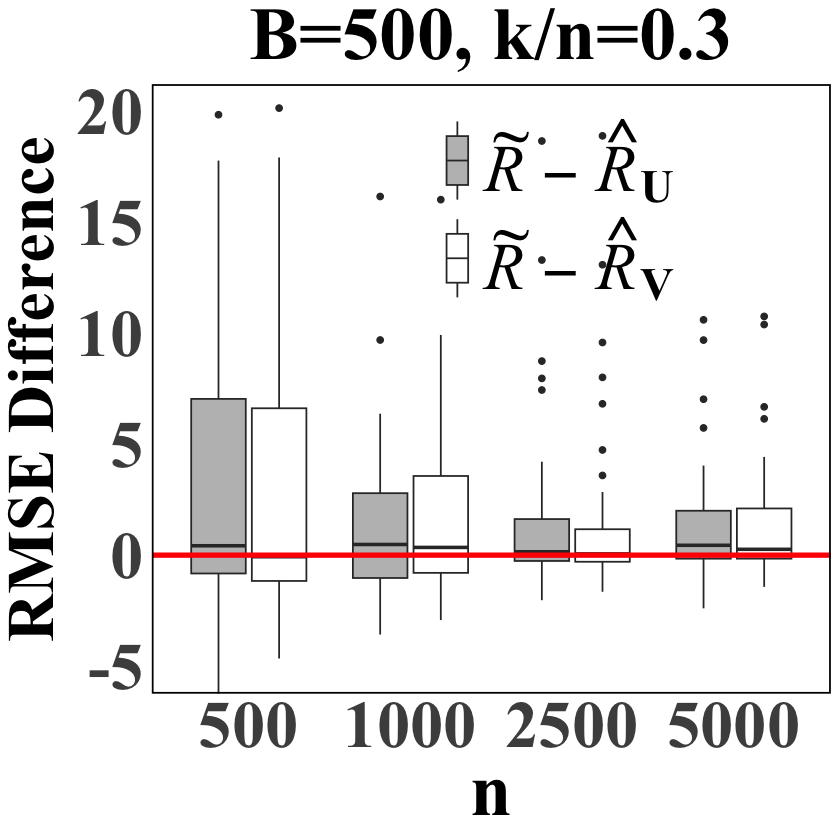} 
\includegraphics[width=0.15\textwidth]{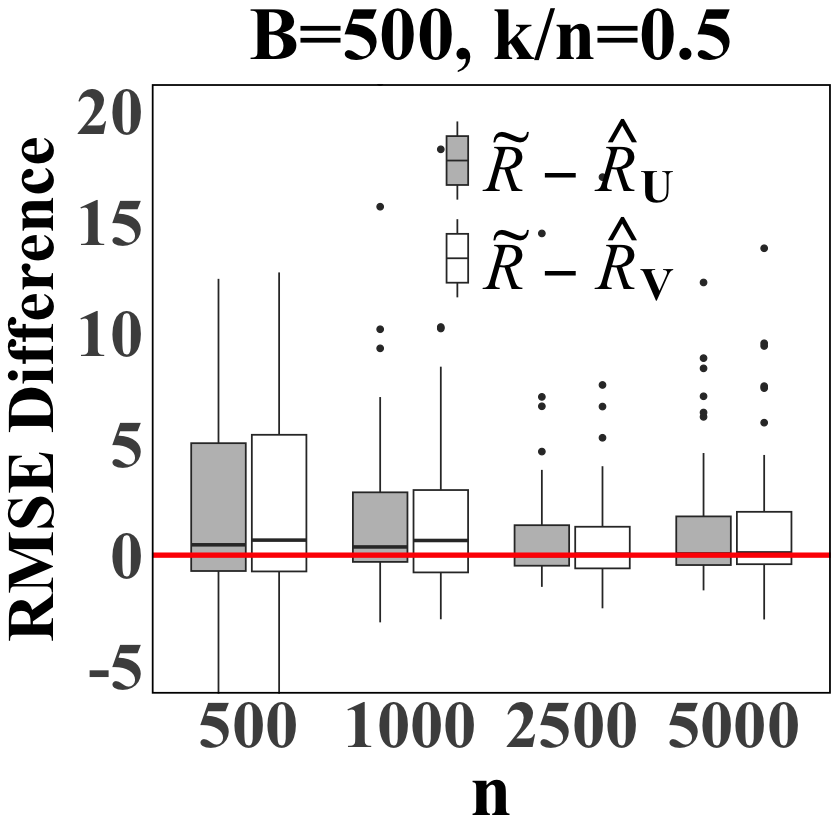} 
\includegraphics[width=0.15\textwidth]{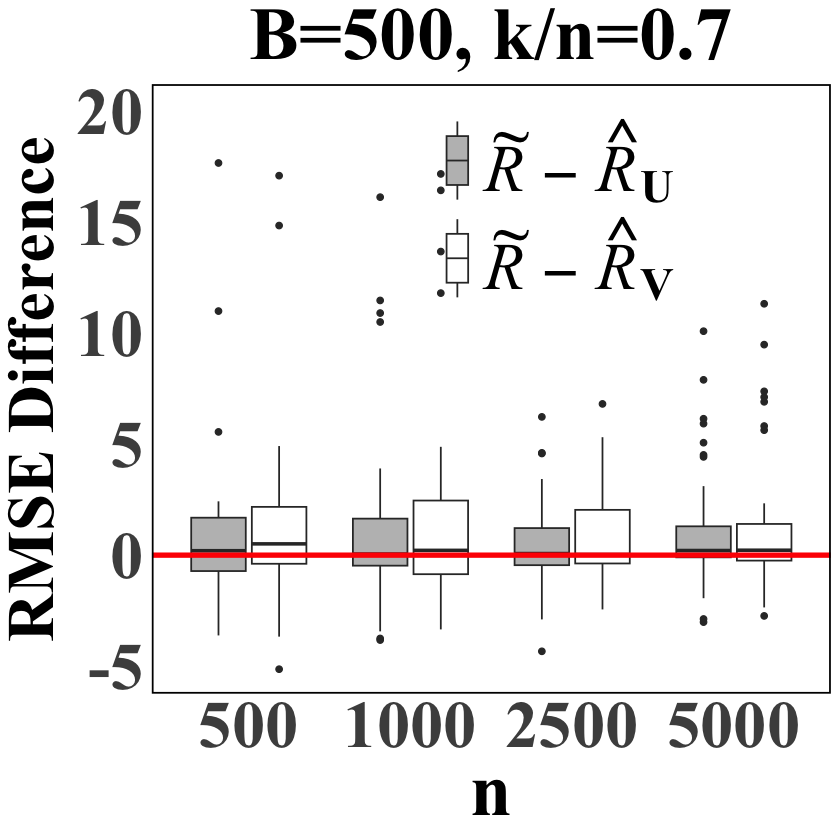} \\
\includegraphics[width=0.15\textwidth]{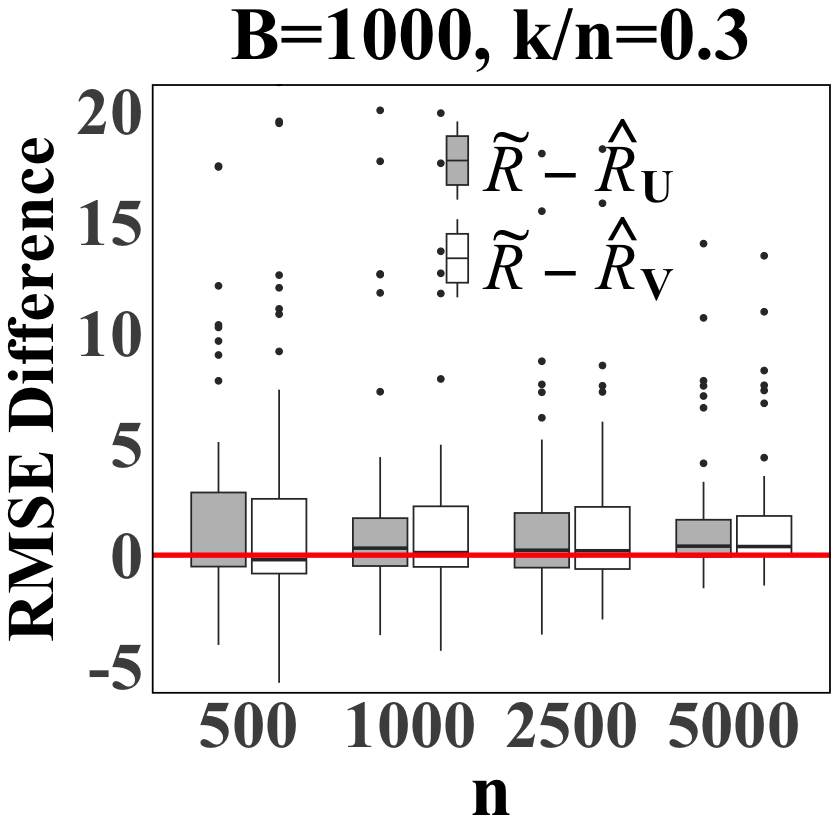} 
\includegraphics[width=0.15\textwidth]{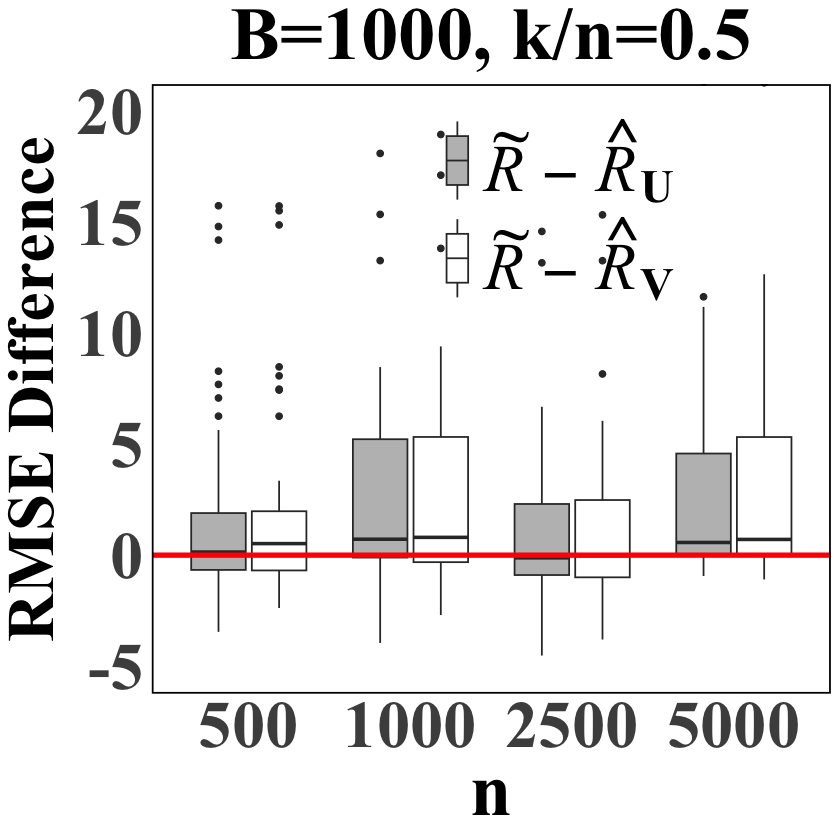} 
\includegraphics[width=0.15\textwidth]{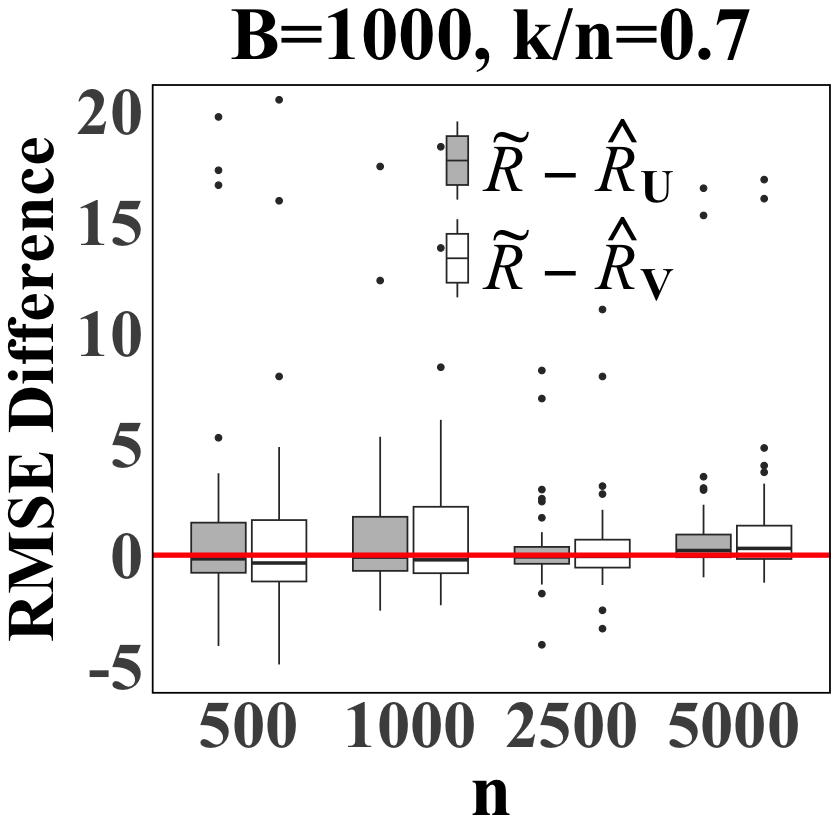} 
\caption{ RMSE differences among the predicted policy values using the first-order PDE-based algorithm with $ m = 50 $ and $ M = 50 $, evaluated across various values of $ n $, $ B $, and $ k/n $. \textcolor{black}{$\tilde{R}$ denotes the RMSE without experience relay, while $\hat{R}_{U}$ and $\hat{R}_{V}$ represent the RMSE with experience replay.} The red line represents the baseline where the variance difference is 0.}
\label{fig:RL_1STrmse}
\end{figure}

Figure \ref{fig:RL_1STrmse} compares the RMSE of the first-order PDE-based algorithm by drawing the boxplots of the differences $\{\tilde{{R}}_i-\hat{{R}}_{i, U}\}_{i=1}^M$ and $\{\tilde{{R}}_i-\hat{{R}}_{i, V}\}_{i=1}^M$, with regard to different $n, B$, and the ratio $k/n$. We select the $n\in\{500, 1000, 2500, 5000\}$, $B\in\{100, 500, 1000\}$, and $k/n\in\{0.3,0.5,0.7\}$. The results further indicate that incorporating experience replay into the PDE-based algorithm not only reduces variance but also consistently achieves reduced prediction errors, underscoring its effectiveness regardless of the resampling method used.

\subsection{Kernel Ridge Regression}\label{rmse simu}
In the experiment setting of Section \ref{sec:kernel}, for each experiment $i=1,2,\dots, M$, we define the following RMSE over the $m$ test points
\begin{equation*}
     \tilde{{R}}_i=\sqrt{\frac{1}{m}\sum_{j=1}^m (\tilde{y}_j^i-y_j)^2}, \quad \hat{{R}}_{i, U}=\sqrt{\frac{1}{m}\sum_{j=1}^m (\hat{y}_{j,U}^i-y_j)^2}, 
\end{equation*}
\begin{equation}\label{kr rmse}
 \hat{{R}}_{i, V}=\sqrt{\frac{1}{m}\sum_{j=1}^m (\hat{y}_{j,V}^i-y_j)^2},
\end{equation}
where \(\tilde{y}_j^i\), \(\hat{y}_{j,U}^i\), and \(\hat{y}_{j,V}^i\) denote the predicted values of \(x_j\) using the three methods in the \(i\)-th experiment.

We compare the prediction errors by comparing $\tilde{{R}}_i, \hat{{R}}_{i, U}$, and $\hat{{R}}_{i, V}$ for $i=1,\dots, M$.

\begin{figure}[htb!]
\centering
\includegraphics[width=0.15\textwidth]{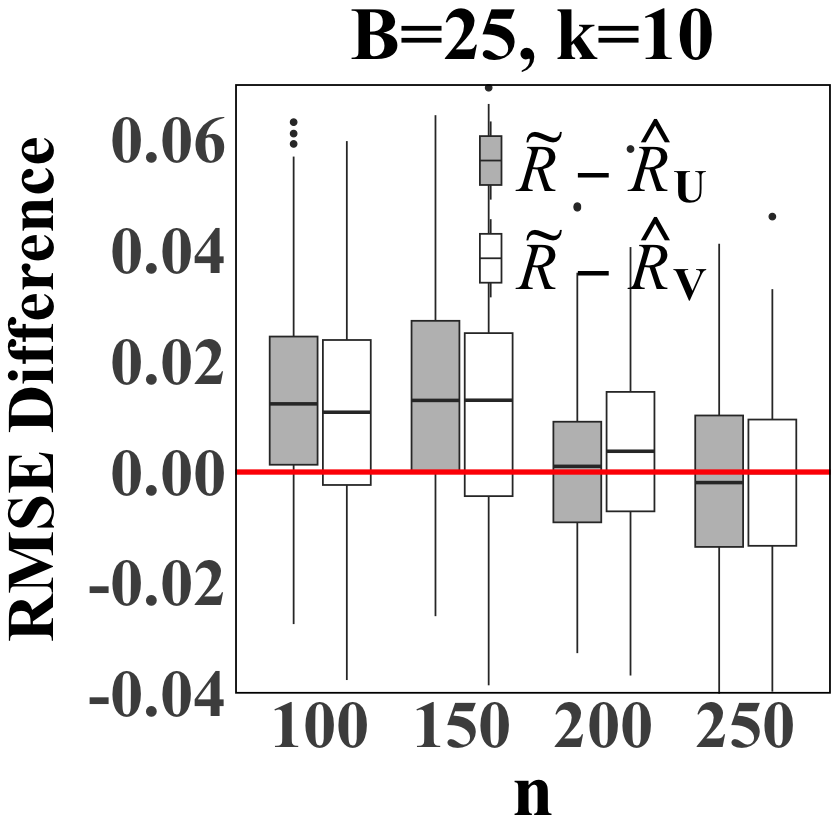} 
\includegraphics[width=0.15\textwidth]{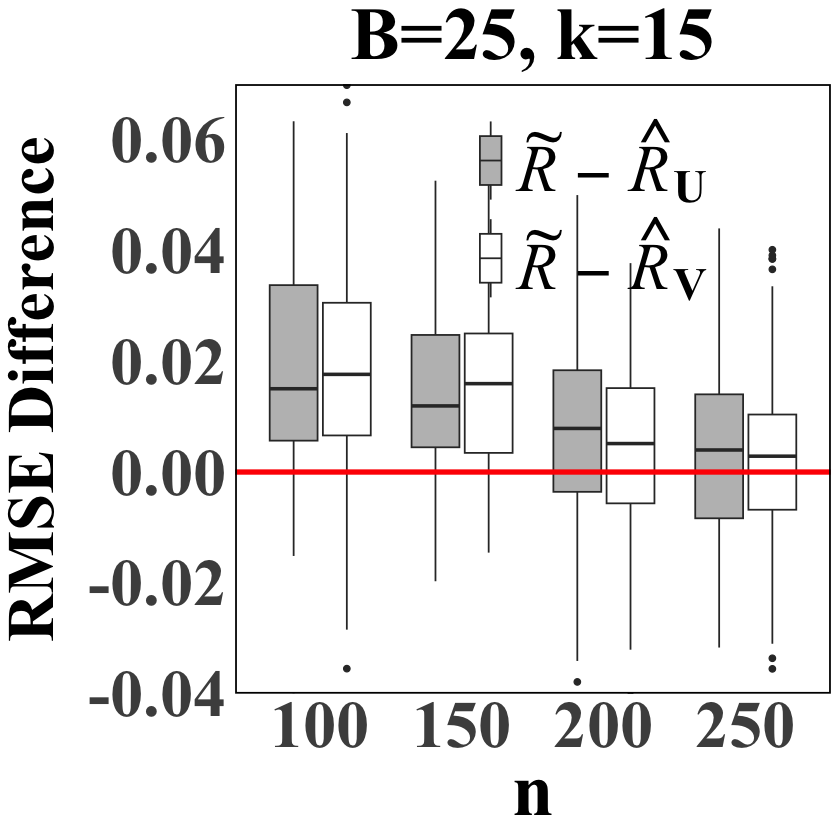} 
\includegraphics[width=0.15\textwidth]{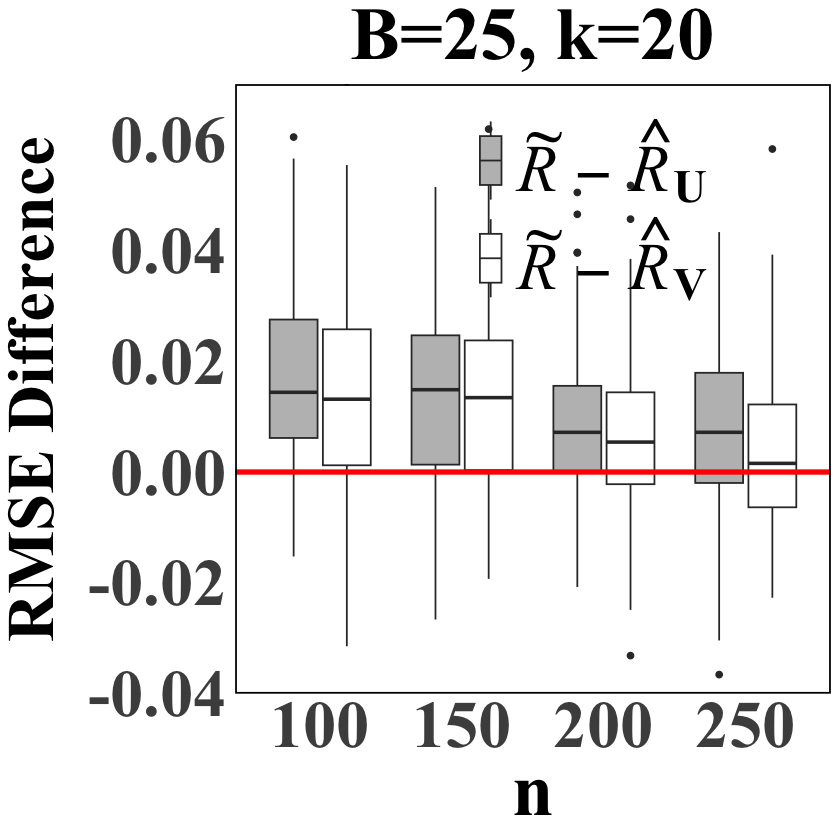} \\
\includegraphics[width=0.15\textwidth]{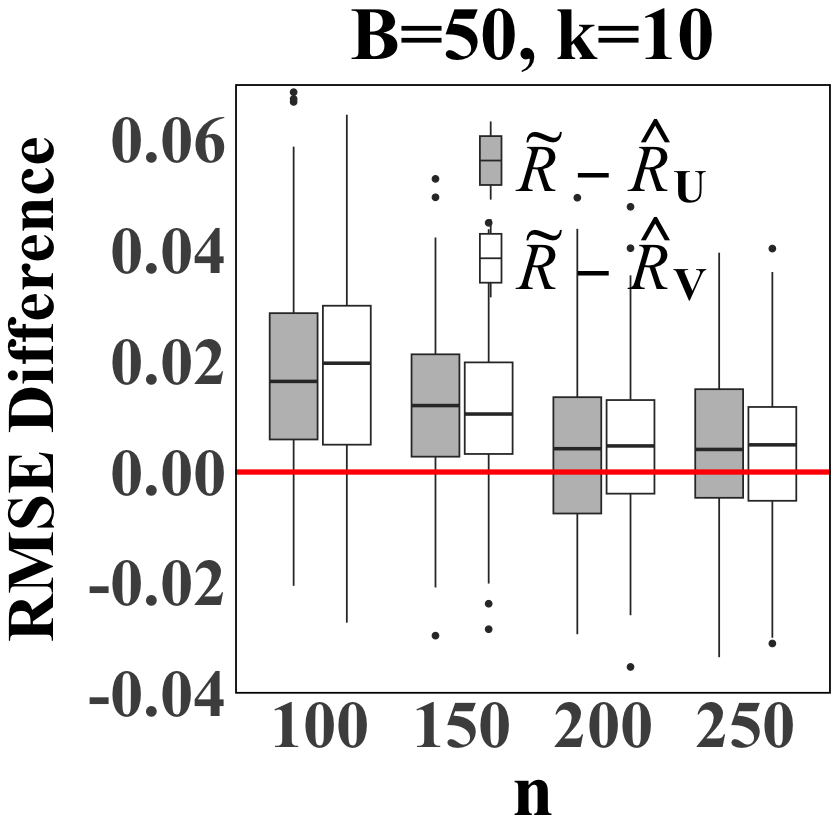} 
\includegraphics[width=0.15\textwidth]{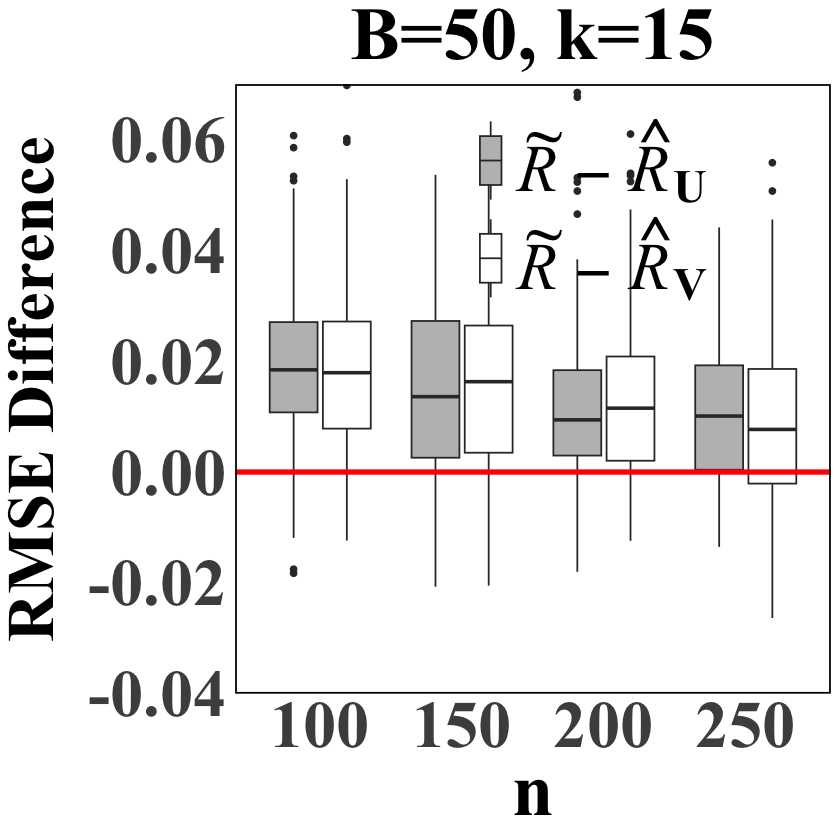} 
\includegraphics[width=0.15\textwidth]{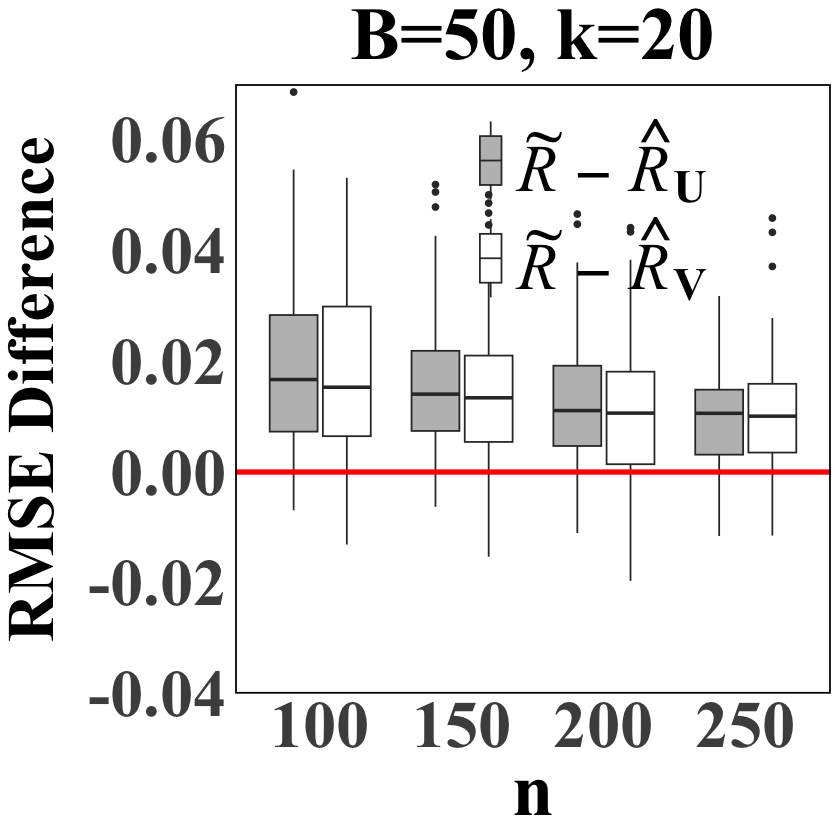} \\
\includegraphics[width=0.15\textwidth]{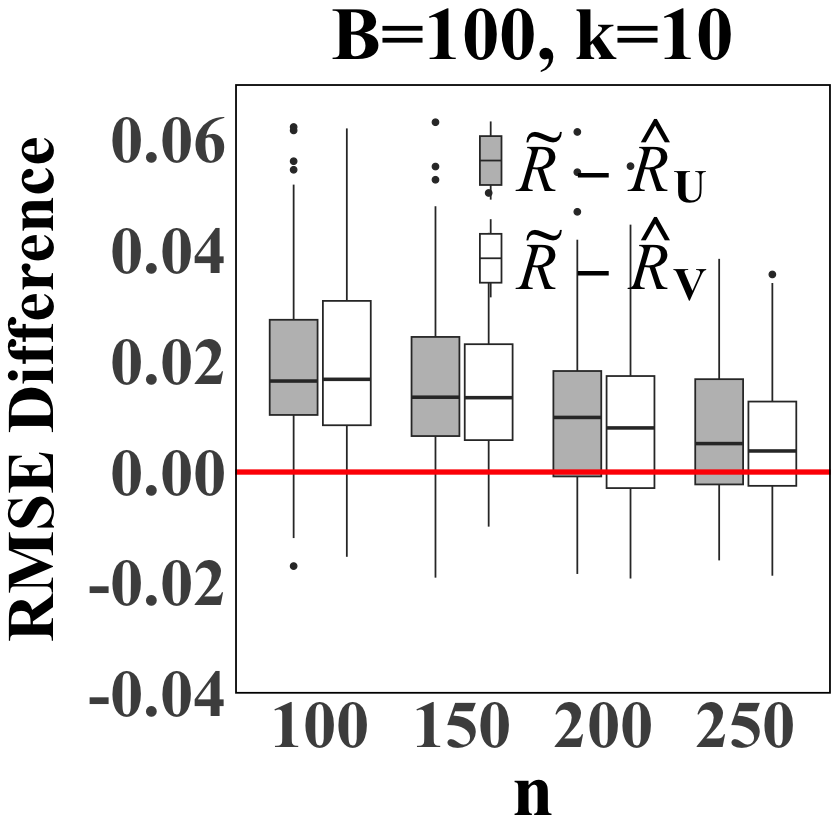} 
\includegraphics[width=0.15\textwidth]{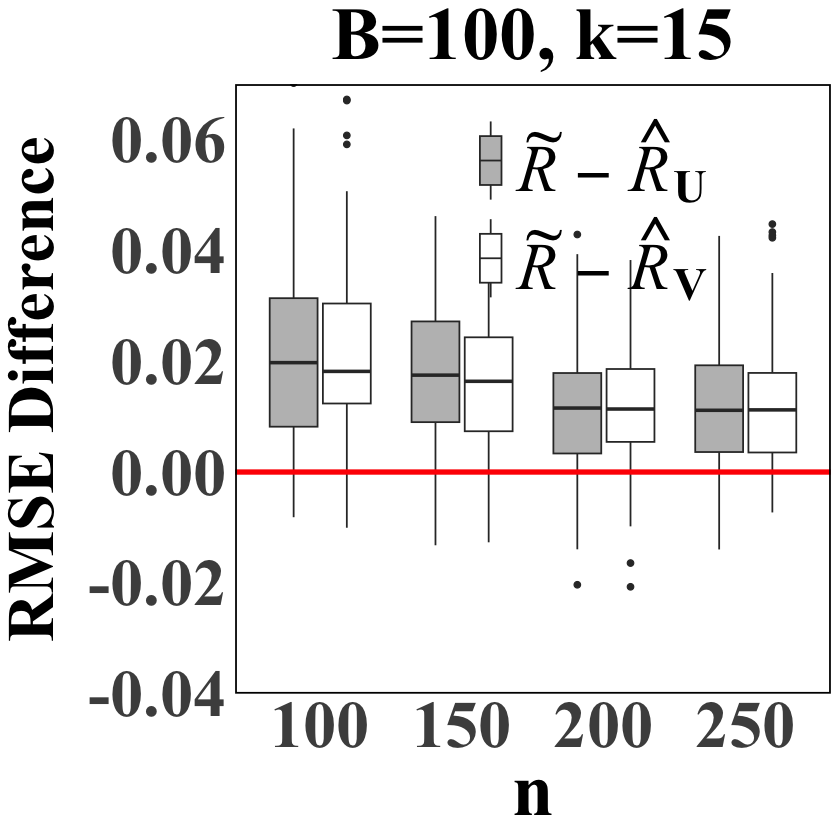} 
\includegraphics[width=0.15\textwidth]{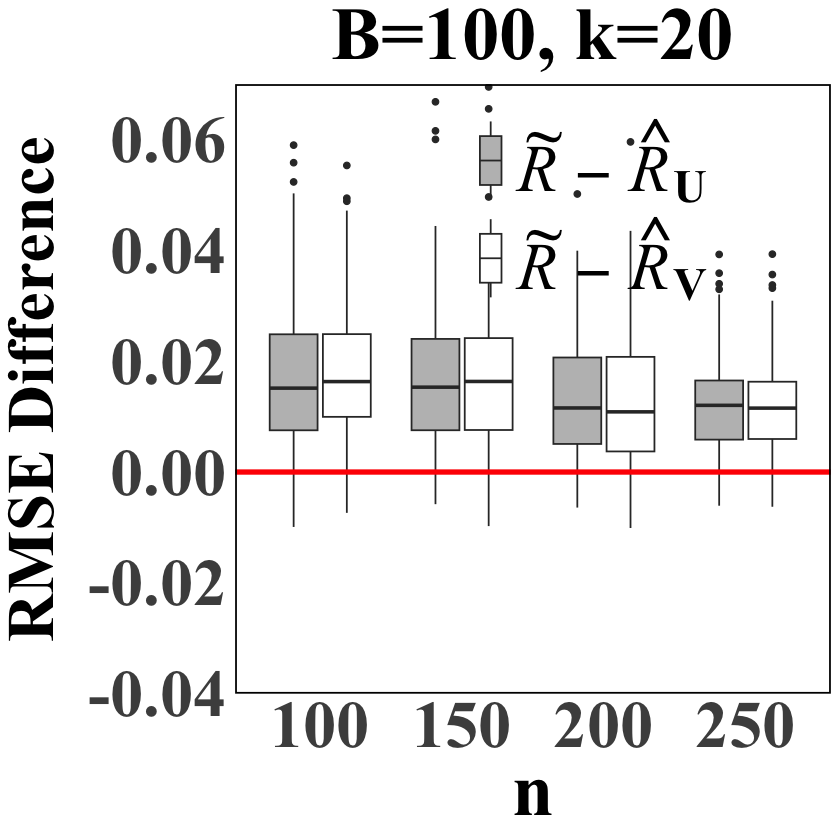} 
\caption{ RMSE differences among the predicted outcome values using the kernel ridge regression with $ m = 100 $ and $ M = 100 $, evaluated across various values of $ n $, $ B $, and $ k $. \textcolor{black}{$\tilde{R}$ denotes the RMSE without experience relay, while $\hat{R}_{U}$ and $\hat{R}_{V}$ represent the RMSE with experience replay.} The red line represents the baseline where the variance difference is 0.}
\label{fig:krmse}
\end{figure}

Figure \ref{fig:krmse} compares the RMSE by drawing the boxplots of the differences $\{\tilde{{R}}_i-\hat{{R}}_{i, U}\}_{i=1}^M$ and $\{\tilde{{R}}_i-\hat{{R}}_{i, V}\}_{i=1}^M$, with regard to different $n, B$, and the ratio $k/n$. We select the $n\in\{ 100, 150, 200, 250\}$, $B\in\{25,50, 100\}$, and $k\in\{10,15, 20\}$. The results indicate that incorporating experience replay not only reduces variance but also decreases errors for all the settings, particularly in data-scarce scenarios.


\end{document}